\documentclass{article}
\usepackage{amsmath, amssymb, amsthm}
\usepackage{graphicx}

\usepackage{geometry}
\usepackage{tikz}
\usepackage{float}
\usepackage{caption}
\usepackage{subcaption}
\usepackage{booktabs}
\usepackage{multirow}
\usepackage{longtable}
\usepackage{array}
\usepackage{ltcaption}
\usepackage{siunitx}
\usepackage{hyperref} % Add this line
\usepackage{url} % Add this line
\usepackage{longtable} 
\usepackage{mathtools}
\usepackage[authoryear]{natbib}
\hypersetup{
    colorlinks=true,
    linkcolor=blue,
    urlcolor=cyan,
}

\newtheorem{definition}{Definition}

\title{Ubiquitous Symmetry at Critical Points Across Diverse Optimization Landscapes}
\author{Irmi Schneider \\ 
  The Hebrew University of Jerusalem\\
  Department of Computer Science \\
  irmi.schneider@mail.huji.ac.il
}

\date{} 

\begin{document}
\maketitle
\begin{abstract}
Symmetry plays a crucial role in understanding the properties of mathematical structures and optimization problems. Recent work has 
explored this phenomenon in the context of neural networks, where the loss function is invariant under column and row permutations of the network weights. It has 
been observed that local minima exhibit significant symmetry 
with respect to the network weights 
(invariance to row and column permutations). And moreover no critical point was found that lacked symmetry.
We extend this line of inquiry by investigating symmetry 
phenomena in real-valued loss functions defined on a broader 
class of spaces. We will introduce four more cases: the projective case over a finite field, the octahedral graph case, the perfect matching case, and the particle attraction case. We show that as in the neural network case, all the critical points observed have non-trivial symmetry.
Finally we introduce a new measure of symmetry in the system and show that it reveals additional symmetry structures not captured by the previous measure.
\end{abstract}

\tableofcontents
\section{Introduction}

The optimization of neural networks is a central challenge in machine learning. Recent research has emphasized the importance of \emph{symmetry} in understanding the optimization landscape, particularly for shallow ReLU networks. In this context, \cite{arjevani2019principle} consider the population loss minimization problem
\begin{equation} \label{eq:nn_loss}
    \mathcal{L}(W) \coloneqq \frac{1}{2} \, \mathbb{E}_{x \sim \mathcal{N}(0,I_d)}\Bigl[\Bigl(1_d^\top \phi(Wx) - 1_d^\top \phi(Vx)\Bigr)^2\Bigr],
\end{equation}
where:
\begin{itemize}
    \item \(W \in M(d,d)\) is the weight matrix,
    \item \(V \in M(d,d)\) is the target weight matrix, fixed as \(V = I_d\),
    \item \(\phi \colon \mathbb{R} \to \mathbb{R}\) is the ReLU activation function, defined by \(\phi(z) \coloneqq \max\{0,z\}\),
    \item \(1_d \in \mathbb{R}^d\) is the vector of all ones.
\end{itemize}

Similarly, \cite{arjevani2021symmetry} investigate the symmetry of local minima in symmetric tensor decomposition. They study the optimization problem
\begin{equation} \label{eq:tensor_loss}
    \min_{W \in M(d,d)} L_\star(W) \coloneqq \left\|\sum_{i=1}^d w_i^{\otimes n} - T\right\|^2_\star,
\end{equation}
where:
\begin{itemize}
    \item \(w_i\) denotes the \(i\)th row of \(W\),
    \item \(\otimes n\) represents the \(n\)-fold tensor product,
    \item \(T\) is a target symmetric tensor, typically expressed as
    \[
    T = \sum_{i=1}^d v_i^{\otimes n},
    \]
    with \(v_i\) being the \(i\)th row of \(V = I_d\),
    \item \(\|\cdot\|_\star\) denotes a norm (e.g., the Frobenius norm or a Gaussian norm).
\end{itemize}

\subsection*{Invariance Under Permutations}

In both settings, the loss function is invariant under the action of the group
\[
S_d \times S_d,
\]
which acts on a \(d \times d\) weight matrix \(W\) by independently permuting its rows and columns. Formally, for any \((\pi_1,\pi_2) \in S_d \times S_d\) with corresponding permutation matrices \(P_{\pi_1}\) and \(P_{\pi_2}\), the action is defined as
\begin{equation} \label{eq:group_action}
    (\pi_1,\pi_2)\cdot W = P_{\pi_1}\, W\, P_{\pi_2}^{-1}.
\end{equation}
The \emph{stabilizer} (or isotropy group) of a weight matrix \(W\) is the subgroup of \(S_d \times S_d\) that leaves \(W\) invariant.

\subsubsection*{Empirical Observations}
\begin{figure}[h]
    \centering
    \includegraphics[width=0.6\textwidth]{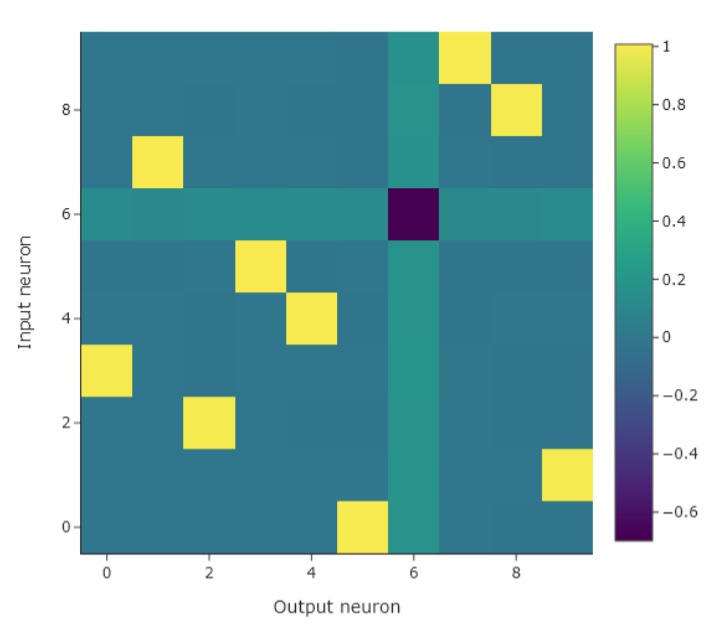}
    \caption{Typical minima observed for ReLU networks with an identity target matrix. The stabilizer under the \(S_d \times S_d\) action is conjugate to \(\Delta S_{d-1}\).}
    \label{fig:typical_minima}
\end{figure}

Empirical studies reveal that:
\begin{itemize}
    \item \textbf{ReLU Networks:} For networks with an identity target matrix (which has full isotropy group 
    \(\Delta S_d = \{(\pi,\pi) \mid \pi \in S_d\}\)), local minima consistently exhibit nontrivial stabilizers. In particular, the observed stabilizers are often conjugate to subgroups such as
    \[
    \Delta S_{d-1} \times \Delta S_1 \quad \text{or} \quad \Delta S_{d-2} \times \Delta S_2,
    \]
    which are maximal proper subgroups of \(\Delta S_d\). See Figure~\ref{fig:typical_minima} for an illustration.
    
    \item \textbf{Tensor Decomposition:} For the tensor decomposition problem with the target tensor
    \[
    T = \sum_{i=1}^d v_i^{\otimes n},
    \]
    where the \(v_i\)'s are the rows of \(V = I_d\), all local minima are observed to have stabilizers conjugate to subgroups such as
    \[
    \Delta(S_{d-1} \times S_1) \quad \text{or} \quad \Delta(S_{d-3} \times S_2 \times S_1),
    \]
    which are significant subgroups of the isotropy group of \(V\).
\end{itemize}

These findings suggest that optimization landscapes adhere to a principle of \emph{least symmetry breaking}---while spurious local minima may break the symmetry of the global minimum, they do so minimally. Indeed, after numerous optimization runs (using methods such as gradient descent or Adam), every critical point identified exhibits substantial, nontrivial symmetry.

\subsubsection*{Kernel Reformulation}
Both optimization problems admit an elegant reformulation in terms of kernel functions. Let \(\kappa \colon \mathbb{R}^d \times \mathbb{R}^d \to \mathbb{R}\) be a kernel function. Then, the loss functions can be expressed as
\begin{equation} \label{eq:kernel_loss}
    L(W) = \sum_{i,j=1}^d \kappa(w_i,w_j) - 2\sum_{i,j=1}^d \kappa(w_i,v_j) + \sum_{i,j=1}^d \kappa(v_i,v_j),
\end{equation}
where \(w_i\) and \(v_i\) denote the \(i\)th rows of \(W\) and \(V\), respectively.

For the ReLU network case with \(V = I_d\), the corresponding kernel is
\begin{equation} \label{eq:relu_kernel}
    \kappa_{\text{ReLU}}(w,v) = \mathbb{E}_{x \sim \mathcal{N}(0,I_d)}\Bigl[\phi(w\cdot x)\,\phi(v\cdot x)\Bigr].
\end{equation}

For the tensor decomposition problem under the Frobenius norm, the kernel is given by
\begin{equation} \label{eq:tensor_kernel}
    \kappa_{\text{tensor}}(w,v) = \langle w^{\otimes n},v^{\otimes n}\rangle_F = \langle w,v\rangle^n,
\end{equation}
where \(\langle \cdot,\cdot \rangle_F\) denotes the Frobenius inner product and \(\langle \cdot,\cdot \rangle\) is the standard Euclidean inner product.

\subsection{Main results of this paper}

Previous research has examined the loss function on shallow ReLU network and tensor decomposition optimization problem. Both are invariant under the $S_d \times S_d$ group action. A key finding was that for these loss functions, every observed local minimum exhibits substantial, nontrivial symmetry through its stabilizer group. In this work, we extend this investigation to several new entirely different symmetry groups.

\subsubsection*{Projective Space Case}

We begin by examining the projective linear group over a finite field $\text{PGL}(n+1, \mathbb{F}_q)$, where $n$ is the dimension of the projective space and $q$ is a prime number. This group acts naturally on the projective space $\mathbb{P}^n(\mathbb{F}_q)$.

We assign a real-valued function to the projective space and define a loss function that remains invariant under the induced action of $\text{PGL}(n+1, \mathbb{F}_q)$ on the space of real-valued functions. The loss function is constructed by summing pairwise interactions (kernel function) between restrictions of the real-valued function on pairs of projective hyperplanes.
We initialize the system and optimize the loss function. After optimization, we analyze the stabilizer (denoted as $I_V$) of the resulting configuration to 
understand which symmetries are preserved.
Our experiments across a wide range of dimensions $n$, polynomial degree $d$ of the kernel function, and prime numbers $q$ demonstrate that local minima of this loss function consistently possess non-trivial stabilizer subgroups of $\text{PGL}(n+1, \mathbb{F}_q)$.

\subsubsection*{Graph-Based Cases}

We investigate two specific graph structures:
\begin{itemize}
    \item \textbf{Octahedral Graph}: A 6-vertex, 12-edge undirected graph representing an octahedron
    \item \textbf{Perfect Matching Graph}: An undirected perfect matching graph
\end{itemize}

Though these graphs are undirected, we treat them as directed by assigning two directed edges $(i,j)$ and $(j,i)$ for each undirected edge $\{i,j\} \in E$ . The symmetry group acting on each graph consists of its automorphisms, i.e. bijections of the vertex set that preserve the edge structure.

For each vertex $i$ we assign a real value $v_i$, then for each directed edge $e = (i_1,i_2)$, we define an edge vector $(v_{i_1}, v_{i_2})$.
We then construct the loss function as the sum of a kernel function evaluated on all pairs of these edge vectors. 
This kernel function represents the interaction between edges.

We initialize the system and optimize the loss function. After optimization, we analyze the stabilizer (denoted as $I_V$) of the resulting configuration to 
understand which symmetries are preserved. Notably, we show that in both graph cases, every observed critical point demonstrates non-trivial, substantial symmetry.

\subsubsection*{Particle Attraction Case}

Finally, we explore a system of $n$ particles in $\mathbb{R}^2$. In this model, a kernel function defines the attraction and repulsion between particle pairs, and the overall loss aggregates all pairwise interactions. This loss function is invariant under the action of the symmetric group $S_n$ on particle positions. Our experiments show that almost all critical points possess non-trivial symmetry.
\subsubsection*{New Measure of Symmetry}

To capture more subtle symmetry properties, we introduce a new measure called the \textbf{edge isotropy group} ($I_E$). This group generalizes the vertex isotropy group $I_V$ by identifying all permutations of particles that preserve kernel interactions between pairs. Our results demonstrate that $I_E$ reveals additional symmetry structures not captured by $I_V$ alone.

We demonstrate this by showing that even in the few cases in the particle attraction model where $I_V$ symmetry is trivial, non-trivial $I_E$ symmetry exists. Furthermore, we introduce kernel functions with repulsive terms $1/d^2$ (where $d$ is distance between particles) that force $I_V$ symmetry to be trivial, and we show that they still maintain non-trivial $I_E$ symmetry, demonstrating the power of the $I_E$ measure.
\section{Optimization on the Projective Space}
\subsection{Introduction}

Projective spaces are fundamental geometric objects that arise by considering directions rather than points in a vector space. Formally, for a given field \(F\), the \(n\)-dimensional projective space \(\mathbb{P}^n(F)\) is defined as the set of all one-dimensional subspaces (i.e., lines through the origin) in \(F^{n+1}\). In other words,
\[
\mathbb{P}^n(F) = \left(F^{n+1} \setminus \{0\}\right) \big/ \sim,
\]
where 
\[
x \sim y \quad \text{if and only if} \quad \exists\, \lambda \in F^\times \text{ such that } y = \lambda x.
\]

For example, the projective line \(\mathbb{P}^1(\mathbb{R})\) consists of all lines through the origin in \(\mathbb{R}^2\) and can be identified with the extended real line \(\mathbb{R} \cup \{\infty\}\).

The symmetries of projective spaces are captured by the projective general linear group, denoted \(\mathrm{PGL}(n+1,F)\). This group is defined as the quotient
\[
\mathrm{PGL}(n+1,F) = \frac{\mathrm{GL}(n+1,F)}{F^\times},
\]
where \(\mathrm{GL}(n+1,F)\) is the group of all invertible \((n+1) \times (n+1)\) matrices over \(F\) and \(F^\times\) is the multiplicative group of nonzero elements in \(F\). The quotient removes the ambiguity of scalar multiples, ensuring that each element of \(\mathrm{PGL}(n+1,F)\) corresponds to a well-defined transformation on the set of lines. This group naturally acts on \(\mathbb{P}^n(F)\) by sending a point (a line)
\[
[x_0: x_1: \cdots : x_n]
\]
to
\[
g \cdot [x_0: x_1: \cdots : x_n] = \left[ g \begin{pmatrix} x_0 \\ x_1 \\ \vdots \\ x_n \end{pmatrix} \right],
\]
which is well-defined because any scalar multiple in \(g\) cancels out in the projective coordinates. 

\subsection{Loss Definition}
In this section, we explore high-symmetry phenomena in critical points by defining a loss function that is invariant under the projective general linear group \(G = \mathrm{PGL}(n+1, \mathbb{F}_q)\), where \(\mathbb{F}_q\) is a finite field. Consider the canonical embedding
\[
i: \mathbb{P}^{n-1}(\mathbb{F}_q) \hookrightarrow \mathbb{P}^n(\mathbb{F}_q), \quad i(x) = [x : 0].
\]
Let \(\mathcal{F}_m\) denote the space of real-valued functions on \(\mathbb{P}^m(\mathbb{F}_q)\). The group \(G\) acts on \(\mathcal{F}_m\) by
\[
(g \cdot f)(x) = f(g^{-1}x), \quad \forall\, f \in \mathcal{F}_m,\, g \in G.
\]
We define the restriction map \(r: \mathcal{F}_n \to \mathcal{F}_{n-1}\) as
\[
r(f)(x) = f(i(x)), \quad \forall\, x \in \mathbb{P}^{n-1}(\mathbb{F}_q).
\]
Endow \(\mathcal{F}_{n-1}\) with the inner product
\[
\langle f, g \rangle = \sum_{x \in \mathbb{P}^{n-1}(\mathbb{F}_q)} f(x) g(x),
\]
and define the kernel
\[
\kappa(f, g) = \langle f, g \rangle^d,
\]
for a fixed \(d \in \mathbb{N}\).

The loss function \(L: \mathcal{F}_n \to \mathbb{R}\) is then defined as
\begin{align}\label{eq:loss-function}
L(f) &= \sum_{g_1, g_2 \in G} \kappa\big(r(g_1 \cdot f), r(g_2 \cdot f)\big) \\
&\quad - 2 \sum_{g_1, g_2 \in G} \kappa\big(r(g_1 \cdot f), r(g_2 \cdot 1)\big) \notag \\
&\quad + \sum_{g_1, g_2 \in G} \kappa\big(r(g_1 \cdot 1), r(g_2 \cdot 1)\big), \notag
\end{align}
where \(1\) denotes the constant function on \(\mathbb{P}^n(\mathbb{F}_q)\).
For even \(d\), the loss satisfies \(L(f) \geq 0\) and attains its global minimum precisely when \(f \equiv \pm 1\).

\subsection{Results}

We conducted experiments by varying three key parameters: the dimension of the projective space, the finite field, and the degree of the kernel (i.e., half of the degree of the polynomial). For every configuration of these parameters, we define a loss function and then initialize numerous points and apply either gradient descent or Newton's method to converge to a critical point. We then analyze the stabilizer of this critical point.

For example, in the projective plane (\(n=2\)), we considered finite fields with 2, 3, 5, 7, and 11 elements. In these settings, we achieved convergence for all even-degree kernels ranging from 4 up to 14; furthermore, for the field with 2 elements, we also obtained convergence for kernels of degrees 50 and 100.

Similarly, in the three-dimensional projective space (\(n=3\)), our experiments over fields with 2 and 3 elements—for every even-degree kernel between 4 and 12—always revealed the symmetry phenomenon. In the case of four-dimensional projective spaces (\(n=4\)), we observed convergence over the field with 2 elements for kernels of degrees 4 to 14.

In all these experiments, most of the minima have fixed value on a projective hyperplane. Therefore the stabilizer at these critical points contains a substantial subgroup of permutations that move this hyperplane to itself. The points identified using Newton's method occasionally exhibited slightly reduced symmetry, but the overall pattern is clear: every critical point we observed possesses nontrivial symmetry.
We refer to Table~\ref{tab:f2_degree8} in the Appendix for the detailed results obtained on the projective plane over \(\mathbb{F}_2\) for \(n=2\) and \(d=8\).

\subsection{Results For Reformulated Loss}

We obtain similar results when considering an alternative formulation of the loss function that avoids using a target function. In this setting, we define a function 
\[
L:\mathbb{R}^{\mathbb{P}^n(\mathbb{F}_q)}  \to \mathbb{R}
\]
by
\[
L(f) = \sum_{g_1, g_2 \in G} \kappa\big(r(g_1 \cdot f), r(g_2 \cdot f)\big).
\]
Here, the kernel function \(\kappa\) is given by
\[
\kappa(f, g) = \langle f, g \rangle^d - c\, \langle f, g \rangle^q,
\]
where \(c > 0\) is a positive constant, \(d\) is an even integer, and \(q\) is an integer with \(q < d\). As in the previous cases, all the critical points identified by Newton's method and GD exhibit nontrivial symmetry.
\section{Optimization on the Octahedral Graph}
\subsection{Loss Definition}
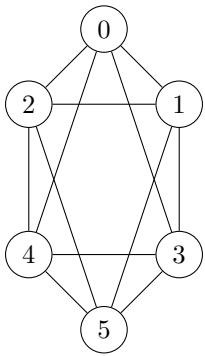
\begin{figure}[H]
    \centering
    \begin{tikzpicture}[scale=2, every node/.style={circle, draw, fill=white!20}]
    \node (2) at (0, 1) {2};
    \node (1) at (1, 1) {1};
    \node (4) at (0, 0) {4};
    \node (3) at (1, 0) {3};
    \node (0) at (0.5, 1.5) {0};
    \node (5) at (0.5, -0.5) {5};
    \draw (0) -- (1);
    \draw (0) -- (2);
    \draw (0) -- (3);
    \draw (0) -- (4);
    \draw (1) -- (2);
    \draw (1) -- (3);
    \draw (1) -- (5);
    \draw (2) -- (4);
    \draw (2) -- (5);
    \draw (3) -- (4);
    \draw (3) -- (5);
    \draw (4) -- (5);
    \end{tikzpicture}
    \caption{Octahedral Graph}
    \label{fig:octahedral_graph}
    \end{figure}
We study the octahedral graph, a 6-vertex, 12-edge undirected graph representing an octahedron.\\
Let \(G=(V,E)\) be a graph with vertex set 
\[
V = \{0,1,2,3,4,5\},
\]
and edge set:\\
\[
E = \{\{0,1\},\{0,2\},\{0,3\},\{0,4\},\{1,2\},\{1,3\},\{1,5\},\{2,4\},\{2,5\},\{3,4\},\{3,5\},\{4,5\}\}.\\
\]
A \emph{graph configuration} is a function \(v: V \to \mathbb{R}\), assigning a value \(v_i\) to each vertex \(i\). For each directed edge \((i,j)\), the corresponding \emph{edge configuration} is 
\[
a_{ij} = (v_i, v_j).
\]
The loss function is defined by
\[
L = \sum_{(i,j) \in E} \sum_{(k,l) \in E} \kappa(a_{ij}, a_{kl}),
\]
where \(k\) is a bounded-below polynomial kernel function.\\
We also define the \emph{vertex isotropy group} as
\[
I_V(v) = \{\sigma \in \operatorname{Aut}(G) \mid v(\sigma(i)) = v_i \text{ for all } i \in V\}.
\]

\subsubsection*{Kernel Function}

In our experiments, we use the kernel
\[
\kappa(\mathbf{a}, \mathbf{b}) = \left(\langle \mathbf{a}, \mathbf{b} \rangle\right)^p - c\,\left(\langle \mathbf{a}, \mathbf{b} \rangle\right)^q,
\]
with the standard inner product \(\langle \mathbf{a}, \mathbf{b} \rangle = a_1b_1+a_2b_2\), and parameters \(p=6\), \(q=4\), and \(c=7\).

\subsection{Results}

\subsubsection*{Gradient Descent Results}
In our experiments, GD always converged to configurations where all vertex values are equal or where one vertex differs from the others:
\begin{itemize}
\item All vertex values equal:
\[
\text{Loss} = -29269.0
\]
\[
v_0 = v_1 = v_2 = v_3 = v_4 = v_5 = 1.0393
\]

\item One vertex differs:
\[
v_0 = -1.0393, \quad v_1 = v_2 = v_3 = v_4 = v_5 = 1.0393
\]
\end{itemize}
\subsubsection*{Newton's Method Results}

Using Newton's Method for up to 200 times, we identified 8 critical points with varying isotropy groups.
The results are summarized in Table~\ref{tab:newton_results_octahedral}.

\begin{table}[H]
    \centering
    \caption{\textbf{Octahedral Graph} - Critical Points Found Using Newton's Method}
    \label{tab:newton_results_octahedral}
    \footnotesize
    \begin{tabular}{c l c c}
    \hline
    \textbf{Loss} & \textbf{Positions} $(v_0,v_1,v_2,v_3,v_4,v_5)$ & $I_V$ Order & $I_V$ Group Name \\ 
    \hline
    $-29269.0$ & $(1.039, 1.039, 1.039, 1.039, 1.039, 1.039)$ & 48 & $C_2 \times S_4$ \\ 
    \hline
    $-16261.0$ & $(-1.039, 1.039, -1.039, -1.039, -1.039, -1.039)$ & 8 & $D_4$ \\ 
    \hline
    $-8073.9$ & $(1.399, 0.758, 0.758, -0.315, -0.315, 1.399)$ & 4 & $C_2^2$ \\ 
    \hline
    $-7510.4$ & $(-0.279, 1.425, 0.685, 0.685, 1.425, -0.279)$ & 8 & $C_2^3$ \\ 
    \hline
    $-3225.5$ & $(-0.252, -0.252, -0.252, 1.085, 1.085, 1.085)$ & 6 & $S_3$ \\ 
    \hline
    $-2164.1$ & $(0.929, -0.192, -0.105, 1.436, -0.192, -0.200)$ & 2 & $C_2$ \\ 
    \hline
    $0$ & $(0, 0, 0, 0, 0, 0)$ & 48 & $C_2 \times S_4$ \\ 
    \hline
    \end{tabular}
    \end{table}

    \subsection*{Conclusion}
    As we saw in the matrices and in the projective case on the octahedral graph, both Gradient Descent and Newton's Method always converge to points with non-trivial isotropy groups.  
    \section{Optimization on the Perfect Matching Graph}
    \subsection{Loss Definition}
    \begin{figure}[H]
        \centering
        \begin{tikzpicture}[scale=1.5, every node/.style={circle, draw, fill=white!20}]
        % Define the positions of the vertices
        \node (0) at (0, 1) {0};
        \node (3) at (2, 1) {3};
        \node (1) at (0, 0) {1};
        \node (4) at (2, 0) {4};
        \node (2) at (0, -1) {2};
        \node (5) at (2, -1) {5};
        
        % Draw the edges (undirected perfect matching)
        \draw (0) -- (3);
        \draw (1) -- (4);
        \draw (2) -- (5);
        \end{tikzpicture}
        \caption{Perfect Matching Graph with \( n = 3 \)}
        \end{figure}
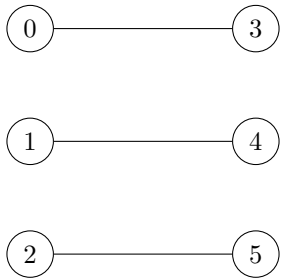
        We now shift our focus from the octahedral graph to the \emph{perfect matching graph}. For vectors \(\mathbf{a}, \mathbf{b} \in \mathbb{R}^2\), we define the kernel function as
        \[
        \kappa(\mathbf{a}, \mathbf{b}) = \left\|\mathbf{a} - \mathbf{b}\right\|^{2p} - c \, \left\|\mathbf{a} - \mathbf{b}\right\|^{2q},
        \]
        where \(p\) and \(q\) are parameters and \(c\) is a constant. The loss function \(L\) is then obtained by summing the kernel over all directed edges \((i,j)\) and \((k,l)\) in the edge set \(E\):
        \[
        L = \sum_{(i,j) \in E} \sum_{(k,l) \in E} \kappa(\mathbf{a}_{ij}, \mathbf{a}_{kl}).
        \]
        \subsection{Results}
        
        To identify graph configurations that minimize \(L\) and to study the symmetries preserved in these configurations, we conducted optimization experiments using Gradient Descent (GD) and Newton's Method. The results from Gradient Descent, ordered by the lowest (most negative) loss values, are presented in Table~\ref{tab:gd_results_PerfectMatching}.

    \subsubsection*{Gradient Descent Results}
    
    The GD results are presented in Table \ref{tab:gd_results_PerfectMatching}, 
    ordered by the lowest (most negative) loss values.
    \begin{table}[H]
        \centering
        \caption{\textbf{Perfect Matching Graph}: Results for Gradient Descent (GD)}
        \label{tab:gd_results_PerfectMatching}
        \footnotesize
        \begin{tabular}{@{}cccc@{}}
        \toprule
        \textbf{Loss} & $I_V$ Order & $I_V$ Group Name & \textbf{Positions} $(v_0,v_1,v_2,v_3,v_4,v_5)$ \\ \midrule
        $-1219.6$ & 4 & $C_2^2$ & $(1.246, -0.174, 0.865, 1.246, 0.865, -0.174)$ \\ \midrule
        $-914.67$ & 8 & $D_4$ & $(1.535, 0.115, 0.115, 0.496, 0.115, 0.115)$ \\ \midrule
        $-813.04$ & 8 & $D_4$ & $(0.916, 0.916, -0.123, 0.916, 0.916, -0.124)$ \\ \bottomrule
        \end{tabular}
        \end{table}
    
    \subsubsection*{Newton's Method Results}
    
    The critical points found using Newton's Method are presented in Table \ref{tab:newton_results_PerfectMatching}, ordered by the lowest (most negative) loss values.
    \begin{table}[H]
        \centering
        \caption{\textbf{Perfect Matching Graph}: Results for Newton's Method}
        \label{tab:newton_results_PerfectMatching}
        \footnotesize
        \begin{tabular}{@{}cccc@{}}
        \toprule
        \textbf{Loss} & $I_V$ Order & $I_V$ Group Name & \textbf{Positions} $(v_0,v_1,v_2,v_3,v_4,v_5)$ \\ \midrule
        $-534.89$ & 4 & $C_2^2$ & $(-0.012, -0.578, 0.848, -0.012, -0.578, -0.198)$ \\ \midrule
        $-415.78$ & 2 & $C_2$ & $(0.885, 0.321, -0.243, -0.155, 0.321, 0.796)$ \\ \midrule
        $-111.02$ & 8 & $D_4$ & $(1.172, 0.649, 0.649, 0.126, 0.649, 0.649)$ \\ \midrule
        $0.0$ & 48 & $C_2 \times S_4$ & $(0, 0, 0, 0, 0, 0)$ \\ \bottomrule
        \end{tabular}
        \end{table}
    \subsection*{Conclusion}
    As we saw in the previous perfect matching case, all the critical points found by Newton's Method and Gradient Descent have non-trivial isotropy groups.
    \section{Particle Attraction Case}
    \subsection{Loss Definition}
    Let $x_1, x_2, \ldots, x_n\in \mathbb{R}^2$ be the positions of $n$ particles in the plane.
    
    The kernel function, which represents the interaction between pairs of particles, is given by
    \[
    \kappa(\mathbf{a}, \mathbf{b}) = \left( \| \mathbf{a} - \mathbf{b} \|^2 \right)^6 - \left( \| \mathbf{a} - \mathbf{b} \|^2 \right)^4
    \]
    The loss function is given by
    \[
    L = \sum_{i=1}^{n} \sum_{j=1}^{n} \kappa(x_i, x_j)
    \]
    \subsection{Results}
    We optimize the loss function for \( n = 4 \) using Gradient Descent and Newton's Method to find the critical points.
    The results are summarized in Table \ref{table:experiment_results_ParticleAttraction_4}.
    \begin{table}[H]
    \centering
    
    \caption{\textbf{Particle Attraction Case}: Results for 4 particles for $k = d^{12} - d^8$ kernel}
    \label{table:experiment_results_ParticleAttraction_4}
    \small
    \begin{tabular}{c c c c c}
    \hline
    Critical point& Loss & $I_V$ Group Name & $I_V$ Order & Minimum? \\
    \hline
    1 & $-1.4815$ & $C_{2}$ & $2$ & Yes \\
    2 & $-1.1852$ & $C_{2}^{2}$ & $4$ & Yes \\
    3 & $-0.88889$ & $S_{3}$ & $6$ & Yes \\
    4 & $-0.94181$ & $I$ & $1$ & No \\
    5 & $-0.91969$ & $I$ & $1$ & No \\
    6 & $-0.60196$ & $C_{2}$ & $2$ & No \\
    7 & $-0.3098$ & $C_{2}$ & $2$ & No \\
    8 & $0$ & $S_{4}$ & $24$ & No \\
    \hline
    \end{tabular}
    \caption*{The corresponding plots for each critical point can be found in Figure~\ref{fig:plots_particle_attarciton_n_4_reg_ker} in the Appendix.}
    \end{table}
    
    \paragraph{Observations:}
    
    The results show that the vertex isotropy group \( I_V \) is non-trivial at most of the observed critical points.
    Later we will see that the only points (number 5, 6) with trivial symmetry have non-trivial edge isotropy group \( I_E \).
    
    We conducted the same experiments with $7$ particles. The results summarized in Table \ref{table:experiment_results_ParticleAttraction_7} 
    show that all the critical points have non-trivial isotropy groups.
\begin{table}[H]
\centering
\caption{\textbf{Particle Attraction Case}: - Results for 7 particles for $k = d^{12} - d^8$ kernel}
\label{table:experiment_results_ParticleAttraction_7}
\small
\begin{tabular}{c c c c c}
\hline
Critical point& Loss & $I_V$ Group Name & $I_V$ Order & Minimum? \\
\hline
1 & $-4.7407$ & $C_{2}^{2} \times S_{3}$ & $24$ & Yes \\
2 & $-4.4444$ & $S_{3}^{2}$ & $36$ & Yes \\
3 & $-2.963$ & $C_{2} \times S_{5}$ & $240$ & Yes \\
4 & $0$ & $S_{7}$ & $5040$ & No \\
5 & $-4.1605$ & $D_{6}$ & $12$ & No \\
6 & $-3.8667$ & $D_{6}$ & $12$ & No \\
7 & $-3.5771$ & $D_{6}$ & $12$ & No \\
8 & $-3.0253$ & $C_{2}^{2}$ & $4$ & No \\
9 & $-2.9985$ & $C_{2}^{3}$ & $8$ & No \\
10 & $-2.9799$ & $S_{4}$ & $24$ & No \\
11 & $-2.7265$ & $C_{2}^{2}$ & $4$ & No \\
12 & $-2.715$ & $S_{4}$ & $24$ & No \\
13 & $-2.6967$ & $D_{6}$ & $12$ & No \\
14 & $-2.4711$ & $C_{2}^{2}$ & $4$ & No \\
15 & $-2.3891$ & $S_{3}$ & $6$ & No \\
16 & $-2.1642$ & $C_{2}^{2}$ & $4$ & No \\
17 & $-1.8107$ & $C_{2}$ & $2$ & No \\
18 & $-1.6567$ & $C_{2}^{2}$ & $4$ & No \\
19 & $-1.6279$ & $S_{3}$ & $6$ & No \\
20 & $-1.5994$ & $D_{6}$ & $12$ & No \\
21 & $-1.5652$ & $C_{2}^{2}$ & $4$ & No \\
22 & $-1.4331$ & $C_{2}$ & $2$ & No \\
23 & $-1.3177$ & $C_{2}^{2}$ & $4$ & No \\
24 & $-1.2911$ & $S_{3}$ & $6$ & No \\
25 & $-1.226$ & $S_{3}$ & $6$ & No \\
26 & $-1.2107$ & $C_{2}^{2}$ & $4$ & No \\
27 & $-0.9973$ & $S_{3}$ & $6$ & No \\
28 & $-0.92304$ & $S_{3}^{2}$ & $36$ & No \\
29 & $-0.63063$ & $C_{2} \times S_{4}$ & $48$ & No \\
30 & $-0.33078$ & $S_{5}$ & $120$ & No \\
\hline
\end{tabular}
\caption*{The corresponding plots for each critical point can be found in Figure~\ref{fig:plots_particle_attarciton_n_7_reg_ker} in the Appendix.}

\end{table}
\paragraph{Observations:}

All the critical points except two points in the $n=4$ case, have non-trivial isotropy groups, and most of them have large symmetry.
We will see later that the two points with the trivial $I_V$ symmetry have other ($I_E$) non-trivial symmetry.
\subsection{Robustness Under Varying Kernel Functions and High-Degree Polynomials}
In our particle attraction experiments, the loss function is generated by a kernel function, 
which in our initial setup is based on the distance between particles. 
However, our observations indicate that the resulting behavior is not restricted solely to distance-based kernels. 
In fact, when we replace the distance with the inner product between particles, 
and even when considering the inner product raised to high powers 
(e.g., the inner product to the 50th, resulting in the loss being a polynomial of degree 100), the critical points also always possessed non-trivial high symmetry. In fact, in these cases, 
the critical points appear as a perturbation of the critical points in lower degree polynomial kernels.

Moreover, it seems that similar phenomena occur when the kernel function is a randomly chosen polynomial. 
For instance, in the appendix~\ref{app:RandomKer}, we fixed the kernel as a bounded below polynomial of degree 4. 
Our experiments again reveal that the critical points in those cases always possess large symmetry. 
This robustness persists regardless of whether the kernel is derived from the distance or from an inner product or perhaps (needs more experiments) from a randomly chosen polynomial.
\subsection{Narrow Attraction Basins of Non-Symmetric Minima (if they indeed exist)}

In all the cases we have studied—whether it’s the projective plane, the octahedral graph, or the perfect matching case—every local minimum observed exhibited non-trivial symmetry. In those experiments, roughly 1,000 trials per case consistently yielded symmetric minima. To rigorously test the possibility that a non-symmetric minimum might exist, we selected one representative scenario using an inner-product-based kernel powered by 8 (i.e., a 16th-degree polynomial) and ran an extensive series of about 30 million experiments. Not a single one of these runs produced a non-symmetric minimum; all converged to one of the four symmetric minima we had already identified.

This overwhelming evidence suggests that, while we cannot theoretically rule out the existence of non-symmetric minima, if they do exist their basins of attraction must be extraordinarily narrow—so much so that they are effectively undetectable even under exhaustive search. For further details, see Appendix~\ref{app:RandomPolynomialExperiment1}.

\section{New Measure of Symmetry - Edge Isotropy Group \( I_E \)}
\subsection{Introduction}
In this section, we will define and demonstrate the importance of the edge isotropy group \( I_E \) 
in capturing symmetries that are not evident from the $I_V$ group alone.
We start by examining the edge isotropy group for the particle attraction loss function as defined in the previous section.
We showed that the group \( I_E \) measures symmetries that the group \( I_V \) does not capture
in the case of $4$ and $7$ particles.
\subsection{Edge Isotropy Group Definition \( I_E \)}  
\begin{definition}[Edge Isotropy Group \( I_E \)]
The \emph{edge isotropy group} \( I_E(a) \) consists of automorphisms that preserve the kernel values between edge pairs:

\[
I_E(a) = \left\{ \sigma \in \mathrm{Aut}(G) \mid k\left( \mathbf{a}_{\sigma(i)\sigma(j)}, \mathbf{a}_{\sigma(k)\sigma(l)} \right) = k\left( \mathbf{a}_{ij}, \mathbf{a}_{kl} \right) \quad \forall (i,j),(k,l) \in E \right\}
\]
\end{definition}
Obviously we have the relation \( I_V \subset I_E \), because each permutation that fixes the vertex values also fixes the kernel values between edges.
However, as we will see in the following examples, the edge isotropy group \( I_E \) in many cases is larger than the vertex isotropy group \( I_V \).
\subsection{$I_E$ Isotropy Group for \( n = 4 \), \( n = 7 \) ($\kappa(x,y) = ||x-y||^{12} - ||x-y||^8$ kernel)}
We begin by presenting the previous tables for $4$ and $7$ particles experiments.
We have added a column of the edge isotropy group \( I_E \) and its order.

The extended results for \( n = 4 \) are summarized in Table \ref{table:experiment_results_ParticleAttraction_regular_kernel_4}.\begin{table}[H]
\centering
\caption{Extended table with $I_E$ groups, $n=4$, $\kappa(x,y) = ||x-y||^{12} - ||x-y||^8$ kernel}
\label{table:experiment_results_ParticleAttraction_regular_kernel_4}
\small
\begin{tabular}{c c c c c c c}
\hline
Critical point& Loss & $I_V$ Group Name & $I_V$ Order & $I_E$ Group Name & $I_E$ Order & Minimum? \\
\hline
1 & $-1.4815$ & $C_{2}$ & $2$ & $C_{2}^{2}$ & $4$ & Yes \\
2 & $-1.1852$ & $C_{2}^{2}$ & $4$ & $D_{4}$ & $8$ & Yes \\
3 & $0$ & $S_{4}$ & $24$ & $S_{4}$ & $24$ & No \\
4 & $-0.94181$ & $I$ & $1$ & $C_{2}$ & $2$ & No \\
5 & $-0.91969$ & $I$ & $1$ & $S_{3}$ & $6$ & No \\
6 & $-0.60196$ & $C_{2}$ & $2$ & $C_{2}$ & $2$ & No \\
7 & $-0.3098$ & $C_{2}$ & $2$ & $C_{2}^{2}$ & $4$ & No \\
8 & $-0.88889$ & $S_{3}$ & $6$ & $S_{3}$ & $6$ & Yes \\
\hline
\end{tabular}
\caption*{The corresponding plots for each critical point can be found in Figure~\ref{fig:plots_particle_attarciton_n_4_reg_ker} in the Appendix.}

\end{table}
The extended results for \( n = 7 \) are summarized in Table \ref{table:experiment_results_ParticleAttraction_regular_kernel_7}.
\begin{table}[H]
\centering
\caption{Extended table with $I_E$ groups, $n=7$, $\kappa(x,y) = ||x-y||^{12} - ||x-y||^8$ kernel}
\label{table:experiment_results_ParticleAttraction_regular_kernel_7}
\small
\begin{tabular}{c c c c c c c}
\hline
Critical point& Loss & $I_V$ Group Name & $I_V$ Order & $I_E$ Group Name & $I_E$ Order & Minimum? \\
\hline
1 & $-4.7407$ & $C_{2}^{2} \times S_{3}$ & $24$ & $D_{4} \times S_{3}$ & $48$ & Yes \\
2 & $-4.4444$ & $S_{3}^{2}$ & $36$ & $(S_{3}^{2} \ltimes C_{2})$ & $72$ & Yes \\
3 & $-2.963$ & $C_{2} \times S_{5}$ & $240$ & $C_{2} \times S_{5}$ & $240$ & Yes \\
4 & $0$ & $S_{7}$ & $5040$ & $S_{7}$ & $5040$ & No \\
5 & $-4.1605$ & $D_{6}$ & $12$ & $D_{6}$ & $12$ & No \\
6 & $-3.8667$ & $D_{6}$ & $12$ & $D_{6}$ & $12$ & No \\
7 & $-3.5771$ & $D_{6}$ & $12$ & $D_{6}$ & $12$ & No \\
8 & $-3.0253$ & $C_{2}^{2}$ & $4$ & $C_{2}^{2}$ & $4$ & No \\
9 & $-2.9985$ & $C_{2}^{3}$ & $8$ & $C_{2} \times D_{4}$ & $16$ & No \\
10 & $-2.9799$ & $S_{4}$ & $24$ & $S_{4}$ & $24$ & No \\
11 & $-2.7265$ & $C_{2}^{2}$ & $4$ & $C_{2}^{3}$ & $8$ & No \\
12 & $-2.715$ & $S_{4}$ & $24$ & $C_{2} \times S_{4}$ & $48$ & No \\
13 & $-2.6967$ & $D_{6}$ & $12$ & $D_{6}$ & $12$ & No \\
14 & $-2.4711$ & $C_{2}^{2}$ & $4$ & $D_{4}$ & $8$ & No \\
15 & $-2.3891$ & $S_{3}$ & $6$ & $S_{3}$ & $6$ & No \\
16 & $-2.1642$ & $C_{2}^{2}$ & $4$ & $C_{2}^{3}$ & $8$ & No \\
17 & $-1.8107$ & $C_{2}$ & $2$ & $C_{2}^{2}$ & $4$ & No \\
18 & $-1.6567$ & $C_{2}^{2}$ & $4$ & $D_{4}$ & $8$ & No \\
19 & $-1.6279$ & $S_{3}$ & $6$ & $S_{3}$ & $6$ & No \\
20 & $-1.5994$ & $D_{6}$ & $12$ & $C_{2}^{2} \times S_{3}$ & $24$ & No \\
21 & $-1.5652$ & $C_{2}^{2}$ & $4$ & $C_{2}^{3}$ & $8$ & No \\
22 & $-1.4331$ & $C_{2}$ & $2$ & $C_{2}^{2}$ & $4$ & No \\
23 & $-1.3177$ & $C_{2}^{2}$ & $4$ & $C_{2}^{2}$ & $4$ & No \\
24 & $-1.2911$ & $S_{3}$ & $6$ & $D_{6}$ & $12$ & No \\
25 & $-1.226$ & $S_{3}$ & $6$ & $C_{2}^{2} \times S_{3}$ & $24$ & No \\
26 & $-1.2107$ & $C_{2}^{2}$ & $4$ & $C_{2}^{3}$ & $8$ & No \\
27 & $-0.9973$ & $S_{3}$ & $6$ & $D_{6}$ & $12$ & No \\
28 & $-0.92304$ & $S_{3}^{2}$ & $36$ & $S_{3}^{2}$ & $36$ & No \\
29 & $-0.63063$ & $C_{2} \times S_{4}$ & $48$ & $C_{2} \times S_{4}$ & $48$ & No \\
30 & $-0.33078$ & $S_{5}$ & $120$ & $C_{2} \times S_{5}$ & $240$ & No \\
\hline
\end{tabular}
\caption*{The corresponding plots for each critical point can be found in Figure~\ref{fig:plots_particle_attarciton_n_7_reg_ker} in the Appendix.}
\end{table}
\paragraph{Observations:}
We can see that the edge isotropy group \( I_E \) is consistently larger than the vertex isotropy group \( I_V \) in the experiments.
Furthermore, in the case of $4$ particles, for the $5, 6$ critical points, the vertex isotropy group \( I_V \) is trivial, but the edge isotropy group \( I_E \) is non-trivial.
\subsection{Experiments for \( n = 4 \) and \( n = 7 \) with $\kappa= \|\mathbf{a} - \mathbf{b}\|^{2} + \frac{1}{\|\mathbf{a} - \mathbf{b}\|^2}$ Kernel}
The power of $I_E$ and the idea that it generalizes the symmetries in $I_V$ is reflected in the experiments where our kernel function is given by:

\[
\kappa(\mathbf{a}, \mathbf{b}) = 
\begin{cases}
\| \mathbf{a} - \mathbf{b} \|^2 + \dfrac{1}{\| \mathbf{a} - \mathbf{b} \|^2}, & \text{if } \mathbf{a} \neq \mathbf{b} \\
0, & \text{if } \mathbf{a} = \mathbf{b}
\end{cases}
\]

In this case, the vertex isotropy group \( I_V \) is trivial (\( I_V \) order = 1) for all experiments, as expected due to the repulsion term \( \frac{1}{\|\mathbf{a} - \mathbf{b}\|^2} \) in the kernel. However, the edge isotropy group \( I_E \) is non-trivial, indicating deeper symmetries in the edge interactions.\begin{table}[H]
\centering
\caption{Results for \( n = 4 \) ( $d^2+ \frac{1}{d^2}$ Kernel function)}
\label{table:experiment_results_repelling_particles_4}
\small
\begin{tabular}{c c c c c c c}
\hline
Critical point& Loss & $I_V$ Group Name & $I_V$ Order & $I_E$ Group Name & $I_E$ Order & Minimum? \\
\hline
1 & $25.298$ & $I$ & $1$ & $D_{4}$ & $8$ & Yes \\
2 & $27.708$ & $I$ & $1$ & $C_{2}$ & $2$ & No \\
3 & $33.941$ & $I$ & $1$ & $C_{2}$ & $2$ & No \\
\hline
\end{tabular}
\caption*{The corresponding plots for each critical point can be found in Figure~\ref{fig:plots_particle_attarciton_n_4_repelling_ker} in the Appendix.}

\end{table}

And here is results for the same kernel for \( n = 7 \):\\
\begin{table}[H]
\centering
\caption{Results for \( n = 7 \) ( $d^2+ \frac{1}{d^2}$ Kernel function)}

\label{table:experiment_results_repelling_particles_7}

\small
\begin{tabular}{c c c c c c c}
\hline
Critical point& Loss & $I_V$ Group Name & $I_V$ Order & $I_E$ Group Name & $I_E$ Order & Minimum? \\
\hline
1 & $99.559$ & $I$ & $1$ & $D_{6}$ & $12$ & Yes \\
2 & $102.97$ & $I$ & $1$ & $C_{2}$ & $2$ & No \\
3 & $103.15$ & $I$ & $1$ & $C_{2}$ & $2$ & No \\
4 & $103.35$ & $I$ & $1$ & $C_{2}$ & $2$ & No \\
5 & $104.77$ & $I$ & $1$ & $D_{7}$ & $14$ & No \\
6 & $105.95$ & $I$ & $1$ & $C_{2}$ & $2$ & No \\
7 & $109.12$ & $I$ & $1$ & $C_{2}$ & $2$ & No \\
8 & $113.02$ & $I$ & $1$ & $C_{2}^{2}$ & $4$ & No \\
9 & $114.91$ & $I$ & $1$ & $C_{2}$ & $2$ & No \\
10 & $121.85$ & $I$ & $1$ & $C_{2}$ & $2$ & No \\
11 & $131.51$ & $I$ & $1$ & $C_{2}$ & $2$ & No \\
12 & $157.15$ & $I$ & $1$ & $C_{2}$ & $2$ & No \\
\hline
\end{tabular}
\caption*{The corresponding plots for each critical point can be found in Figure~\ref{fig:plots_particle_attarciton_n_7_repelling_ker} in the Appendix.}

\end{table}

\paragraph{Observations:}

Similar to the \( n = 4 \) case, while \( I_V \) is trivial, \( I_E \) remains non-trivial.
\subsection{Summary of the Edge Isotropy Group $I_E$}
In this section, we introduced the edge isotropy group $I_E$, which generalizes the vertex isotropy group $I_V$ by identifying more complex symmetry structures in optimization landscapes. Through our experiments with particle attraction cases involving 4 and 7 particles, we found that $I_E$ consistently captures symmetries that $I_V$ alone does not detect. Notably, several scenarios exhibited trivial $I_V$ groups while revealing non-trivial $I_E$ groups, highlighting deeper symmetries embedded within edge interactions.

Furthermore, this observation remained valid even for kernels explicitly containing repulsive terms designed to eliminate vertex-level symmetries. Such findings underscore the robustness and importance of the $I_E$ group as an analytical tool. The consistent observation that $I_E$ is larger than $I_V$ across various configurations suggests that $I_E$ is a promising generalization of symmetry measurement. Ultimately, this generalized symmetry measure can enable a broader exploration of symmetry phenomena, potentially uncovering the underlying mechanisms that drive symmetry in critical points.\section{Summary}
In this work, we investigate symmetry phenomena in a range of optimization settings by examining the isotropy groups of the critical points. Building on earlier studies in ReLU networks and symmetric tensor decomposition, we consider four distinct cases: the projective space over finite fields, the octahedral graph, the perfect matching graph, and the particle attraction model. In each scenario, both gradient descent and Newton's method always converge to local minima that exhibit substantial non-trivial symmetry. Moreover, our experiments show that even when varying the kernel functions or employing high-degree polynomial kernels, the optimization process always converges to points with non-trivial symmetry, indicating that the basins of attraction for any non-symmetric minima (if they indeed exist) are negligibly small. To capture subtle symmetries that are not revealed by vertex configurations alone, we finally introduce the \emph{Edge Isotropy Group} ($I_E$). This additional metric provides deeper insight into the symmetric structure of the minima of loss landscapes and reinforces the principle of least symmetry breaking observed throughout our experiments. Detailed experimental data and further analyses supporting these findings are presented in the Appendix.
\section*{Acknowledgements}
I would like to express my sincere gratitude to David Kazhdan, Yossi Arjevani, and Shmuel Weinberger for their invaluable guidance and insightful discussions throughout this work.
% \newpage
\clearpage
\addcontentsline{toc}{section}{Appendix}
\appendix
\section*{Appendix}
\section{Detailed Experimental Data for the Projective Plane over the Field \(\mathbb{F}_2\)}
In this appendix, we present detailed experimental data for the projective plane over the field \(\mathbb{F}_2\). Let \(v\in \mathcal{F}_n\) and denote 
\[
v_1 = v((0,0,1)), \quad v_2 = v((0,1,0)), \quad v_3 = v((0,1,1)),
\]
\[
\quad v_4 = v((1,0,0)), \quad v_5 = v((1,0,1)), \\\quad v_6 = v((1,1,0)), \quad v_7 = v((1,1,1)).
\]
\begin{longtable}{c c c c}
    \caption{Critical points Data for Field \(\mathbb{F}_2\) and Polynomial Degree 16} \label{tab:f2_degree8} \\
    \hline
    \textbf{Loss} & \textbf{Weights} & \(I_V\) (stabilizer group) Order & \(I_V\) Group Name \\
    \hline
    \endfirsthead
    \multicolumn{4}{c}{Table \thetable{} continued} \\
    \hline
    \textbf{Loss} & \textbf{Weights} & \(I_V\) Order & \(I_V\) Group Name \\
    \hline
    \endhead
        0 & \begin{tabular}{@{}c@{}}
        \scriptsize
        $v_{1}$ $v_{2}$ $v_{3}$ $v_{4}$ $v_{5}$ $v_{6}$ $v_{7}$ = 1.0 \\
        \end{tabular} & 168 & $\text{PGL}(3, 2)$ \\
        \hline
        21079.60 & \begin{tabular}{@{}c@{}}
        \scriptsize
        $v_{1}$ $v_{2}$ = 1.27492 \\
        \scriptsize
        $v_{3}$ = 1.27494 \\
        \scriptsize
        $v_{4}$ $v_{7}$ = 0.03937 \\
        \scriptsize
        $v_{5}$ $v_{6}$ = -0.03567 \\
        \end{tabular} & 4 &$C_2\times C_2$ \\
        \hline
        21079.610 & \begin{tabular}{@{}c@{}}
        \scriptsize
        $v_{1}$ $v_{2}$ = 1.27493 \\
        \scriptsize
        $v_{3}$ = 1.27492 \\
        \scriptsize
        $v_{4}$ = -0.05162 \\
        \scriptsize
        $v_{5}$ $v_{6}$ = 0.0025 \\
        \scriptsize
        $v_{7}$ = 0.05402 \\
        \end{tabular} & 2 & $C_2$\\
        \hline
        21079.615 & \begin{tabular}{@{}c@{}}
        \scriptsize
        $v_{1}$ $v_{2}$ $v_{3}$ = 1.27493 \\
        \scriptsize
        $v_{4}$ = -0.06263 \\
        \scriptsize
        $v_{5}$ $v_{6}$ $v_{7}$ = 0.02335 \\
        \end{tabular} & 6 & $S_3$ \\
        \hline
        21079.616 & \begin{tabular}{@{}c@{}}
        \scriptsize
        $v_{1}$ $v_{2}$ $v_{3}$ = 1.27493 \\
        \scriptsize
        $v_{4}$ $v_{6}$ $v_{7}$ = -0.01963 \\
        \scriptsize
        $v_{5}$ = 0.0663 \\
        \end{tabular} & 6 & $S_3$ \\
        \hline
        21080 & \begin{tabular}{@{}c@{}}
        \scriptsize
        $v_{1}$ $v_{2}$ $v_{3}$ = 1.27493 \\
        \scriptsize
        $v_{4}$ $v_{5}$ $v_{6}$ $v_{7}$ = 0.0018 \\
        \end{tabular} & 24 & $S_4$ \\
        \hline
        93233 & \begin{tabular}{@{}c@{}}
        \scriptsize
        $v_{1}$ $v_{2}$ $v_{3}$ = 1.2173 \\
        \scriptsize
        $v_{4}$ $v_{5}$ $v_{6}$ = 0.82001 \\
        \scriptsize
        $v_{7}$ = -0.10076 \\
        \end{tabular} & 6 & $S_3$ \\
        \hline
        104452 & \begin{tabular}{@{}c@{}}
        \scriptsize
        $v_{1}$ $v_{2}$ $v_{3}$ = 1.24105 \\
        \scriptsize
        $v_{4}$ $v_{5}$ $v_{6}$ $v_{7}$ = 0.61958 \\
        \end{tabular} & 24 & $S_4$ \\
        \hline
        11306411 & \begin{tabular}{@{}c@{}}
        \scriptsize
        $v_{1}$ $v_{2}$ $v_{3}$ = 0.0 \\
        \scriptsize
        $v_{4}$ $v_{5}$ = 1.09527 \\
        \scriptsize
        $v_{6}$ $v_{7}$ = -1.09527 \\
        \end{tabular} & 4 &$C_2\times C_2$ \\
        \hline
        11528833 & \begin{tabular}{@{}c@{}}
        \scriptsize
        $v_{1}$ $v_{2}$ = 0.99946 \\
        \scriptsize
        $v_{3}$ = -0.2527 \\
        \scriptsize
        $v_{4}$ $v_{5}$ $v_{6}$ $v_{7}$ = -0.67966 \\
        \end{tabular} & 8 & $D_4$ \\
        \hline
        11544284 & \begin{tabular}{@{}c@{}}
        \scriptsize
        $v_{1}$ $v_{2}$ = 0.7481 \\
        \scriptsize
        $v_{3}$ = -0.92777 \\
        \scriptsize
        $v_{4}$ $v_{7}$ = -0.34603 \\
        \scriptsize
        $v_{5}$ = 1.15453 \\
        \scriptsize
        $v_{6}$ = -0.01432 \\
        \end{tabular} & 2 & $C_2$\\
        \hline
        11560780 & \begin{tabular}{@{}c@{}}
        \scriptsize
        $v_{1}$ $v_{2}$ = 0.51586 \\
        \scriptsize
        $v_{3}$ = -0.46628 \\
        \scriptsize
        $v_{4}$ = -0.89189 \\
        \scriptsize
        $v_{5}$ $v_{6}$ = 0.91016 \\
        \scriptsize
        $v_{7}$ = -0.10749 \\
        \end{tabular} & 2 & $C_2$\\
        \hline
        11564812 & \begin{tabular}{@{}c@{}}
        \scriptsize
        $v_{1}$ $v_{2}$ $v_{5}$ = 0.87152 \\
        \scriptsize
        $v_{3}$ $v_{4}$ $v_{7}$ = -0.45644 \\
        \scriptsize
        $v_{6}$ = -0.08012 \\
        \end{tabular} & 6 & $S_3$ \\
        \hline
        11565260 & \begin{tabular}{@{}c@{}}
        \scriptsize
        $v_{1}$ $v_{2}$ = 0.29492 \\
        \scriptsize
        $v_{3}$ = 0.79675 \\
        \scriptsize
        $v_{4}$ $v_{5}$ $v_{6}$ $v_{7}$ = -0.7562 \\
        \end{tabular} & 8 & $D_4$ \\
        \hline
        11566743 & \begin{tabular}{@{}c@{}}
        \scriptsize
        $v_{1}$ $v_{2}$ $v_{6}$ = 0.61893 \\
        \scriptsize
        $v_{3}$ $v_{4}$ $v_{7}$ = -0.44323 \\
        \scriptsize
        $v_{5}$ = 1.0662 \\
        \end{tabular} & 6 & $S_3$ \\
        \hline
        11567466 & \begin{tabular}{@{}c@{}}
        \scriptsize
        $v_{1}$ $v_{2}$ = 0.74245 \\
        \scriptsize
        $v_{3}$ = -0.3452 \\
        \scriptsize
        $v_{4}$ $v_{7}$ = -0.48573 \\
        \scriptsize
        $v_{5}$ = 0.95005 \\
        \scriptsize
        $v_{6}$ = 0.54832 \\
        \end{tabular} & 2 & $C_2$\\
        \hline
        11567945 & \begin{tabular}{@{}c@{}}
        \scriptsize
        $v_{1}$ $v_{2}$ = 0.83873 \\
        \scriptsize
        $v_{3}$ = -0.49864 \\
        \scriptsize
        $v_{4}$ $v_{7}$ = 0.68525 \\
        \scriptsize
        $v_{5}$ $v_{6}$ = -0.40378 \\
        \end{tabular} & 4 &$C_2\times C_2$ \\
        \hline
        11568165 & \begin{tabular}{@{}c@{}}
        \scriptsize
        $v_{1}$ $v_{2}$ $v_{3}$ = -0.43366 \\
        \scriptsize
        $v_{4}$ $v_{5}$ $v_{6}$ $v_{7}$ = 0.76953 \\
        \end{tabular} & 24 & $S_4$ \\
        \hline
        11571824 & \begin{tabular}{@{}c@{}}
        \scriptsize
        $v_{1}$ $v_{2}$ = 0.60545 \\
        \scriptsize
        $v_{3}$ = -0.93276 \\
        \scriptsize
        $v_{4}$ $v_{7}$ = -0.07086 \\
        \scriptsize
        $v_{5}$ = 0.53502 \\
        \scriptsize
        $v_{6}$ = -0.6828 \\
        \end{tabular} & 2 & $C_2$\\
        \hline
        11572799 & \begin{tabular}{@{}c@{}}
        \scriptsize
        $v_{1}$ $v_{2}$ = 0.75893 \\
        \scriptsize
        $v_{3}$ = -0.57575 \\
        \scriptsize
        $v_{4}$ $v_{7}$ = -0.18882 \\
        \scriptsize
        $v_{5}$ = 0.35018 \\
        \scriptsize
        $v_{6}$ = -0.75595 \\
        \end{tabular} & 2 & $C_2$\\
        \hline
        11573312 & \begin{tabular}{@{}c@{}}
        \scriptsize
        $v_{1}$ = 0.85652 \\
        \scriptsize
        $v_{2}$ $v_{3}$ $v_{4}$ $v_{6}$ $v_{7}$ = -0.0 \\
        \scriptsize
        $v_{5}$ = -0.85652 \\
        \end{tabular} & 4 &$C_2\times C_2$ \\
        \hline
        11573350 & \begin{tabular}{@{}c@{}}
        \scriptsize
        $v_{1}$ $v_{2}$ = 0.23222 \\
        \scriptsize
        $v_{3}$ = 0.19656 \\
        \scriptsize
        $v_{4}$ $v_{7}$ = -0.21457 \\
        \scriptsize
        $v_{5}$ = 0.81487 \\
        \scriptsize
        $v_{6}$ = -0.86003 \\
        \end{tabular} & 2 & $C_2$\\
        \hline
        11573387 & \begin{tabular}{@{}c@{}}
        \scriptsize
        $v_{1}$ $v_{2}$ = 0.33387 \\
        \scriptsize
        $v_{3}$ = 0.0 \\
        \scriptsize
        $v_{4}$ $v_{7}$ = -0.33387 \\
        \scriptsize
        $v_{5}$ = -0.81544 \\
        \scriptsize
        $v_{6}$ = 0.81544 \\
        \end{tabular} & 2 & $C_2$\\
        \hline
        11573462 & \begin{tabular}{@{}c@{}}
        \scriptsize
        $v_{1}$ = 0.93886 \\
        \scriptsize
        $v_{2}$ = -0.59133 \\
        \scriptsize
        $v_{3}$ = 0.39297 \\
        \scriptsize
        $v_{4}$ $v_{5}$ $v_{6}$ $v_{7}$ = -0.0883 \\
        \end{tabular} & 4 &$C_2\times C_2$ \\
        \hline
        11573479 & \begin{tabular}{@{}c@{}}
        \scriptsize
        $v_{1}$ $v_{2}$ = 0.24957 \\
        \scriptsize
        $v_{3}$ = -0.42817 \\
        \scriptsize
        $v_{4}$ $v_{7}$ = -0.09292 \\
        \scriptsize
        $v_{5}$ = -0.89686 \\
        \scriptsize
        $v_{6}$ = 0.60173 \\
        \end{tabular} & 2 & $C_2$\\
        \hline
        11573551 & \begin{tabular}{@{}c@{}}
        \scriptsize
        $v_{1}$ $v_{2}$ = 0.5086 \\
        \scriptsize
        $v_{3}$ = -0.8155 \\
        \scriptsize
        $v_{4}$ $v_{5}$ $v_{6}$ $v_{7}$ = 0.06123 \\
        \end{tabular} & 8 & $D_4$ \\
        \hline
        11573569 & \begin{tabular}{@{}c@{}}
        \scriptsize
        $v_{1}$ $v_{2}$ = 0.60785 \\
        \scriptsize
        $v_{3}$ = -0.63187 \\
        \scriptsize
        $v_{4}$ $v_{5}$ $v_{6}$ $v_{7}$ = -0.00935 \\
        \end{tabular} & 8 & $D_4$ \\
        \hline
        11573599 & \begin{tabular}{@{}c@{}}
        \scriptsize
        $v_{1}$ = 0.7865 \\
        \scriptsize
        $v_{2}$ $v_{3}$ $v_{4}$ $v_{5}$ $v_{6}$ $v_{7}$ = -0.15037 \\
        \end{tabular} & 24 & $S_4$ \\
        \hline
        11573604 & \begin{tabular}{@{}c@{}}
        \scriptsize
        $v_{1}$ $v_{2}$ $v_{3}$ $v_{4}$ $v_{5}$ $v_{6}$ $v_{7}$ = -0.0 \\
        \end{tabular} & 168 & $\text{PGL}(3, 2)$ \\
        \hline
    \end{longtable}

\section{Plots for particle attraction case}
\subsection{Plots for 4 particles Experiments (kernel $d^{12} - d^8$)}
The plots are ordered in the order of the experiments in the table.\\
% I wnat to start figure counting again from 1

\begin{figure}[H]
\caption{Visualization of critical points from Table~\ref{table:experiment_results_ParticleAttraction_4}}
\label{fig:plots_particle_attarciton_n_4_reg_ker}% \centering
% % % Row 1
\begin{minipage}{0.45\textwidth}
\centering
\includegraphics[width=\linewidth]{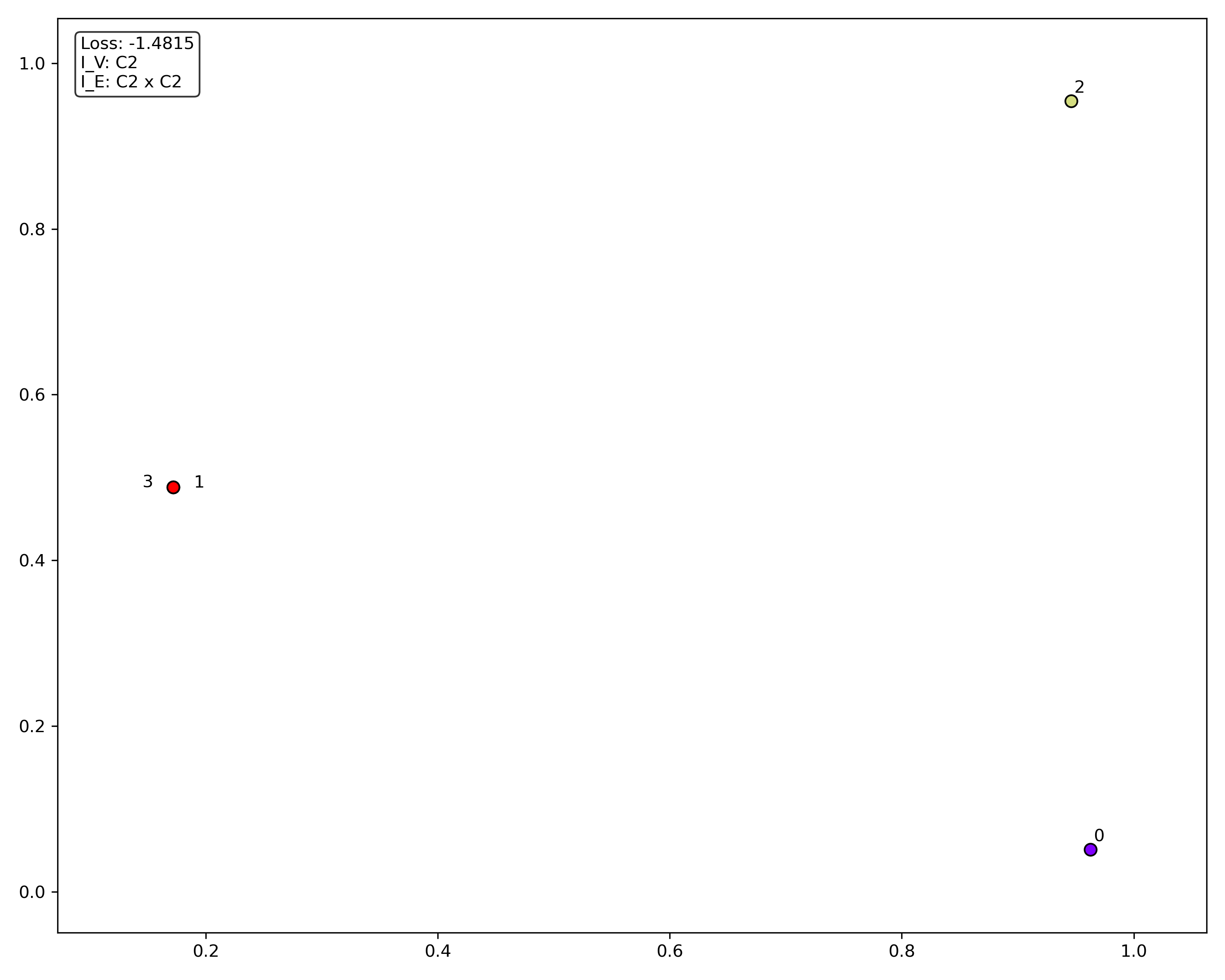}
\caption{}
\end{minipage}\hfill
\begin{minipage}{0.45\textwidth}
\centering
\includegraphics[width=\linewidth]{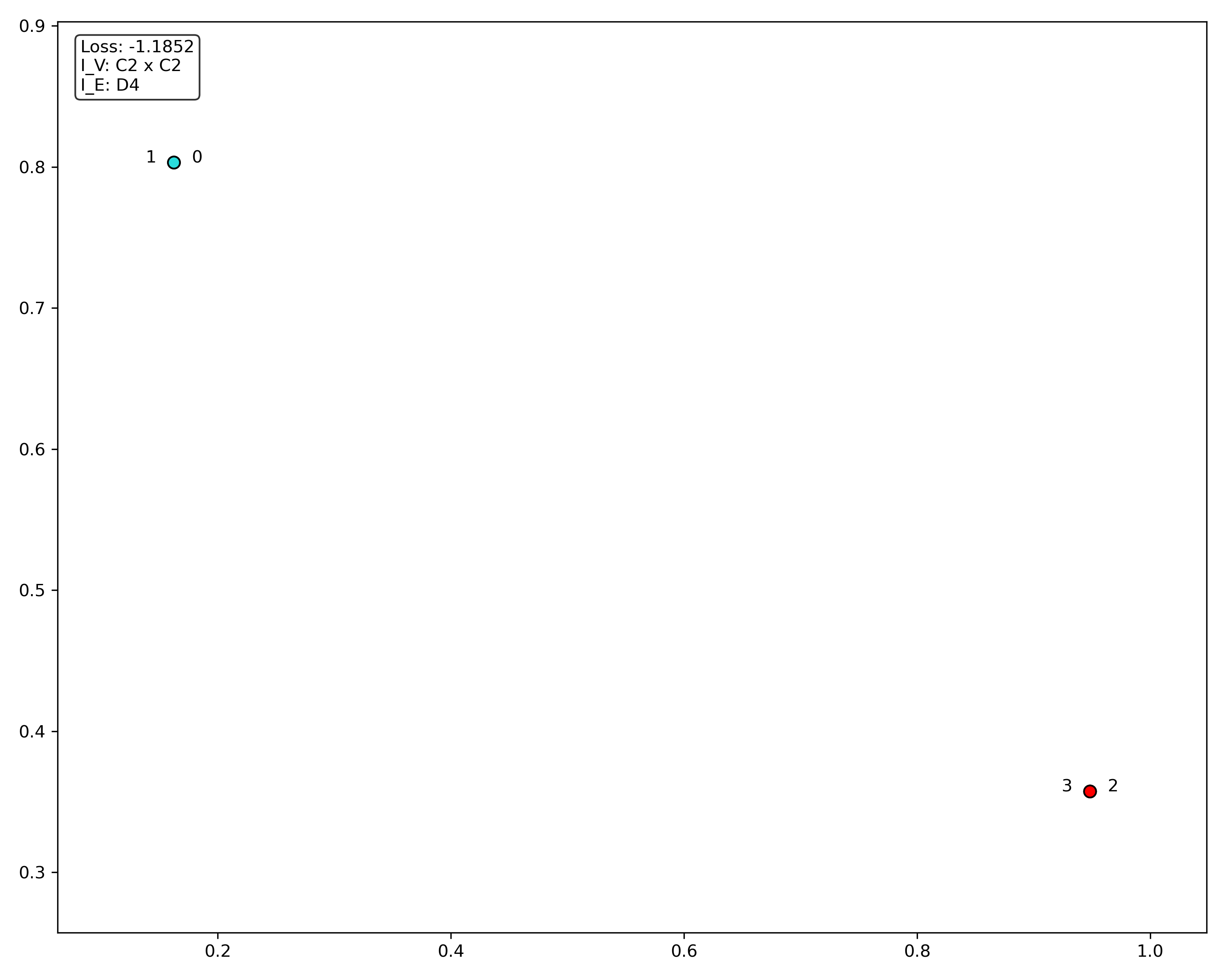}
\caption{}
\end{minipage}
\end{figure}
\begin{figure}[H]
% Row 2
\begin{minipage}{0.45\textwidth}
\centering
\includegraphics[width=\linewidth]{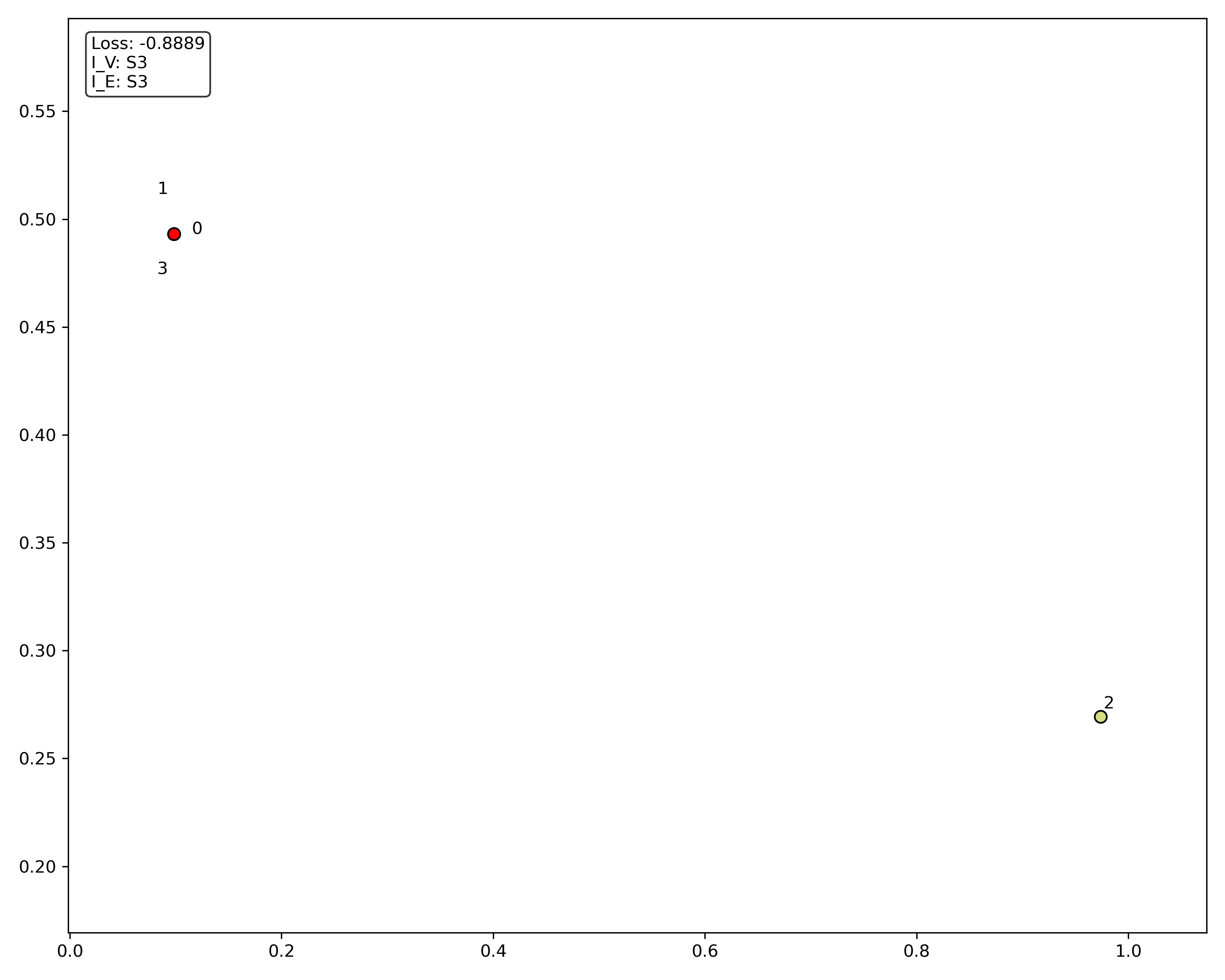}
\caption{}
\end{minipage}\hfill
\begin{minipage}{0.45\textwidth}
\centering
\includegraphics[width=\linewidth]{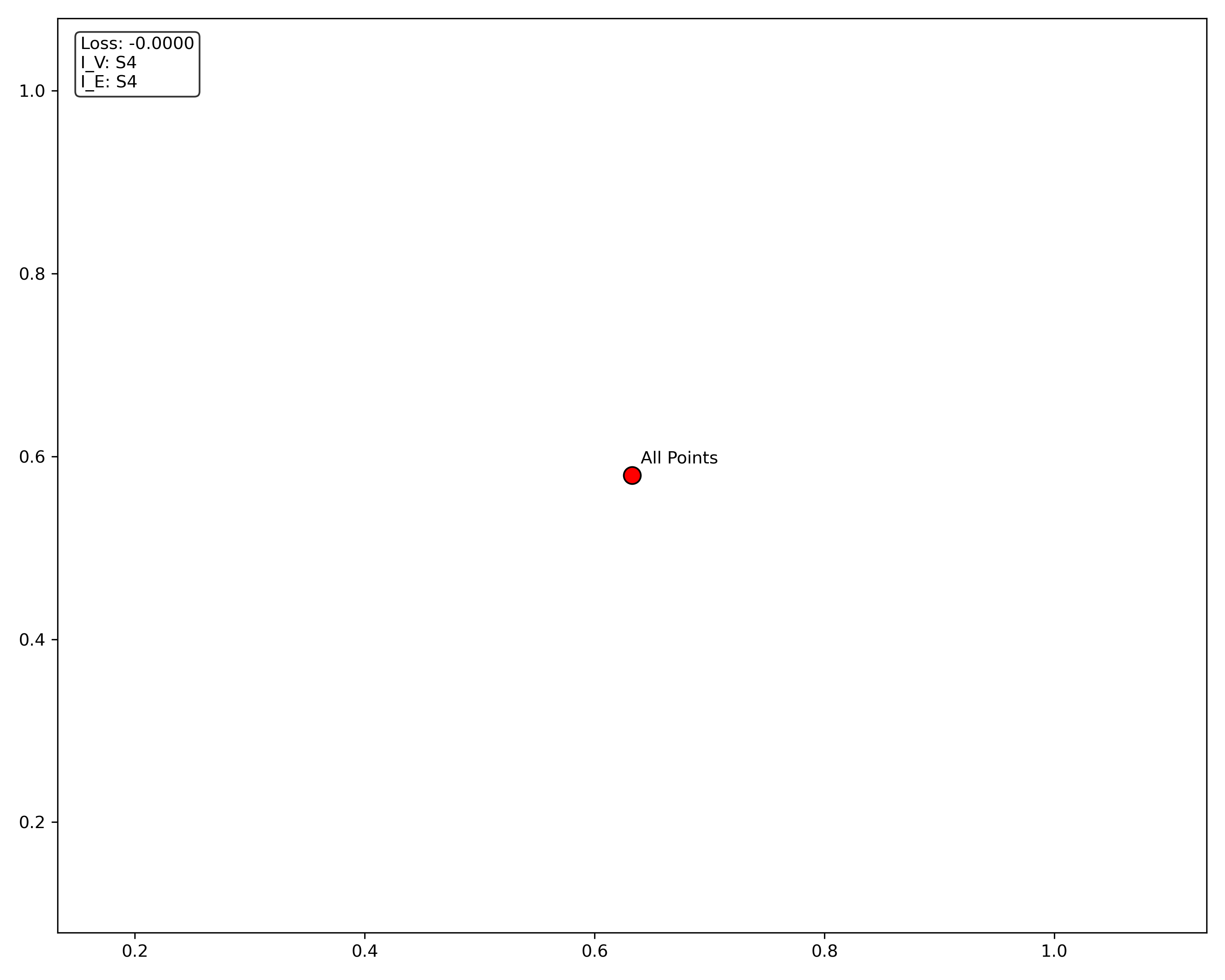}
\caption{}
\end{minipage}
\end{figure}
\begin{figure}[H]
% Row 3
\begin{minipage}{0.45\textwidth}
\centering
\includegraphics[width=\linewidth]{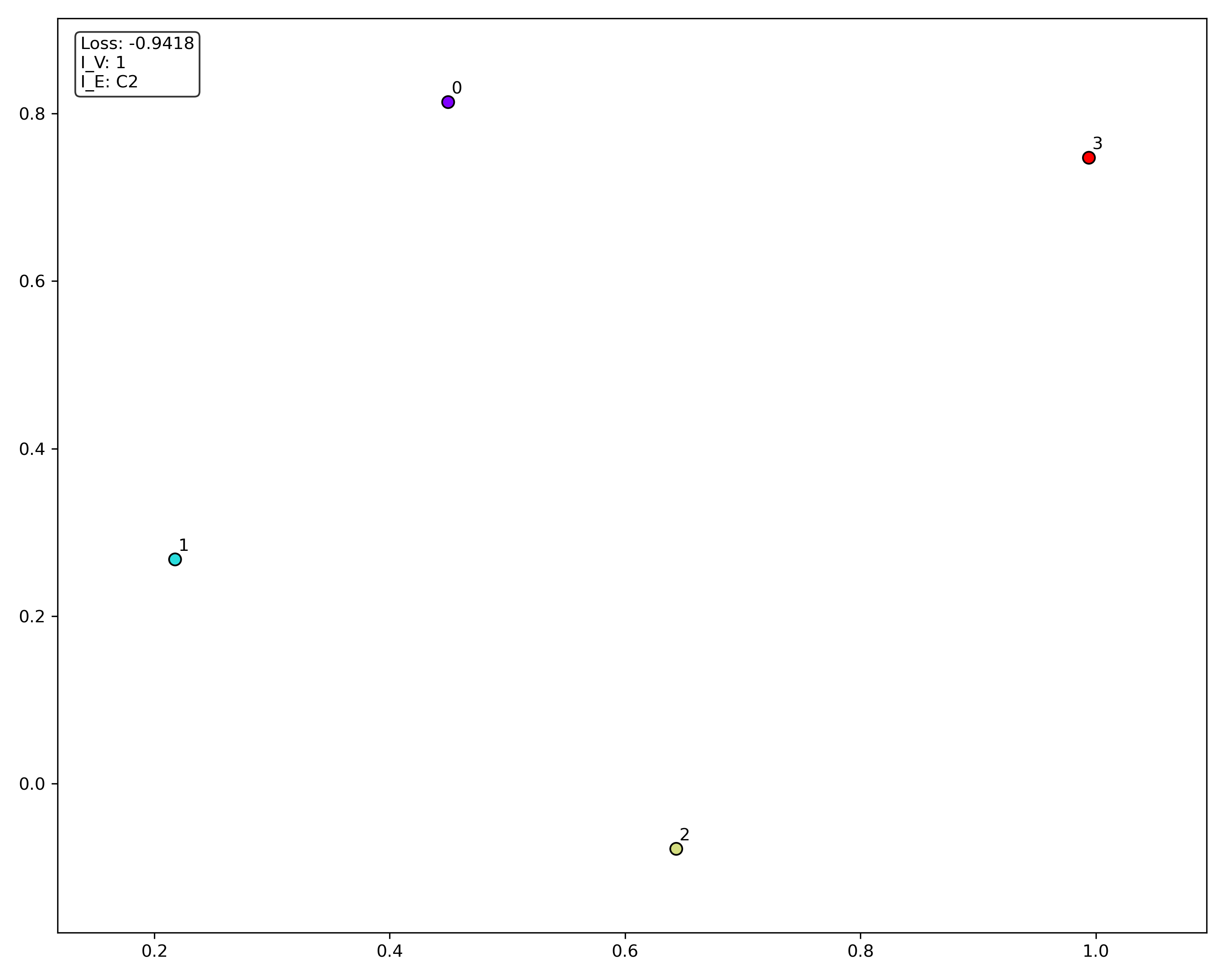}
\caption{}
\end{minipage}\hfill
\begin{minipage}{0.45\textwidth}
\centering
\includegraphics[width=\linewidth]{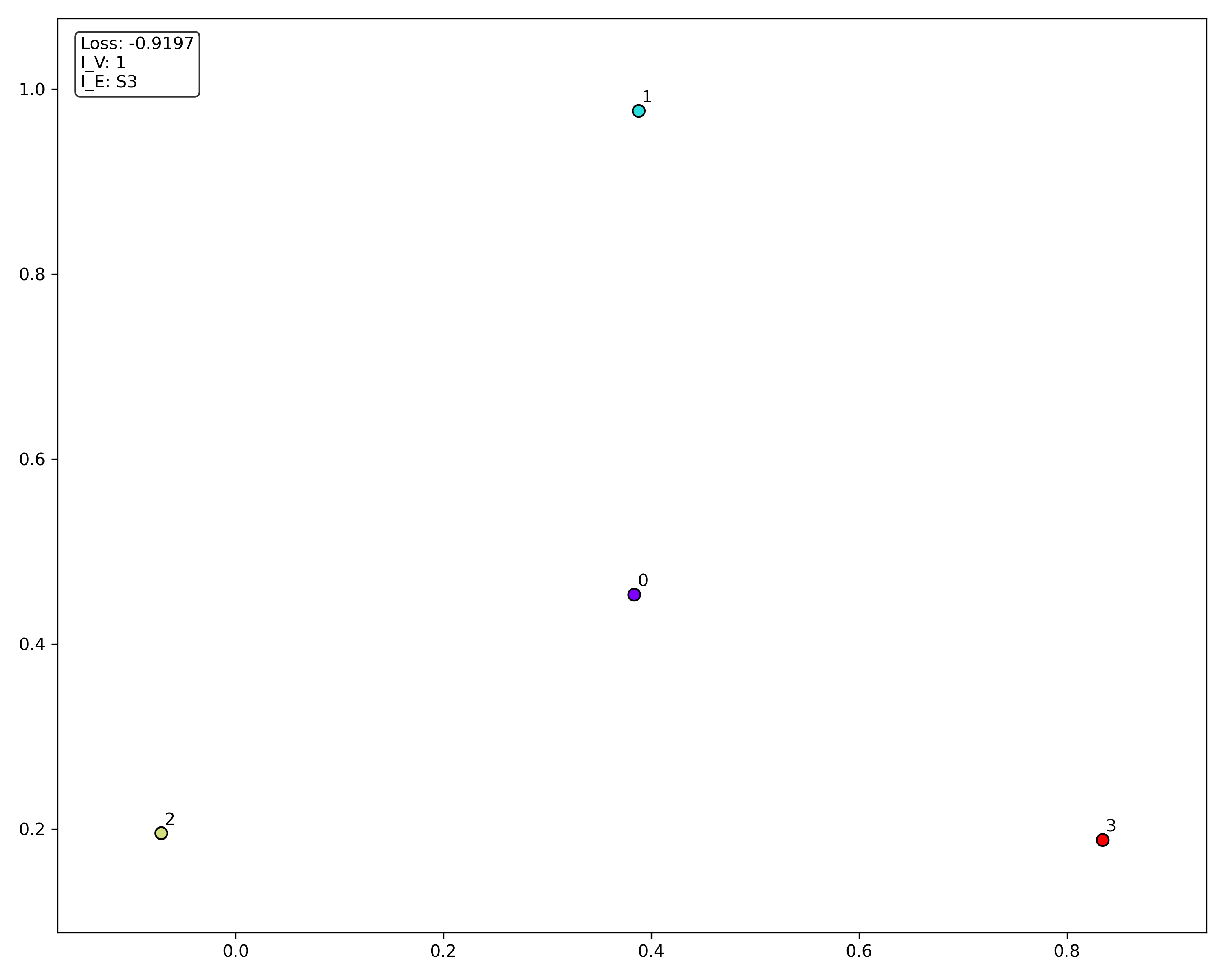}
\caption{}
\end{minipage}

\end{figure}
\begin{figure}[H]
\centering
% Row 1
\begin{minipage}{0.45\textwidth}
\centering
\includegraphics[width=\linewidth]{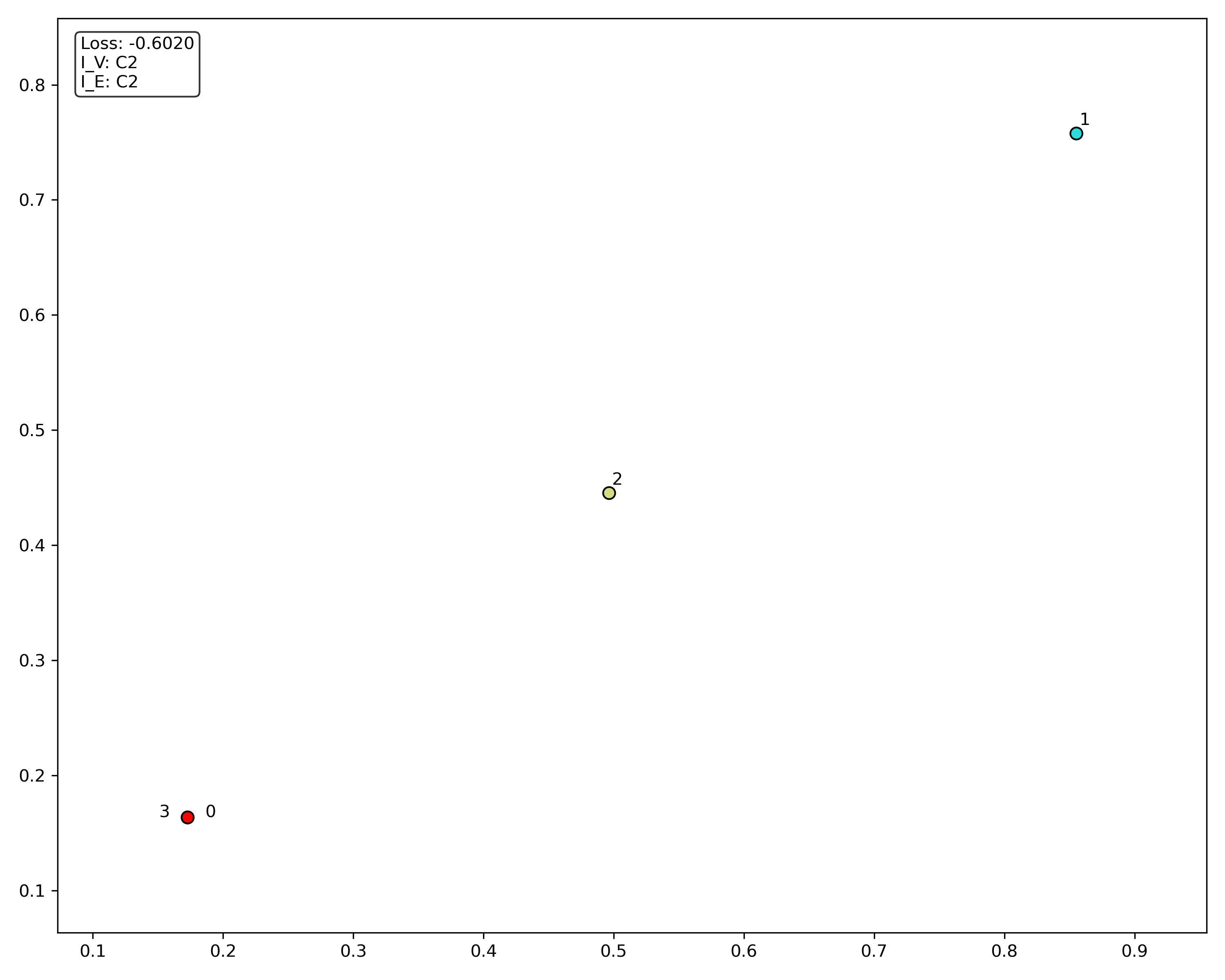}
\caption{}
\end{minipage}\hfill
\begin{minipage}{0.45\textwidth}
\centering
\includegraphics[width=\linewidth]{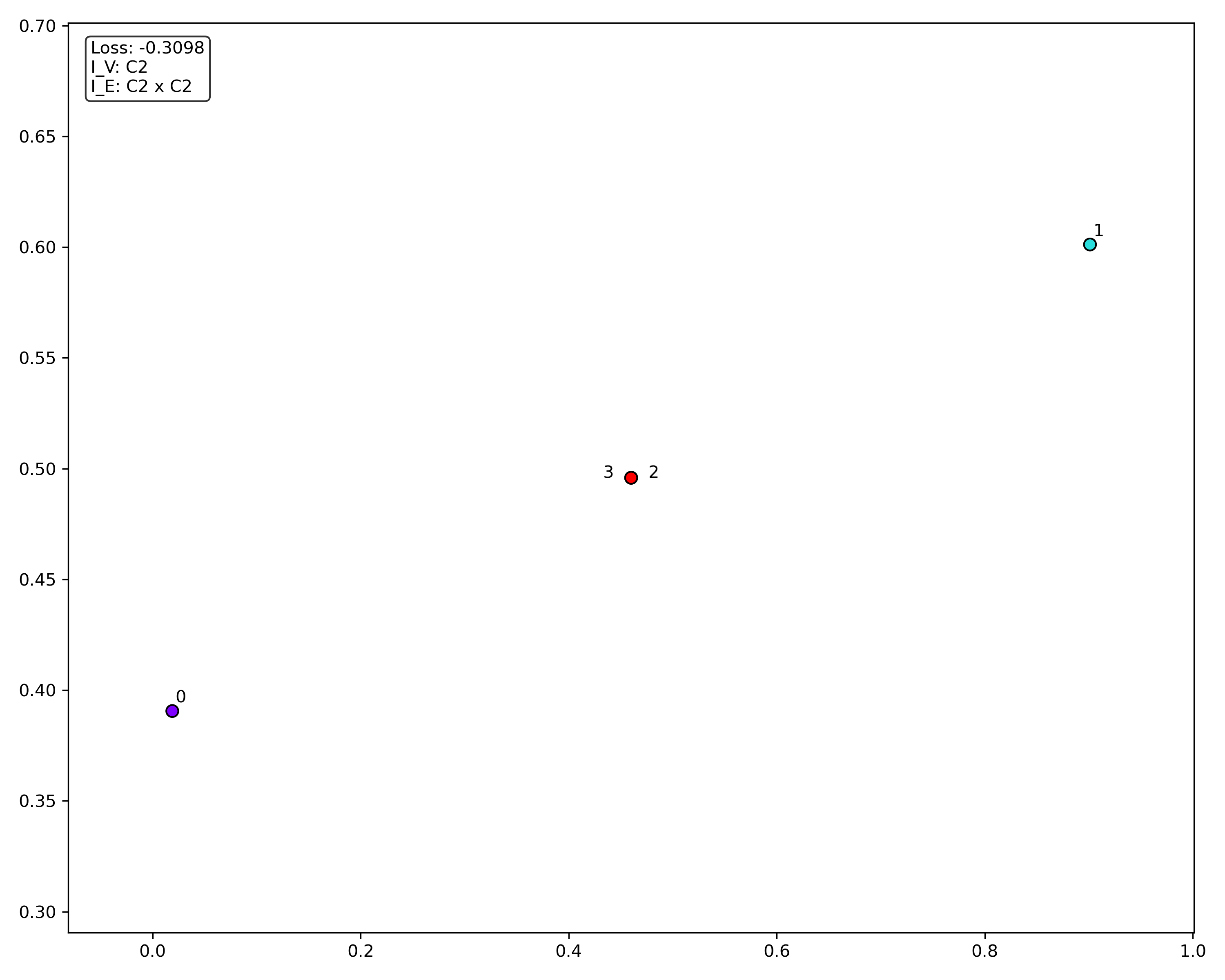}
\caption{}
\end{minipage}

\end{figure}

\subsection{Plots for 7 particles Experiments (kernel $d^{12} - d^8$)}
The plots are ordered in the order of the experiments in the table.\\

\begin{figure}[H]
\caption{Visualization of critical points from Table~\ref{table:experiment_results_ParticleAttraction_7}}
\label{fig:plots_particle_attarciton_n_7_reg_ker}% \centering

\centering
% Row 1
\begin{minipage}{0.45\textwidth}
\centering
\includegraphics[width=\linewidth]{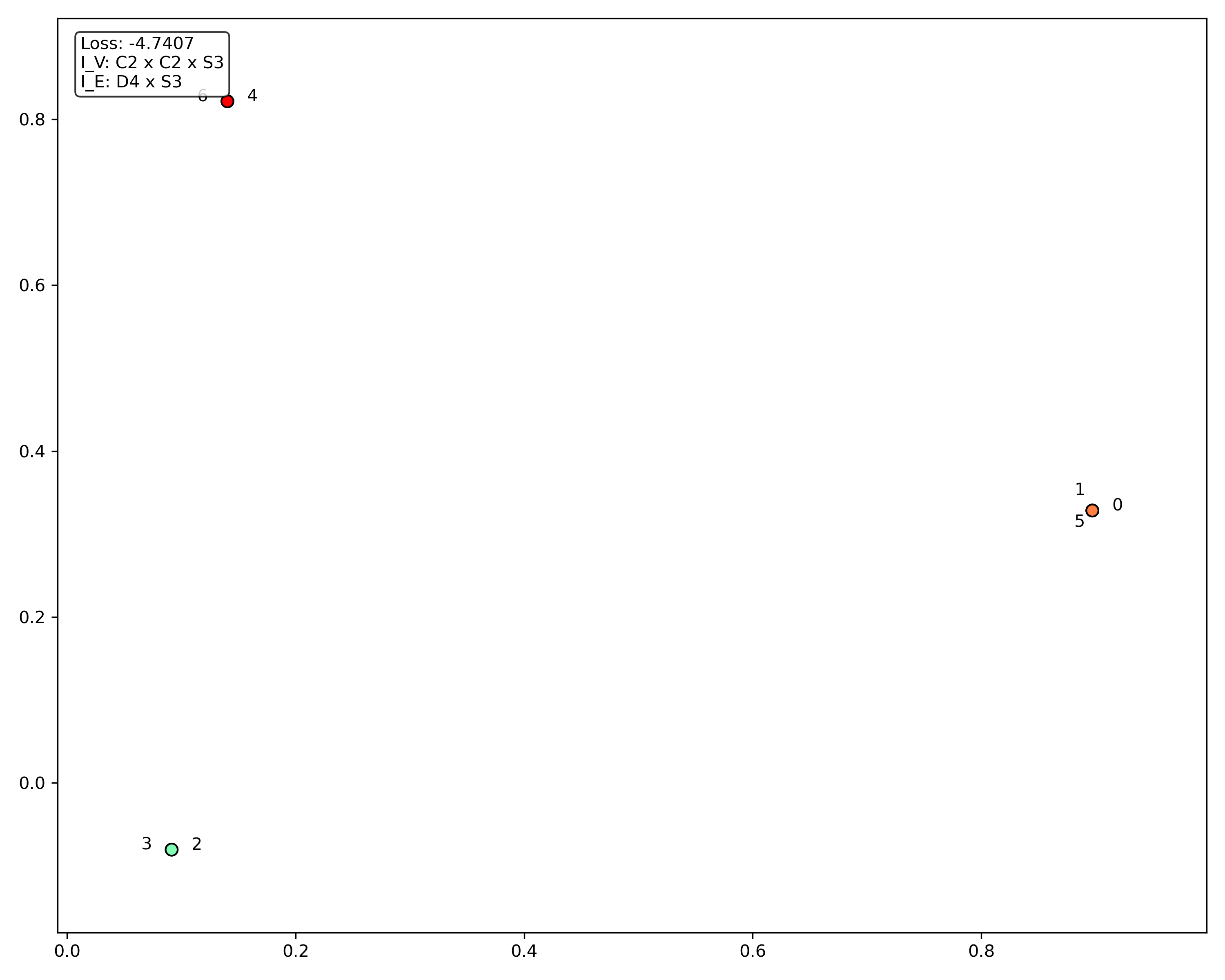}
\caption{}
\end{minipage}\hfill
\begin{minipage}{0.45\textwidth}
\centering
\includegraphics[width=\linewidth]{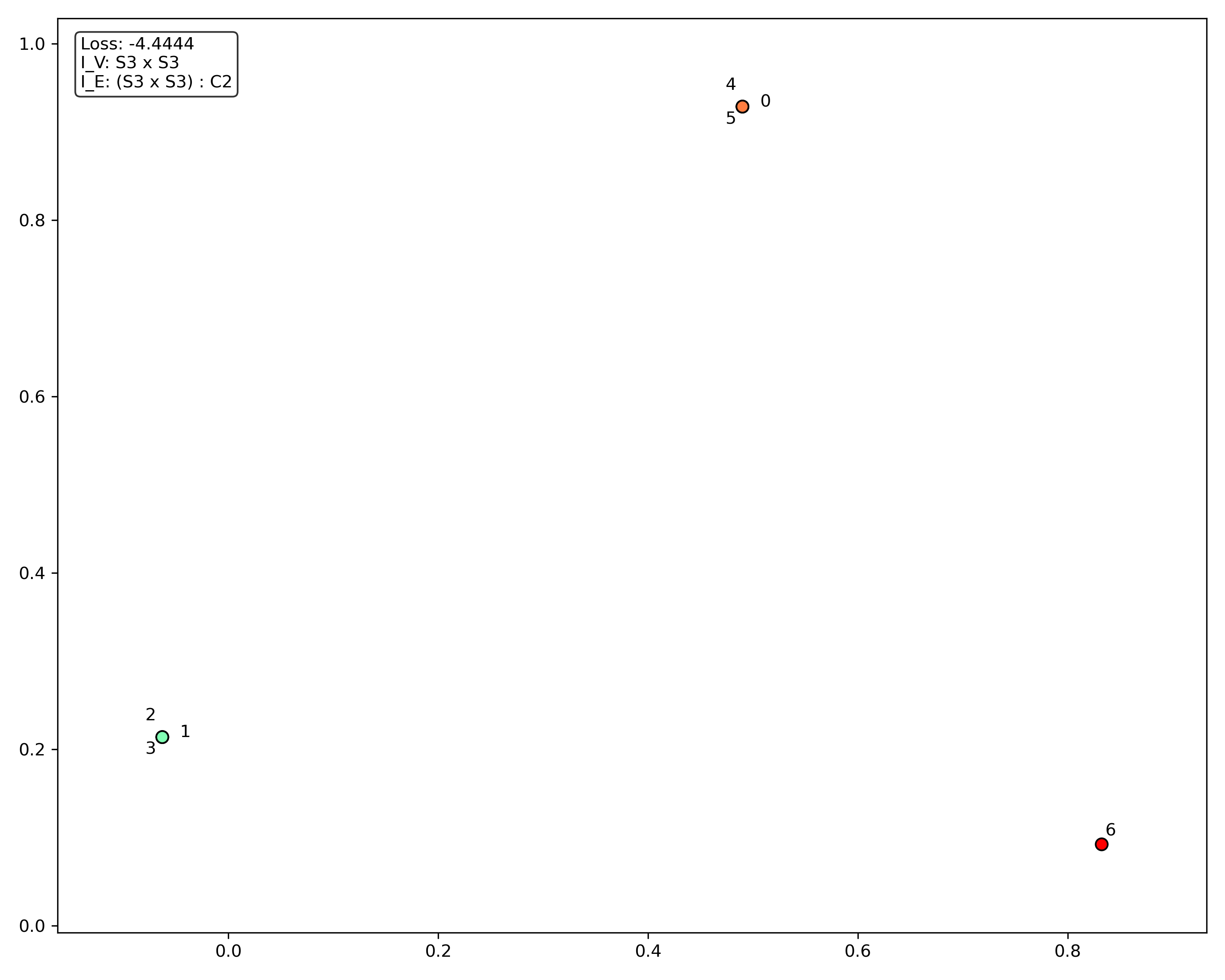}
\caption{}
\end{minipage}
\end{figure}
\begin{figure}[H]
% Row 2
\begin{minipage}{0.45\textwidth}
\centering
\includegraphics[width=\linewidth]{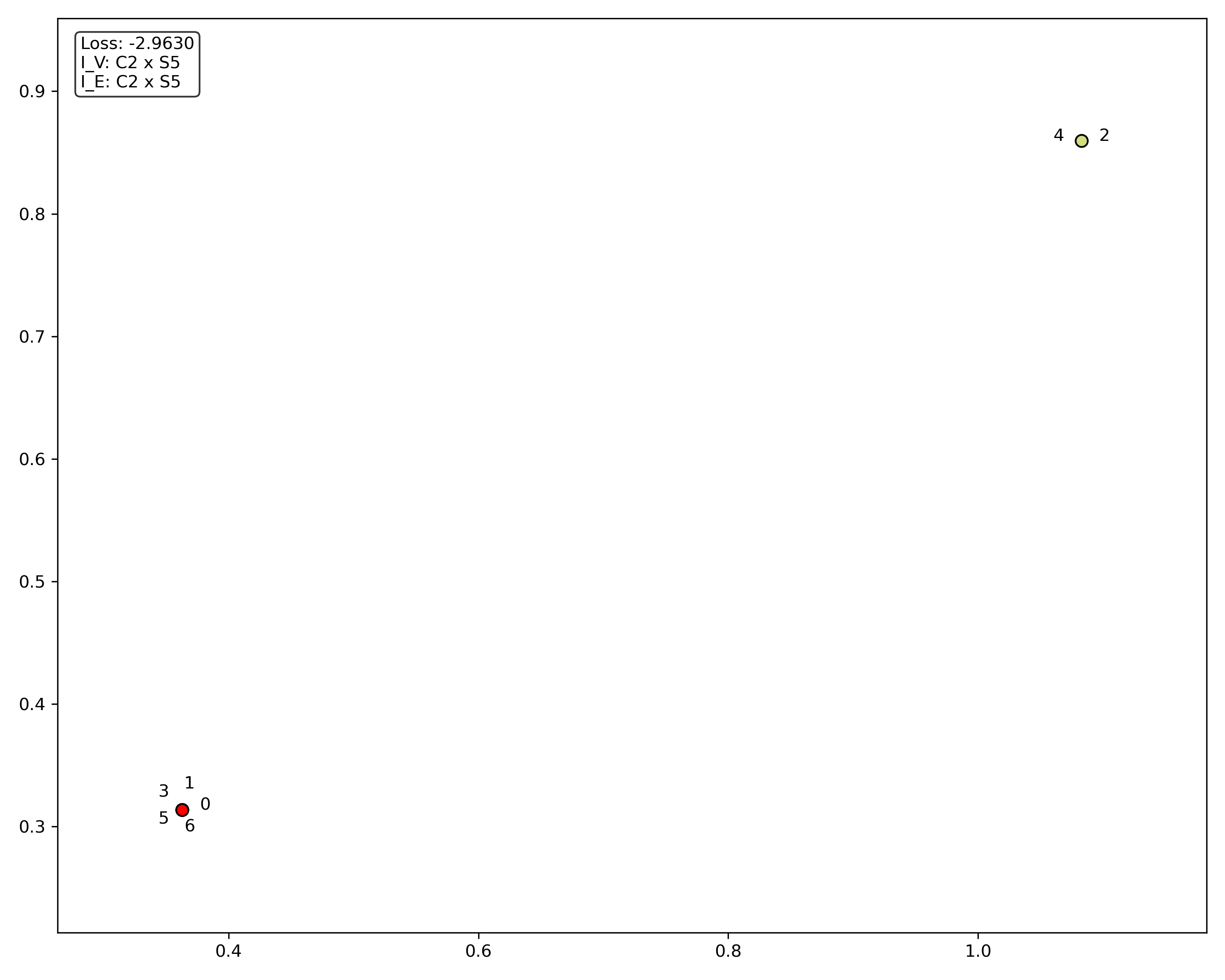}
\caption{}
\end{minipage}\hfill
\begin{minipage}{0.45\textwidth}
\centering
\includegraphics[width=\linewidth]{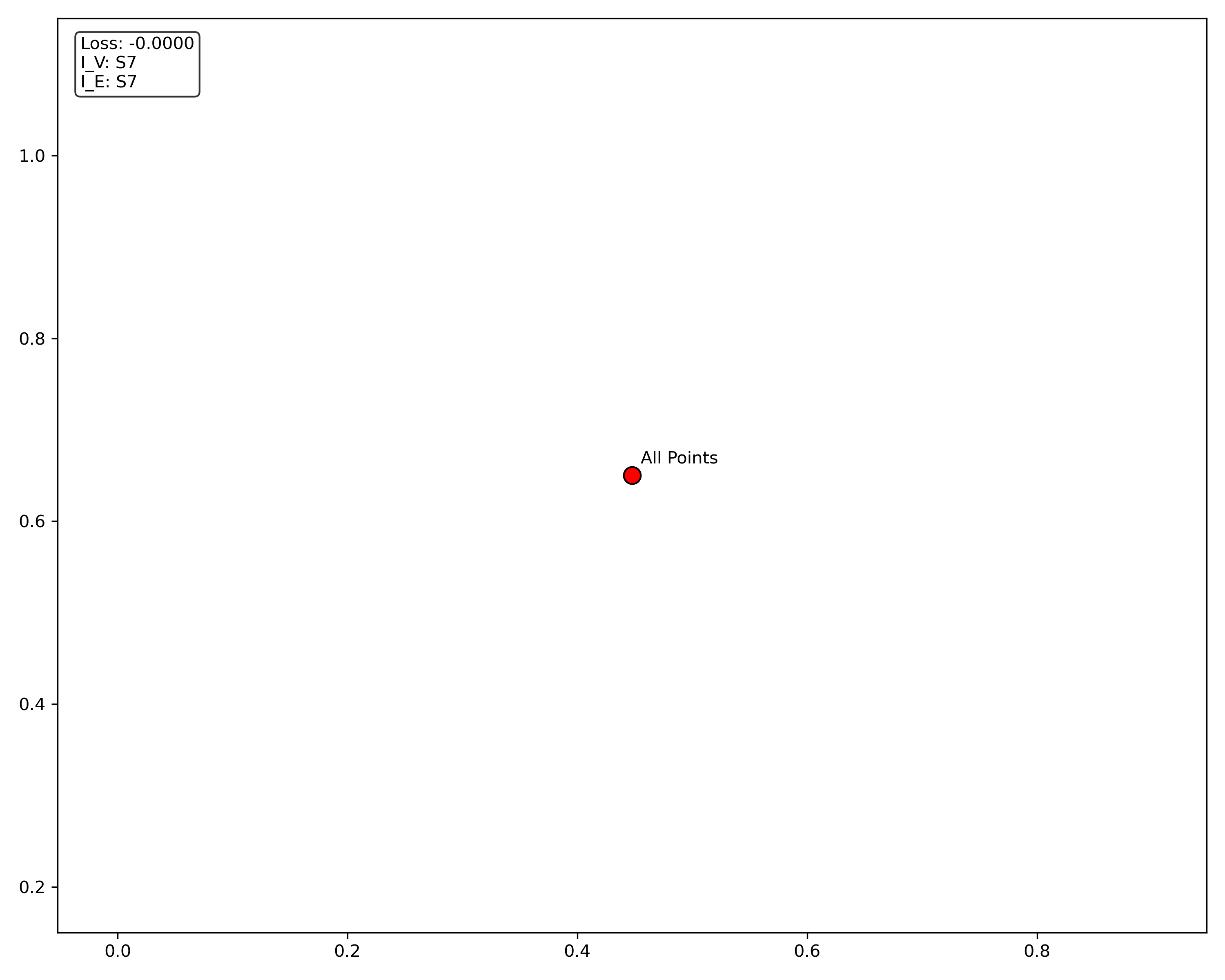}
\caption{}
\end{minipage}
\end{figure}
\begin{figure}[H]
% Row 3
\begin{minipage}{0.45\textwidth}
\centering
\includegraphics[width=\linewidth]{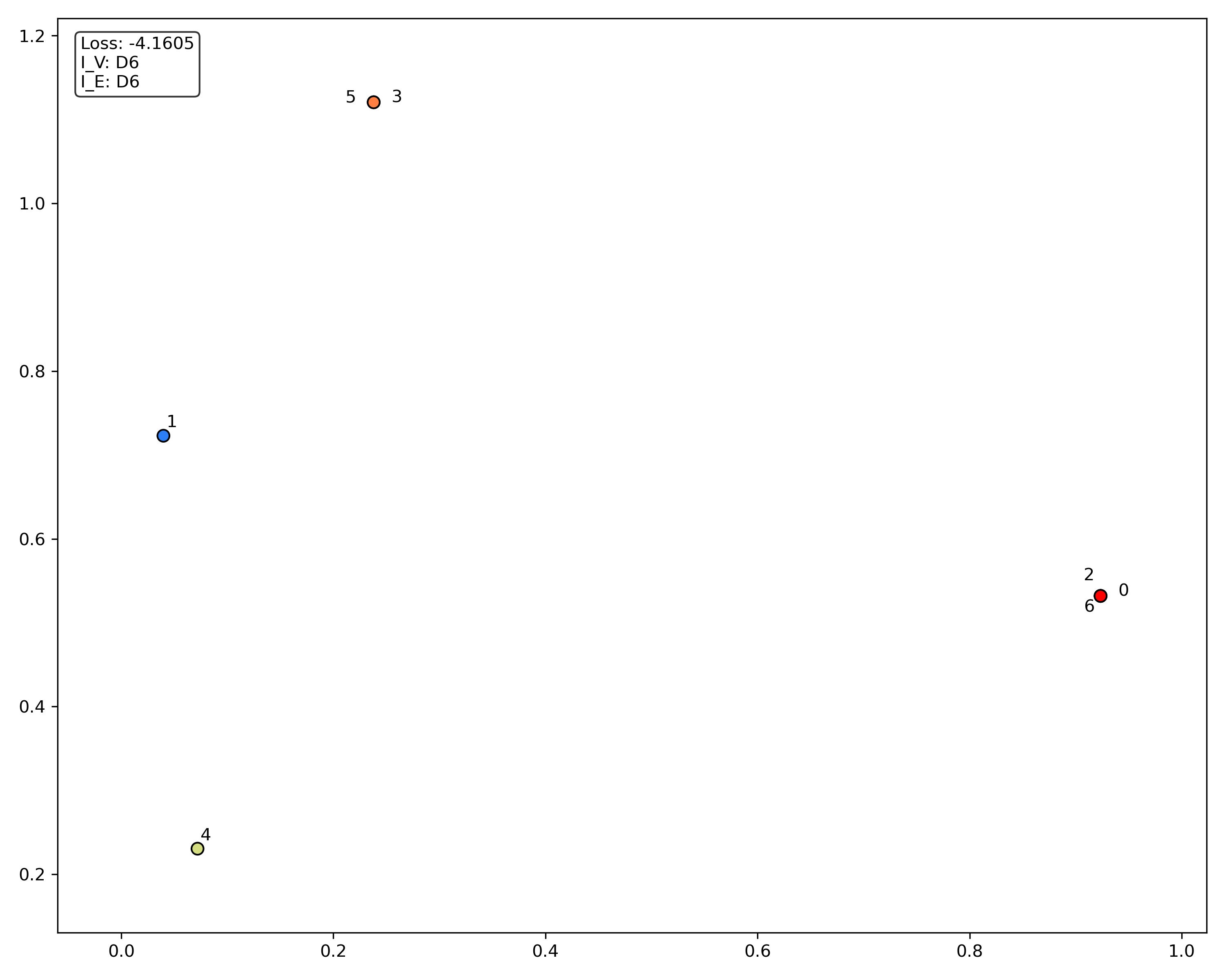}
\caption{}
\end{minipage}\hfill
\begin{minipage}{0.45\textwidth}
\centering
\includegraphics[width=\linewidth]{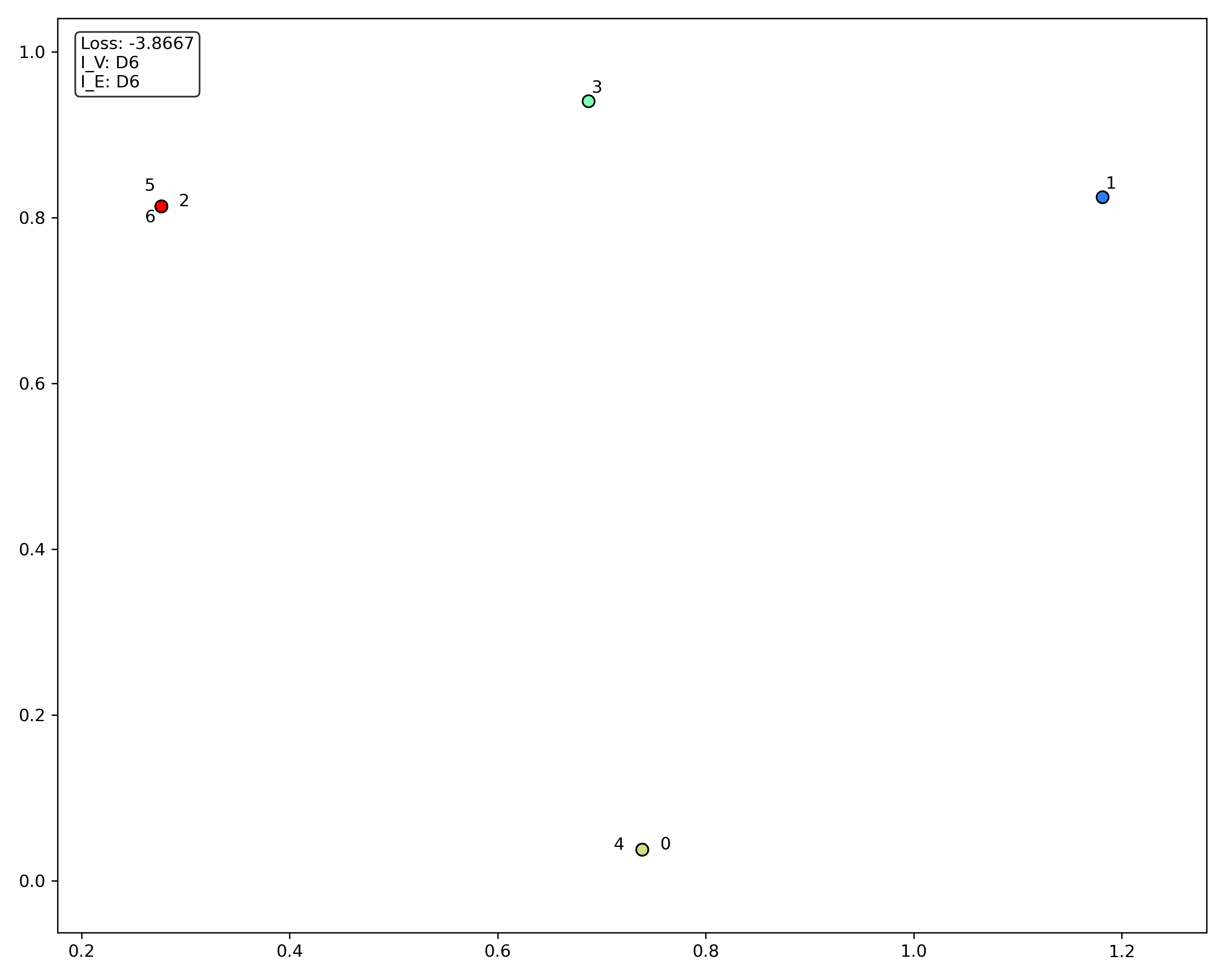}
\caption{}
\end{minipage}

\end{figure}
\begin{figure}[H]
\centering
% Row 1
\begin{minipage}{0.45\textwidth}
\centering
\includegraphics[width=\linewidth]{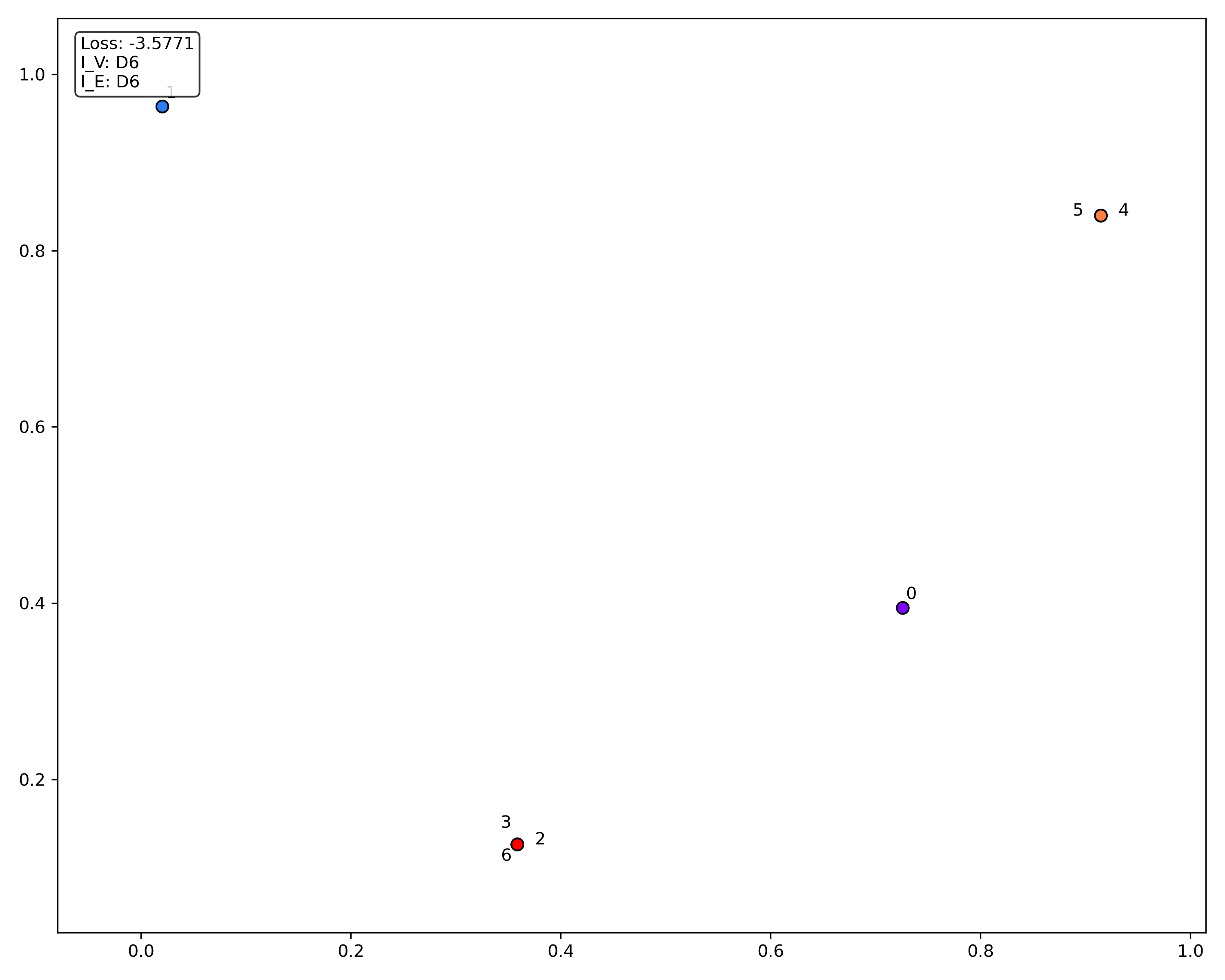}
\caption{}
\end{minipage}\hfill
\begin{minipage}{0.45\textwidth}
\centering
\includegraphics[width=\linewidth]{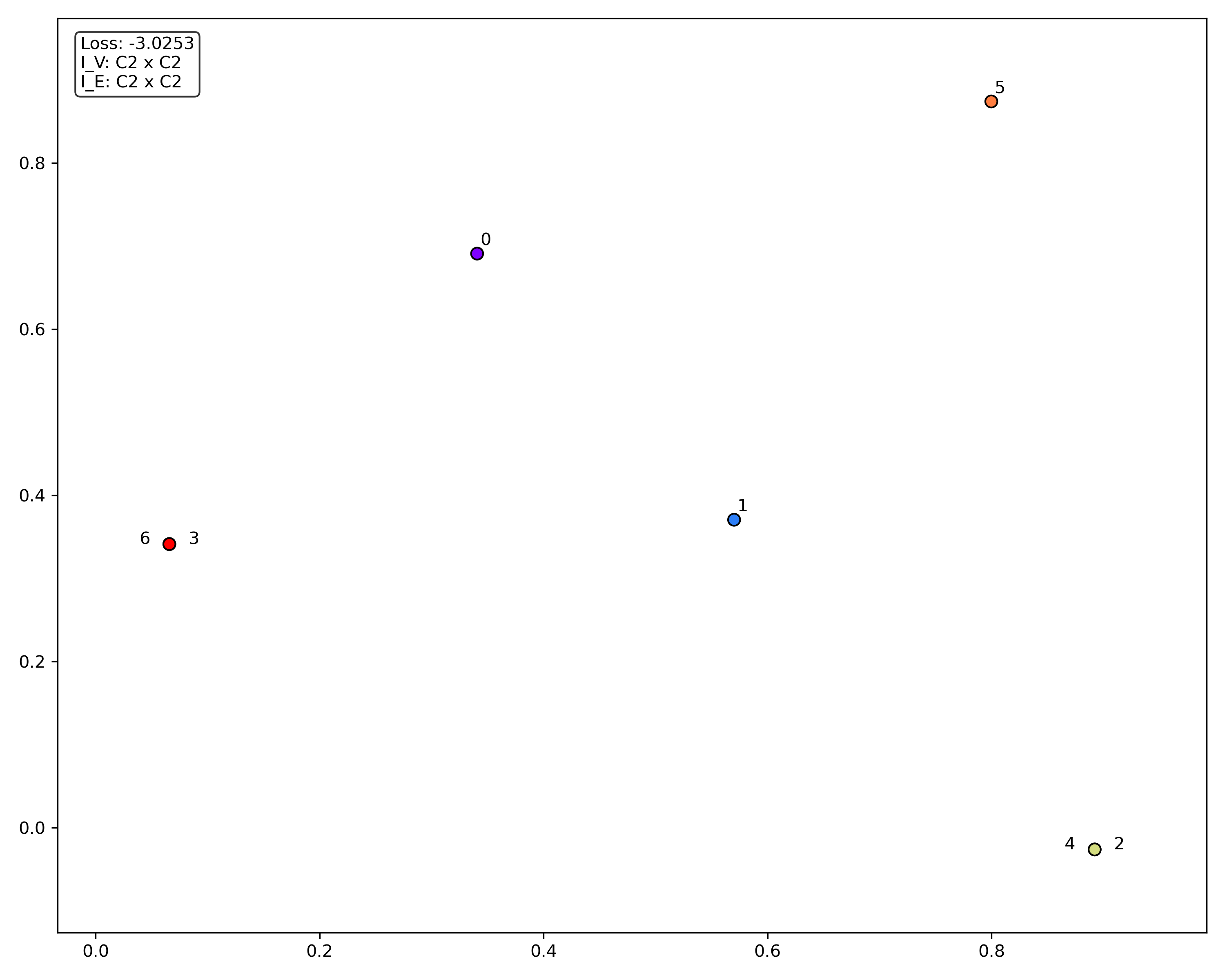}
\caption{}
\end{minipage}
\end{figure}
\begin{figure}[H]
% Row 2
\begin{minipage}{0.45\textwidth}
\centering
\includegraphics[width=\linewidth]{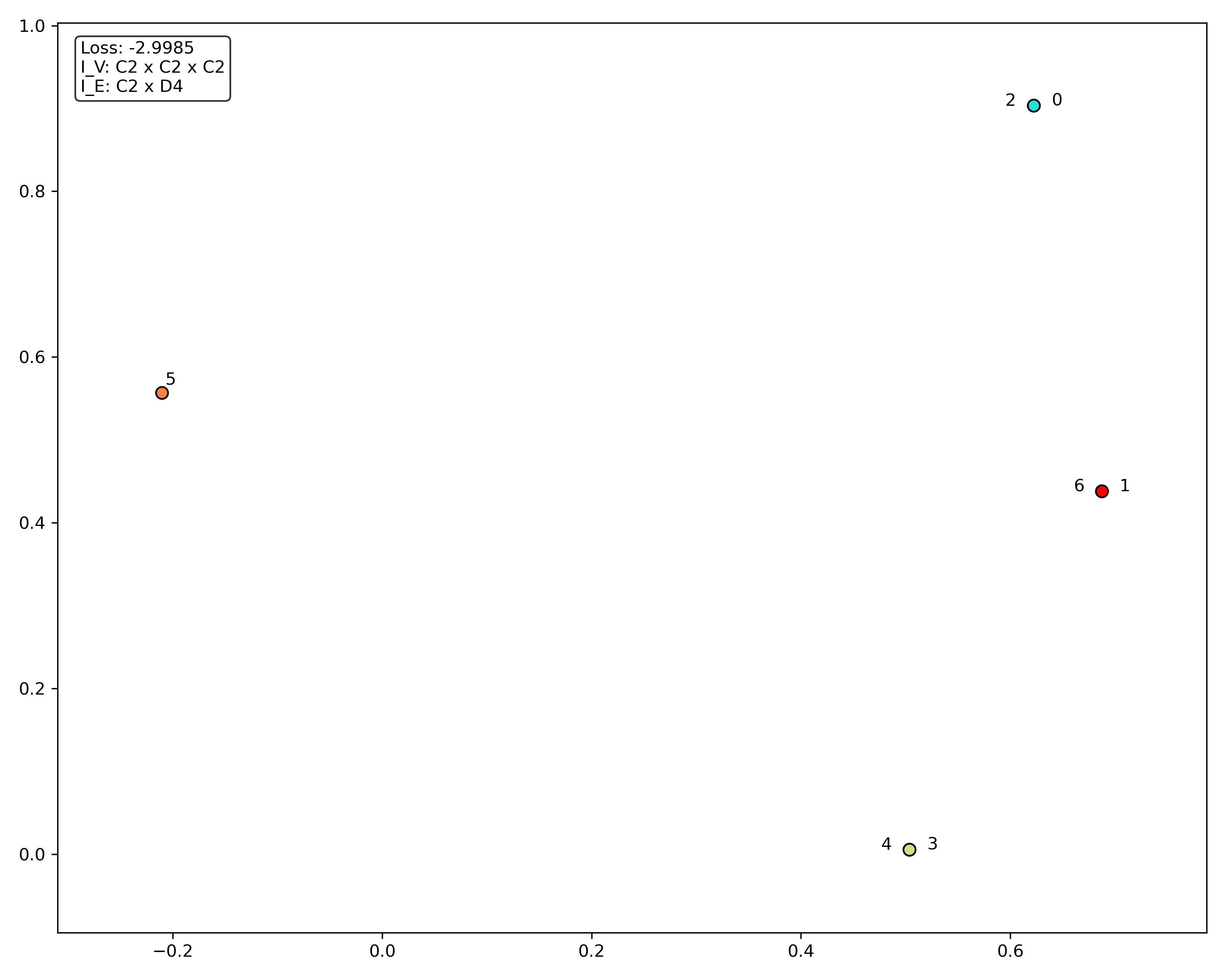}
\caption{}
\end{minipage}\hfill
\begin{minipage}{0.45\textwidth}
\centering
\includegraphics[width=\linewidth]{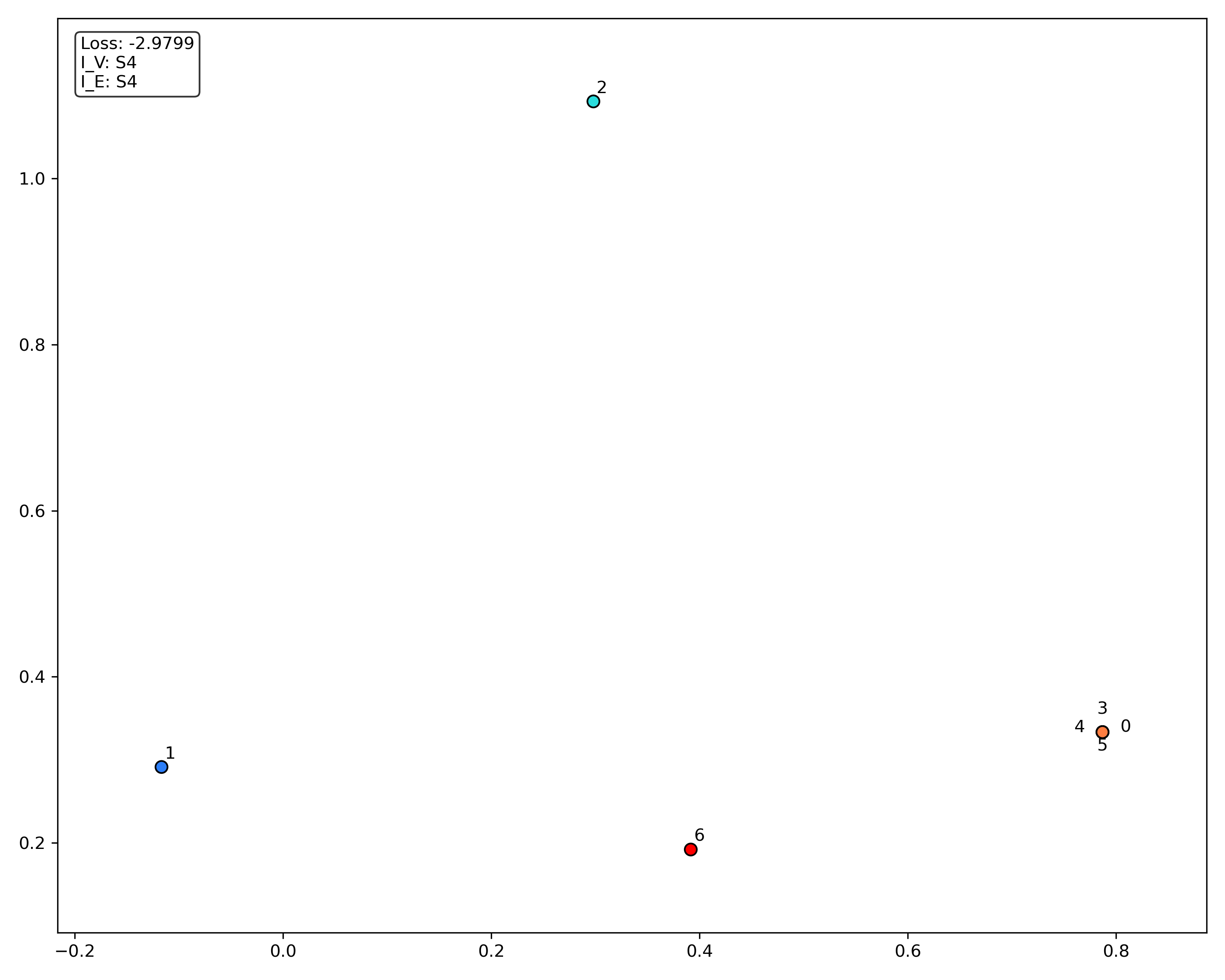}
\caption{}
\end{minipage}
\end{figure}
\begin{figure}[H]
% Row 3
\begin{minipage}{0.45\textwidth}
\centering
\includegraphics[width=\linewidth]{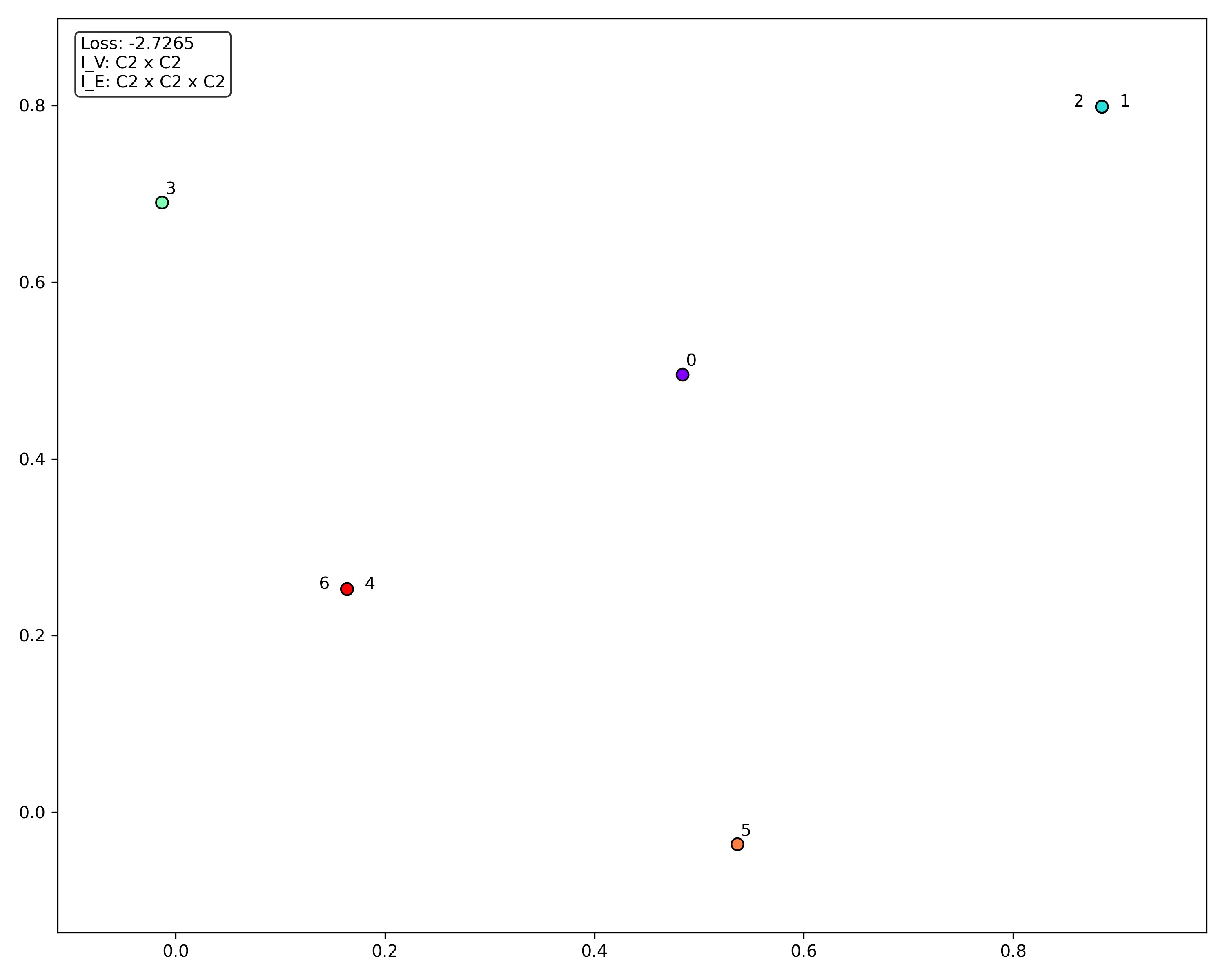}
\caption{}
\end{minipage}\hfill
\begin{minipage}{0.45\textwidth}
\centering
\includegraphics[width=\linewidth]{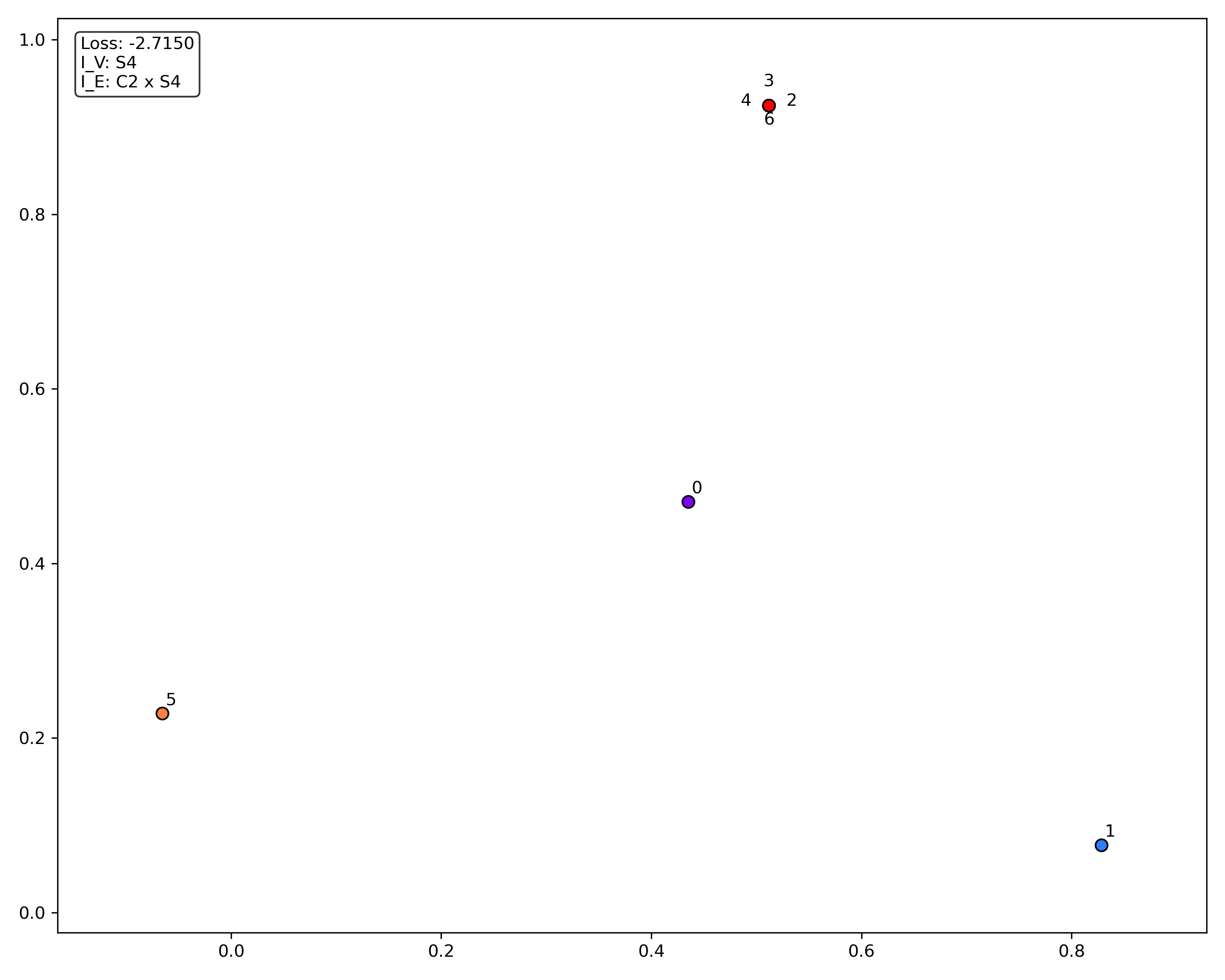}
\caption{}
\end{minipage}

\end{figure}
\begin{figure}[H]
\centering
% Row 1
\begin{minipage}{0.45\textwidth}
\centering
\includegraphics[width=\linewidth]{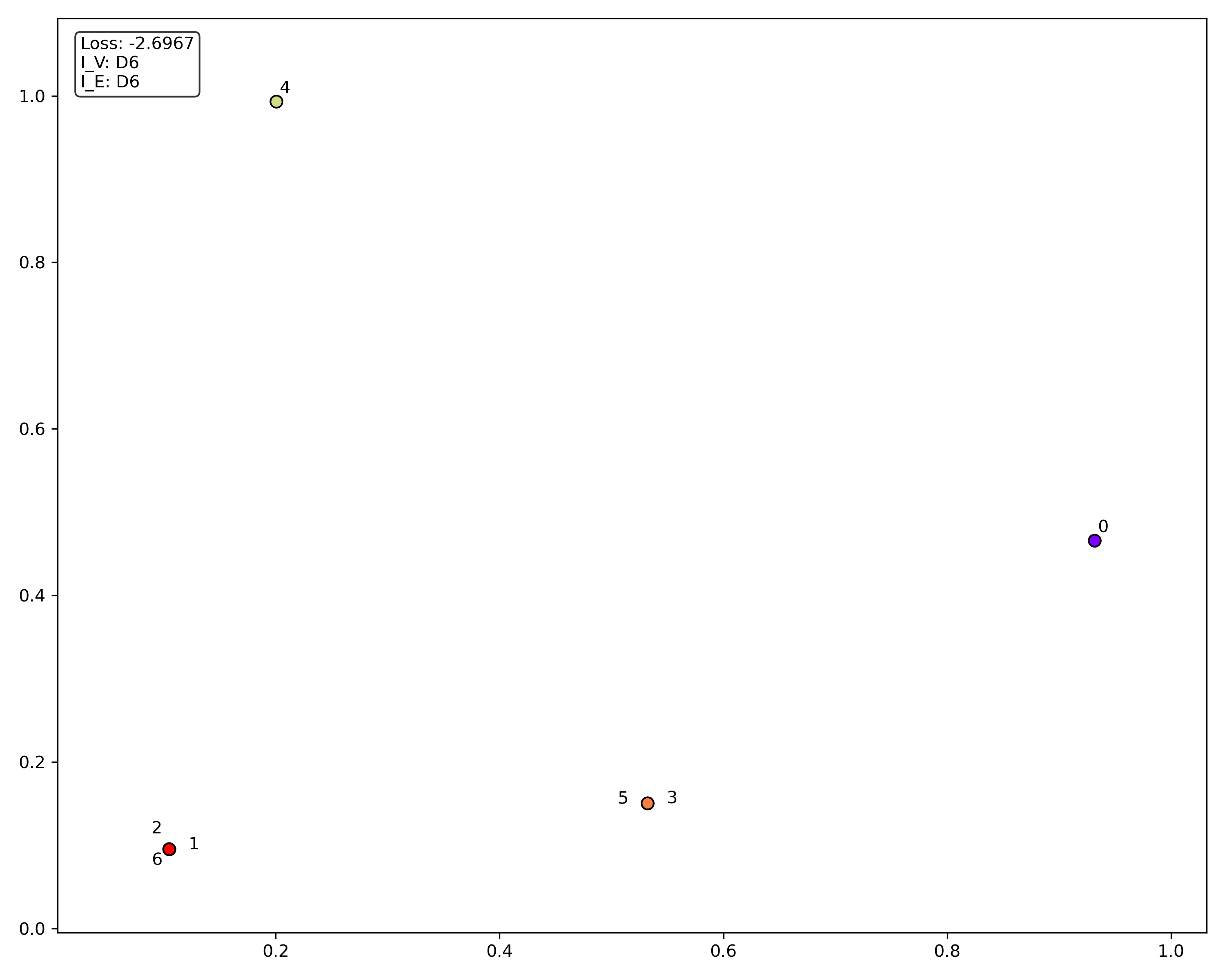}
\caption{}
\end{minipage}\hfill
\begin{minipage}{0.45\textwidth}
\centering
\includegraphics[width=\linewidth]{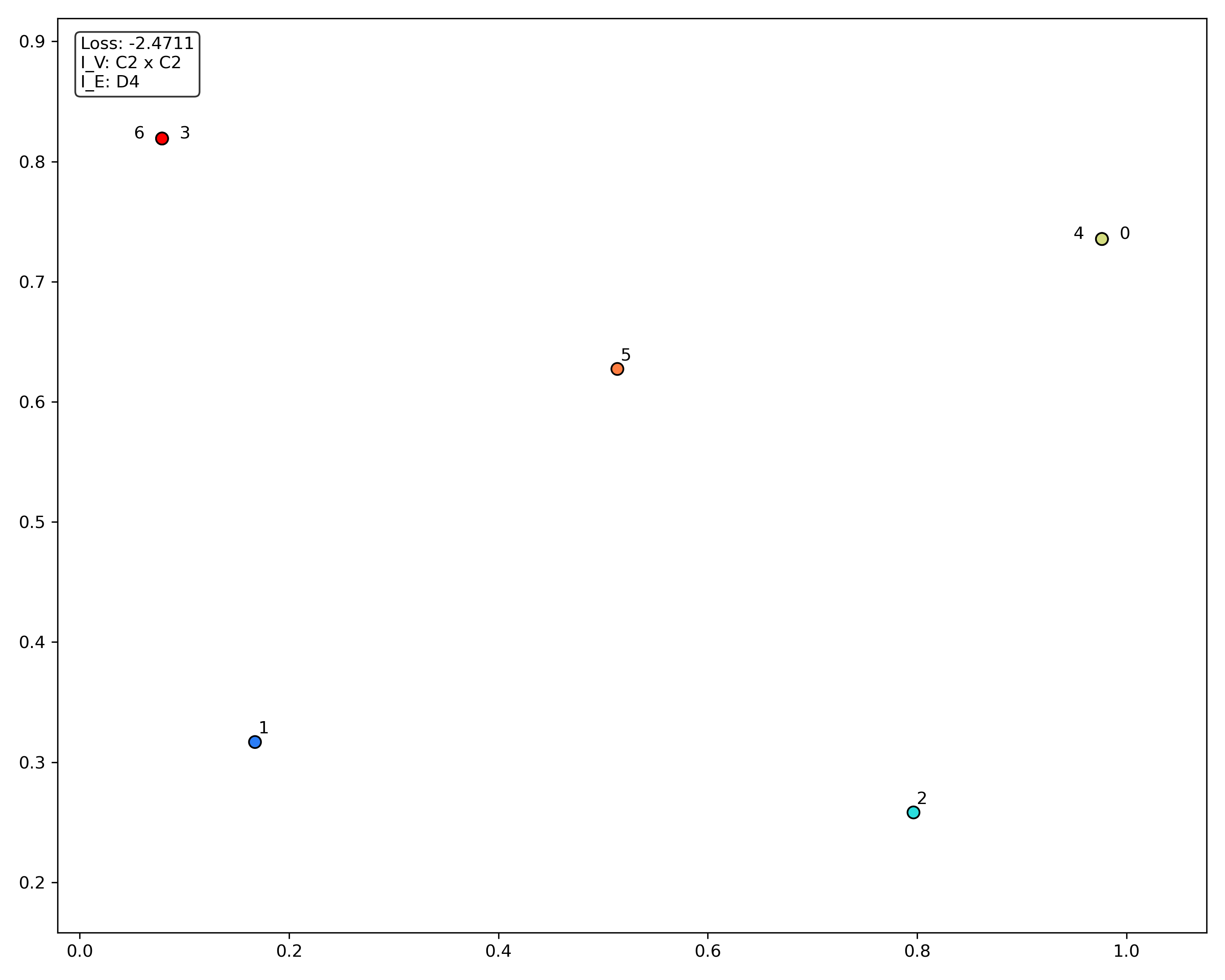}
\caption{}
\end{minipage}
\end{figure}
\begin{figure}[H]
% Row 2
\begin{minipage}{0.45\textwidth}
\centering
\includegraphics[width=\linewidth]{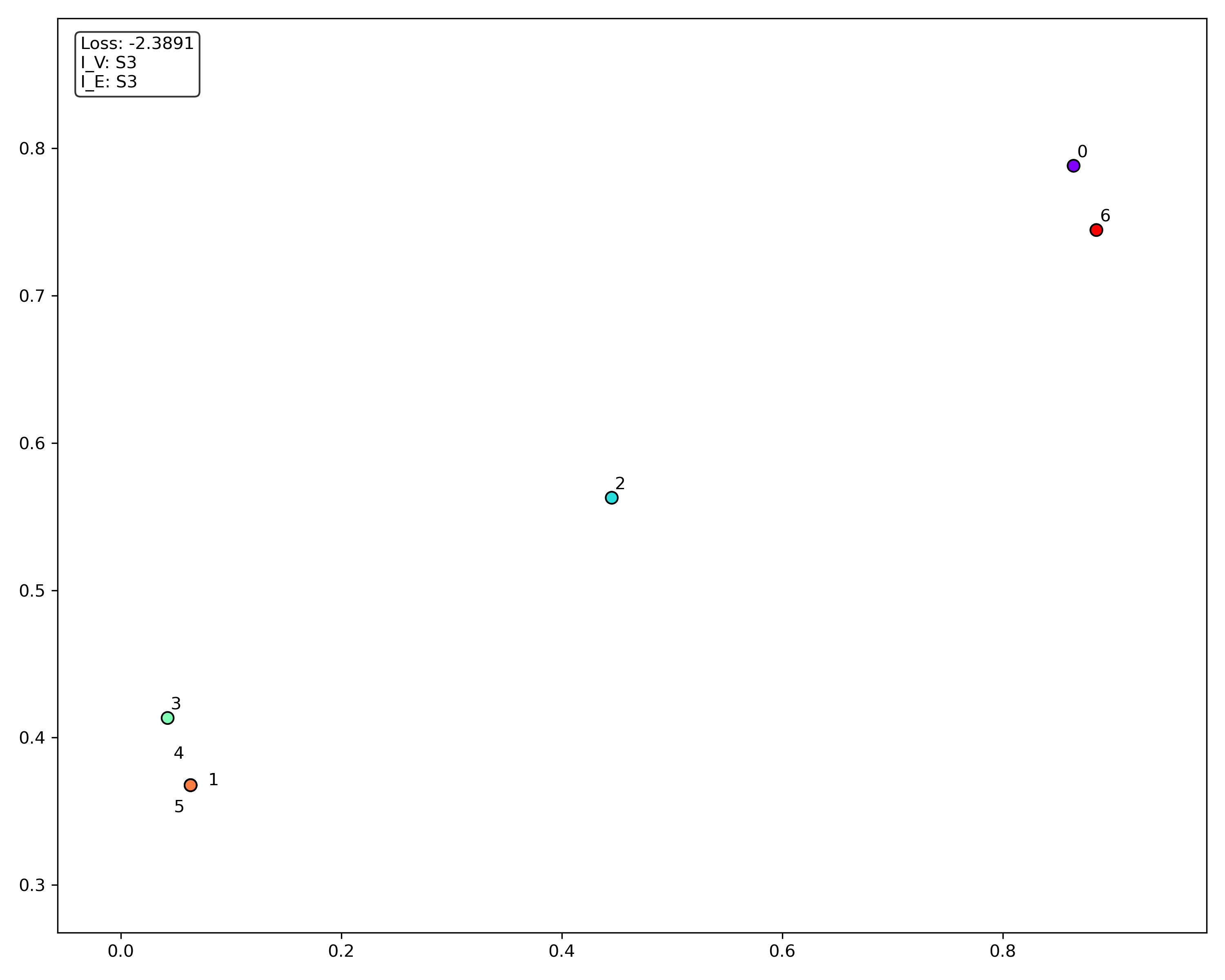}
\caption{}
\end{minipage}\hfill
\begin{minipage}{0.45\textwidth}
\centering
\includegraphics[width=\linewidth]{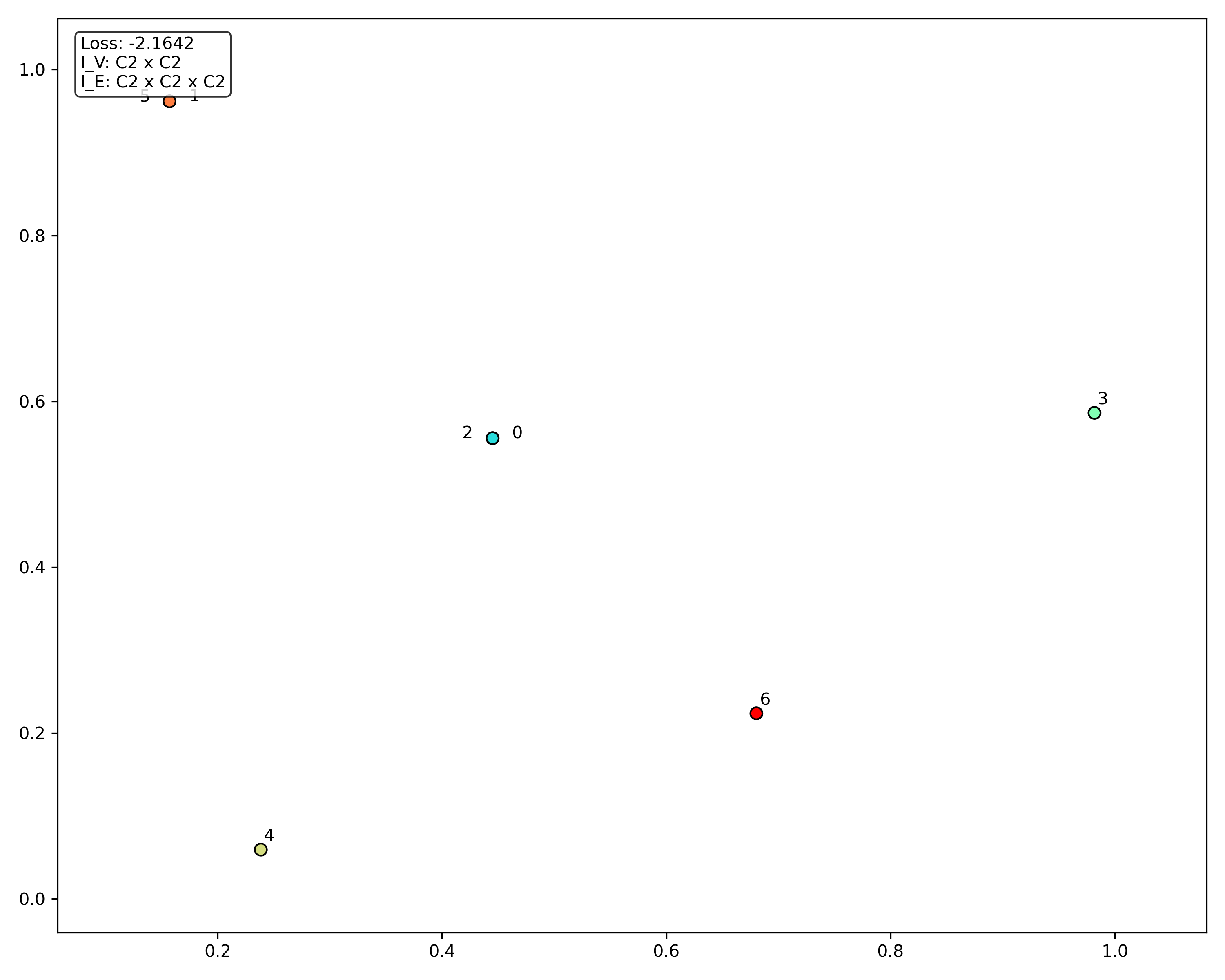}
\caption{}
\end{minipage}
\end{figure}
\begin{figure}[H]
% Row 3
\begin{minipage}{0.45\textwidth}
\centering
\includegraphics[width=\linewidth]{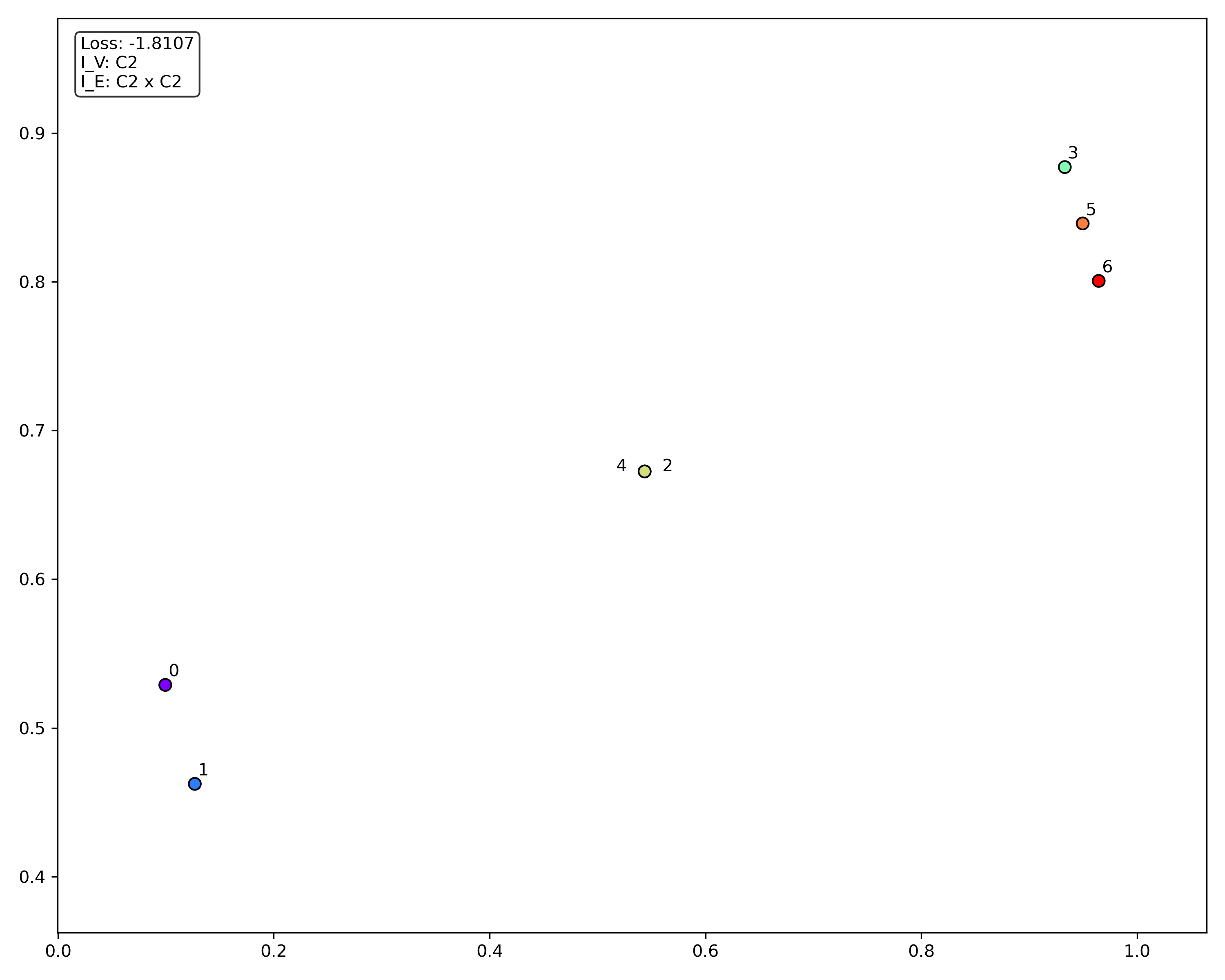}
\caption{}
\end{minipage}\hfill
\begin{minipage}{0.45\textwidth}
\centering
\includegraphics[width=\linewidth]{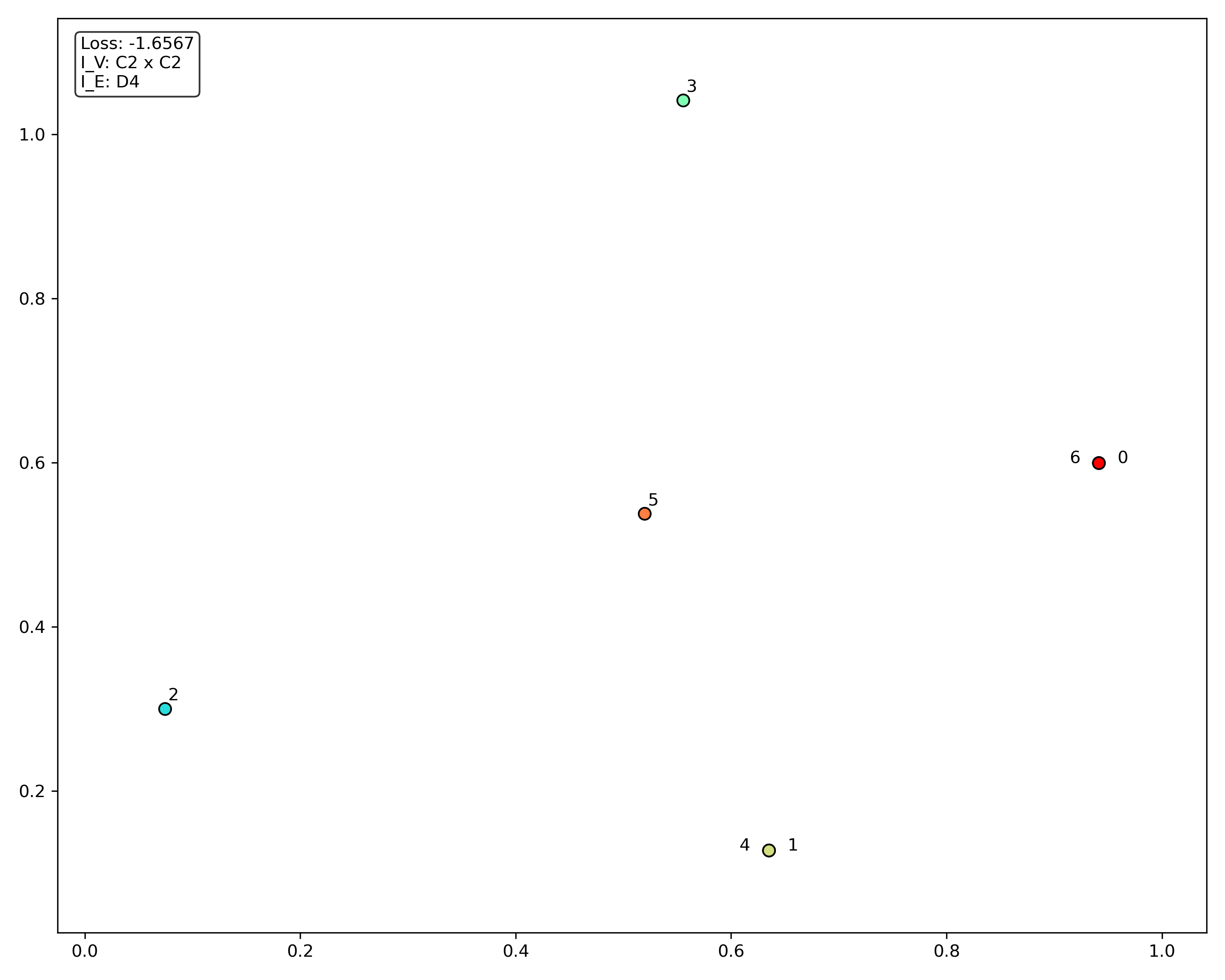}
\caption{}
\end{minipage}

\end{figure}
\begin{figure}[H]
\centering
% Row 1
\begin{minipage}{0.45\textwidth}
\centering
\includegraphics[width=\linewidth]{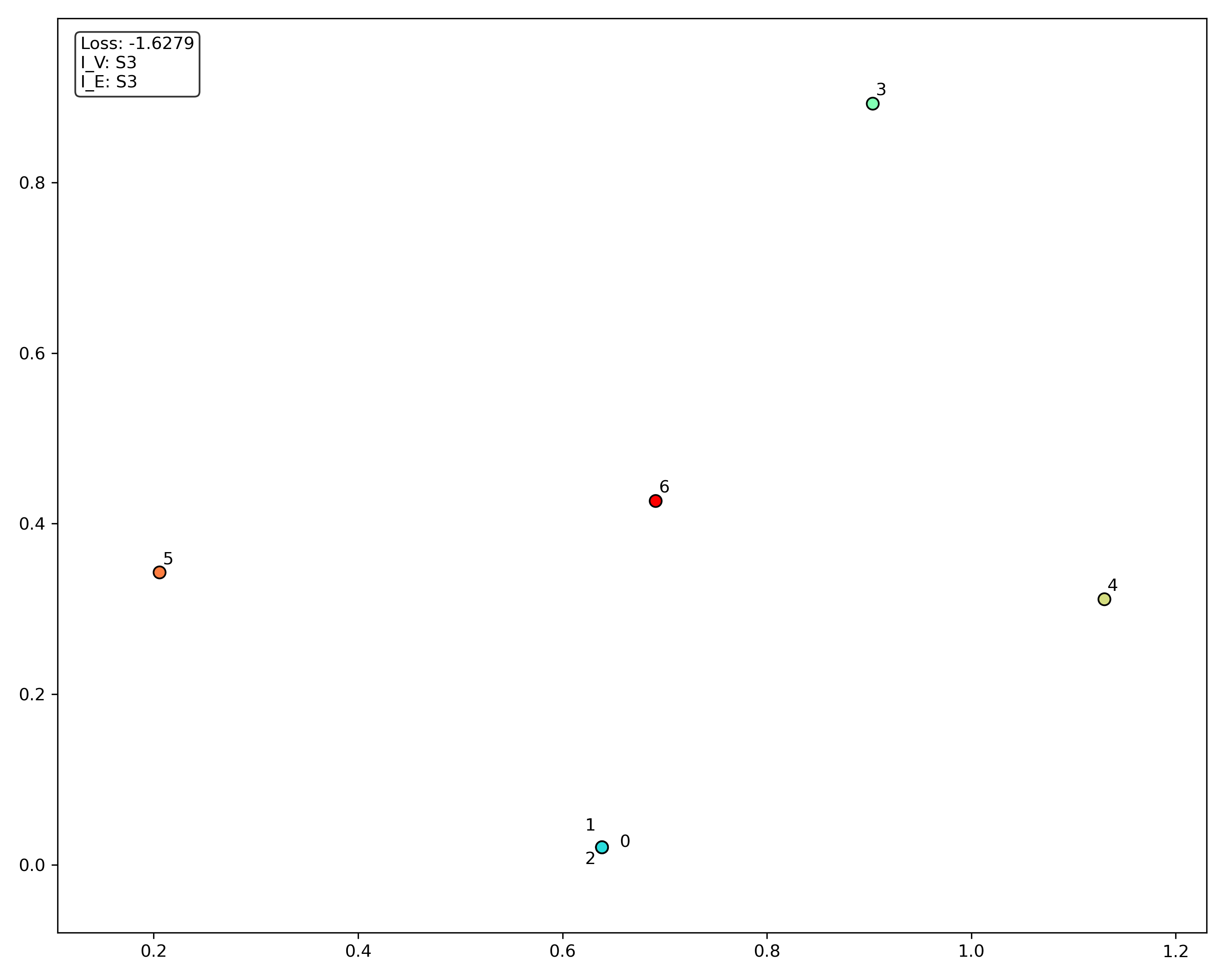}
\caption{}
\end{minipage}\hfill
\begin{minipage}{0.45\textwidth}
\centering
\includegraphics[width=\linewidth]{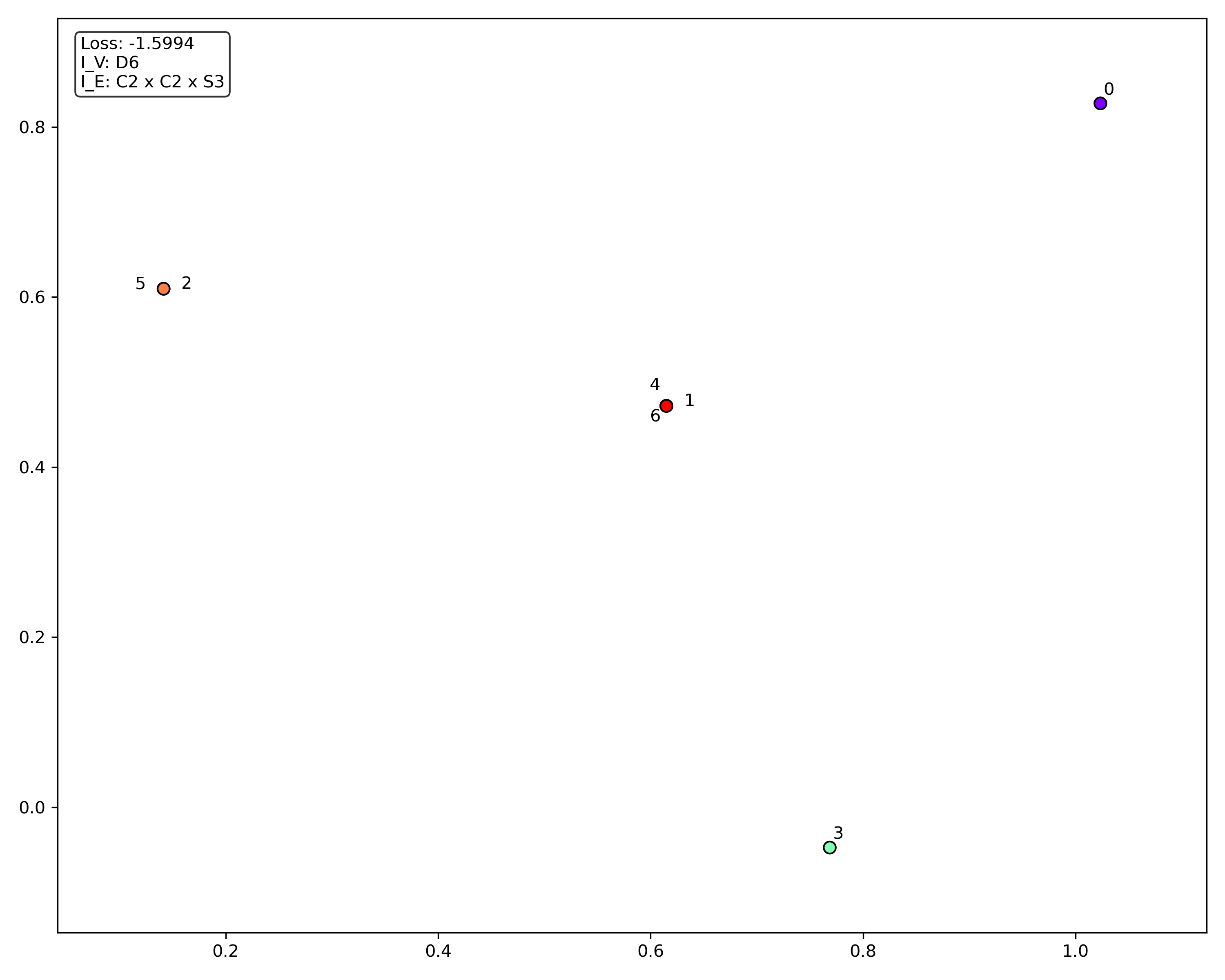}
\caption{}
\end{minipage}
\end{figure}
\begin{figure}[H]
\centering
% Row 2
\begin{minipage}{0.45\textwidth}
\centering
\includegraphics[width=\linewidth]{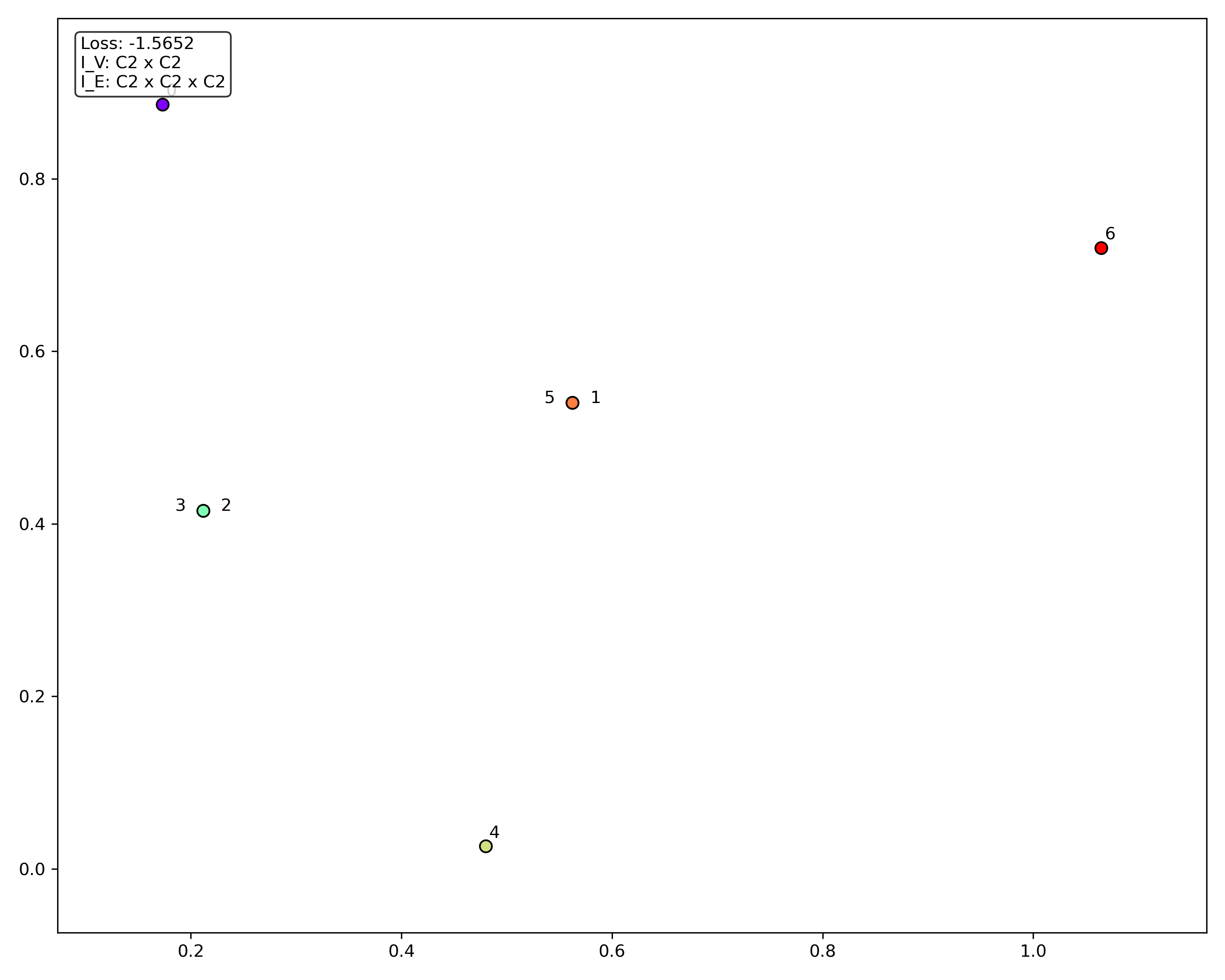}
\caption{}
\end{minipage}\hfill
\begin{minipage}{0.45\textwidth}
\centering
\includegraphics[width=\linewidth]{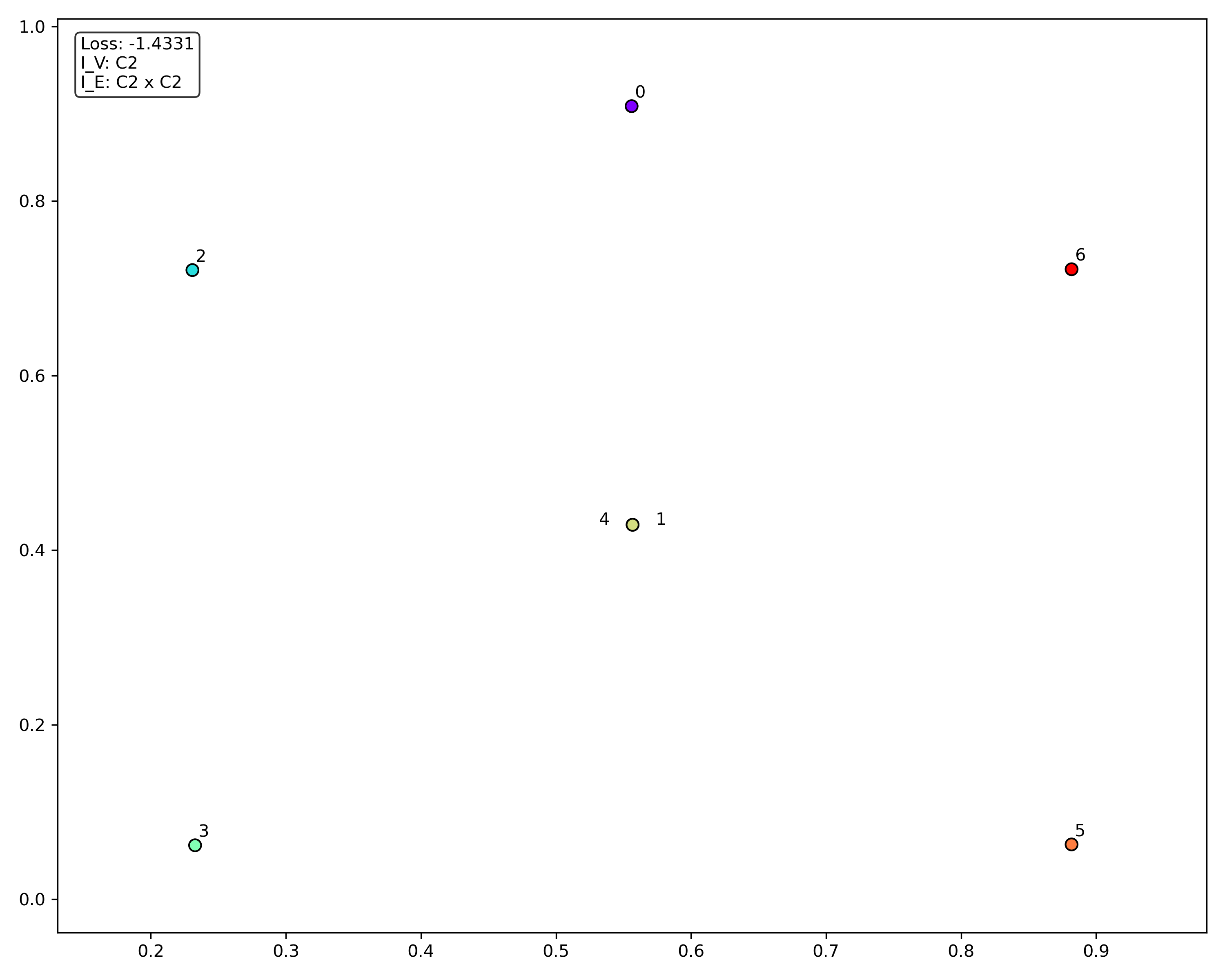}
\caption{}
\end{minipage}
\end{figure}
\begin{figure}[H]
\centering
% Row 3
\begin{minipage}{0.45\textwidth}
\centering
\includegraphics[width=\linewidth]{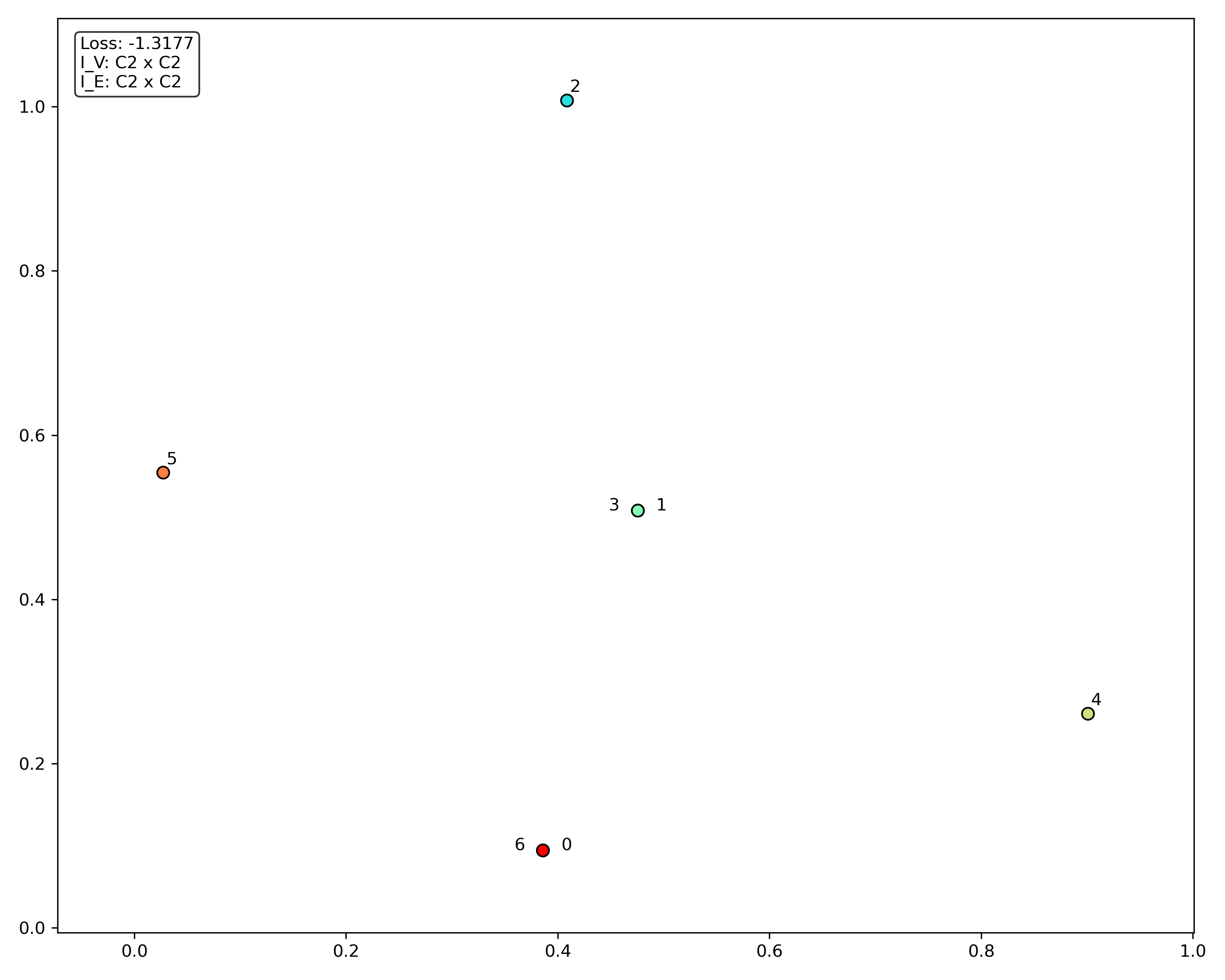}
\caption{}
\end{minipage}\hfill
\begin{minipage}{0.45\textwidth}
\centering
\includegraphics[width=\linewidth]{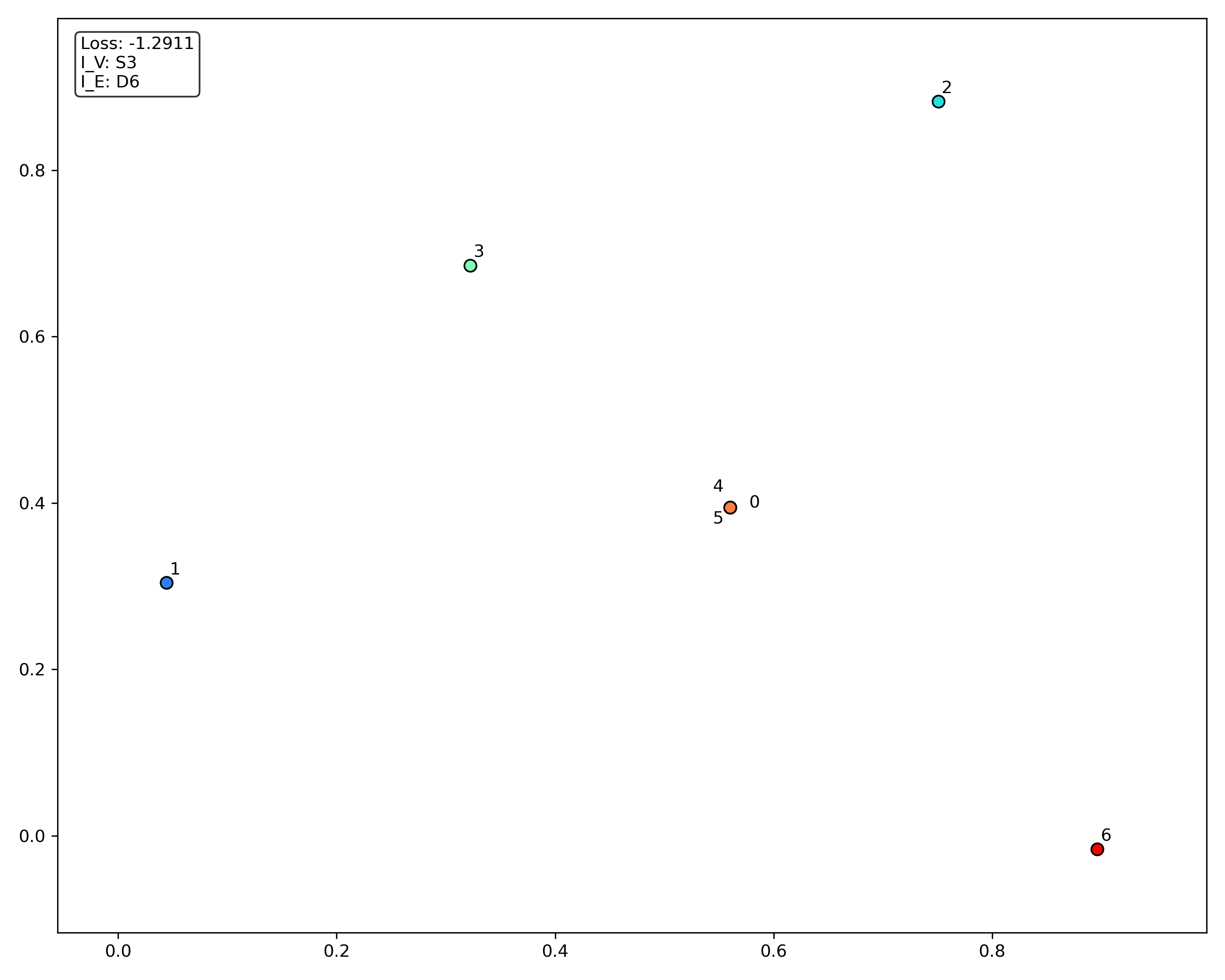}
\caption{}
\end{minipage}

\end{figure}
\begin{figure}[H]
\centering
% Row 1
\begin{minipage}{0.45\textwidth}
\centering
\includegraphics[width=\linewidth]{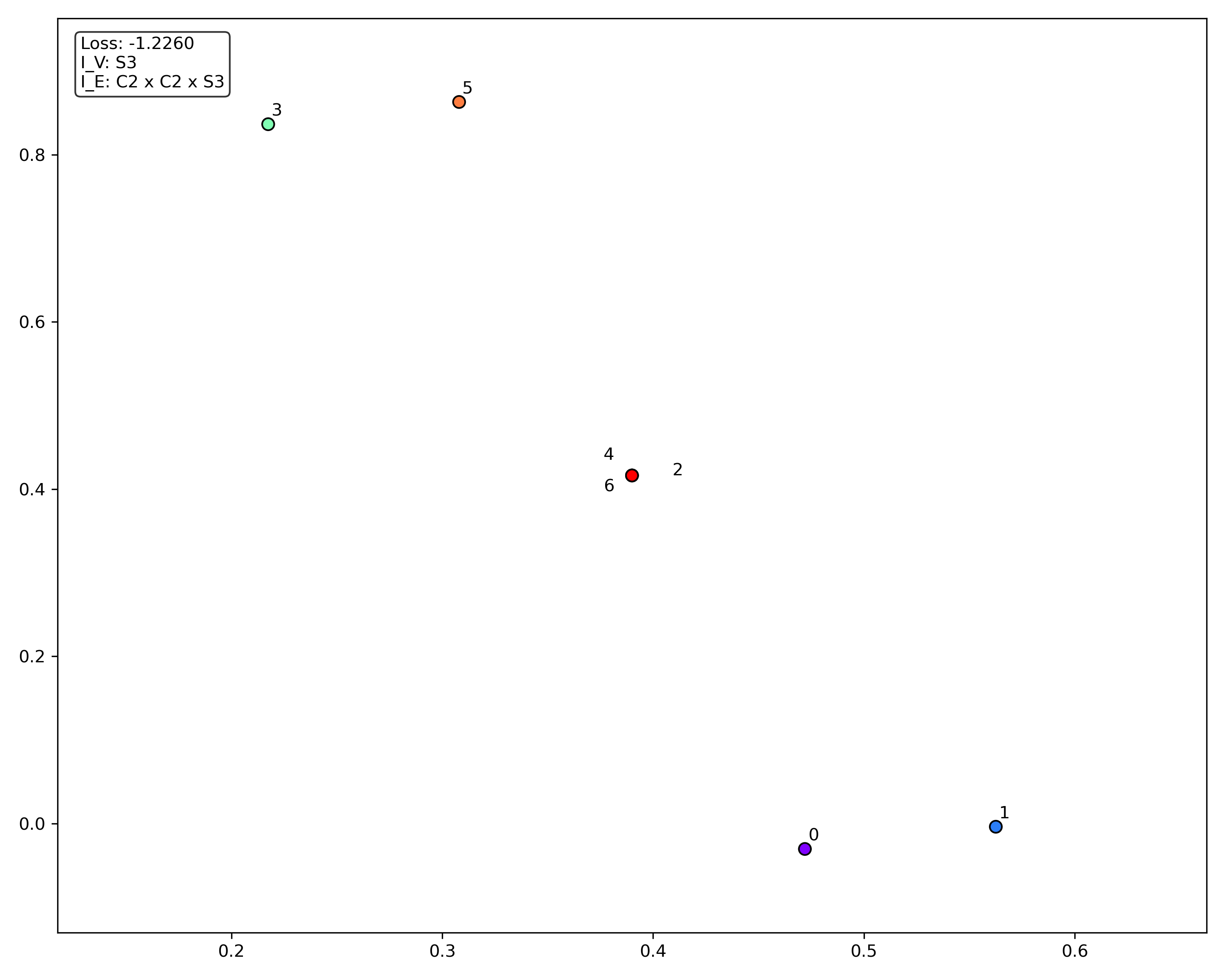}
\caption{}
\end{minipage}\hfill
\begin{minipage}{0.45\textwidth}
\centering
\includegraphics[width=\linewidth]{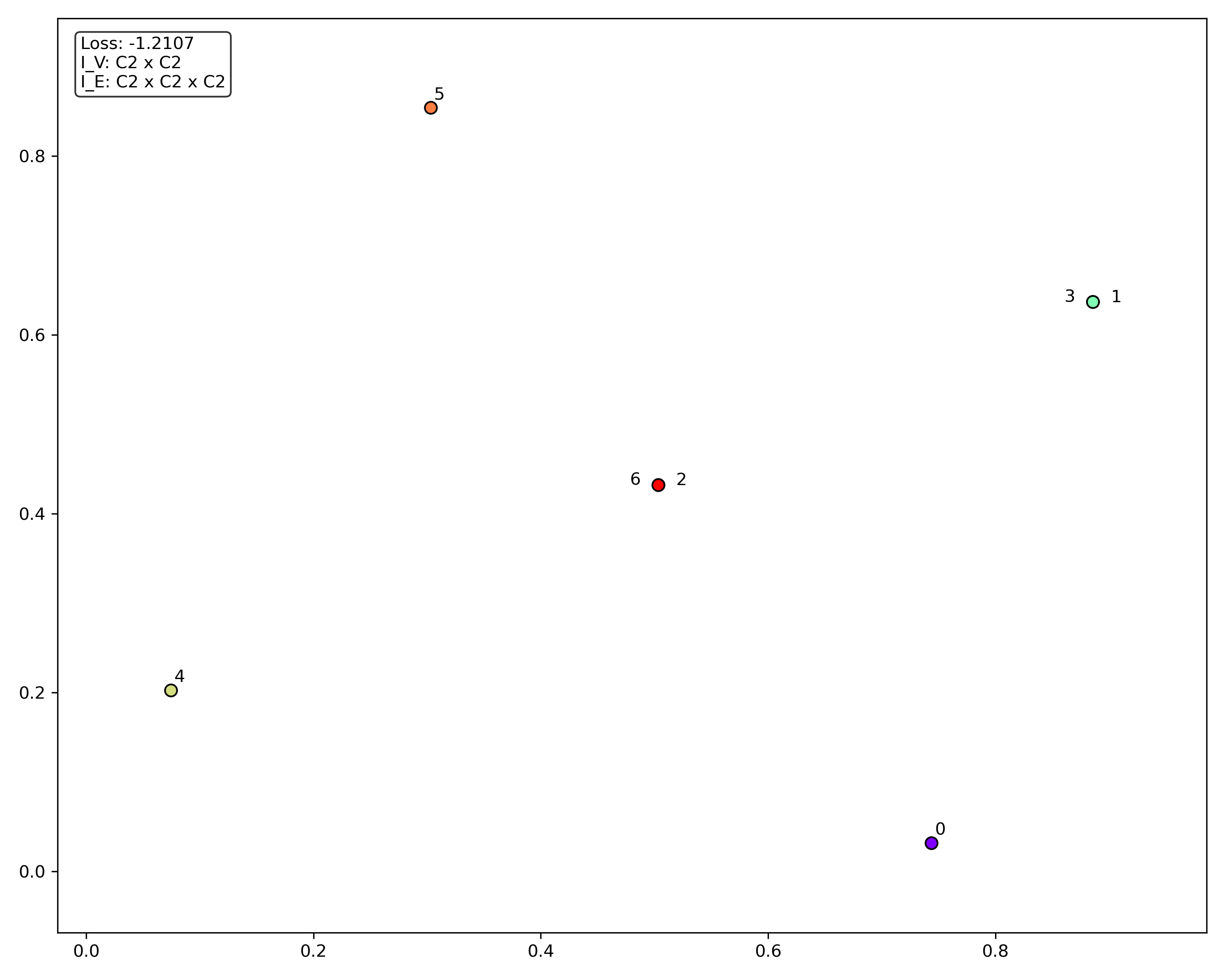}
\caption{}
\end{minipage}
\end{figure}
\begin{figure}[H]
\centering
% Row 2
\begin{minipage}{0.45\textwidth}
\centering
\includegraphics[width=\linewidth]{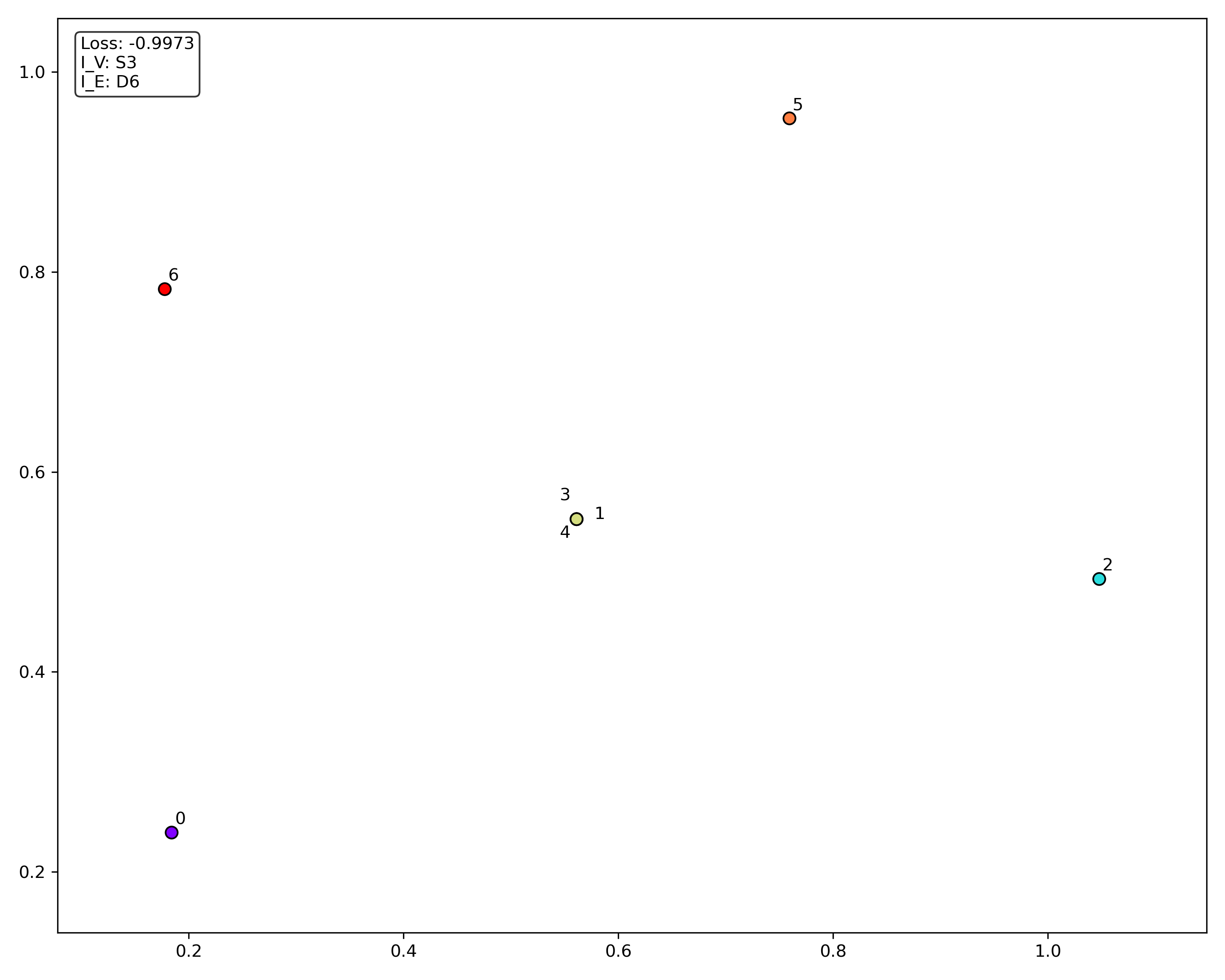}
\caption{}
\end{minipage}\hfill
\begin{minipage}{0.45\textwidth}
\centering
\includegraphics[width=\linewidth]{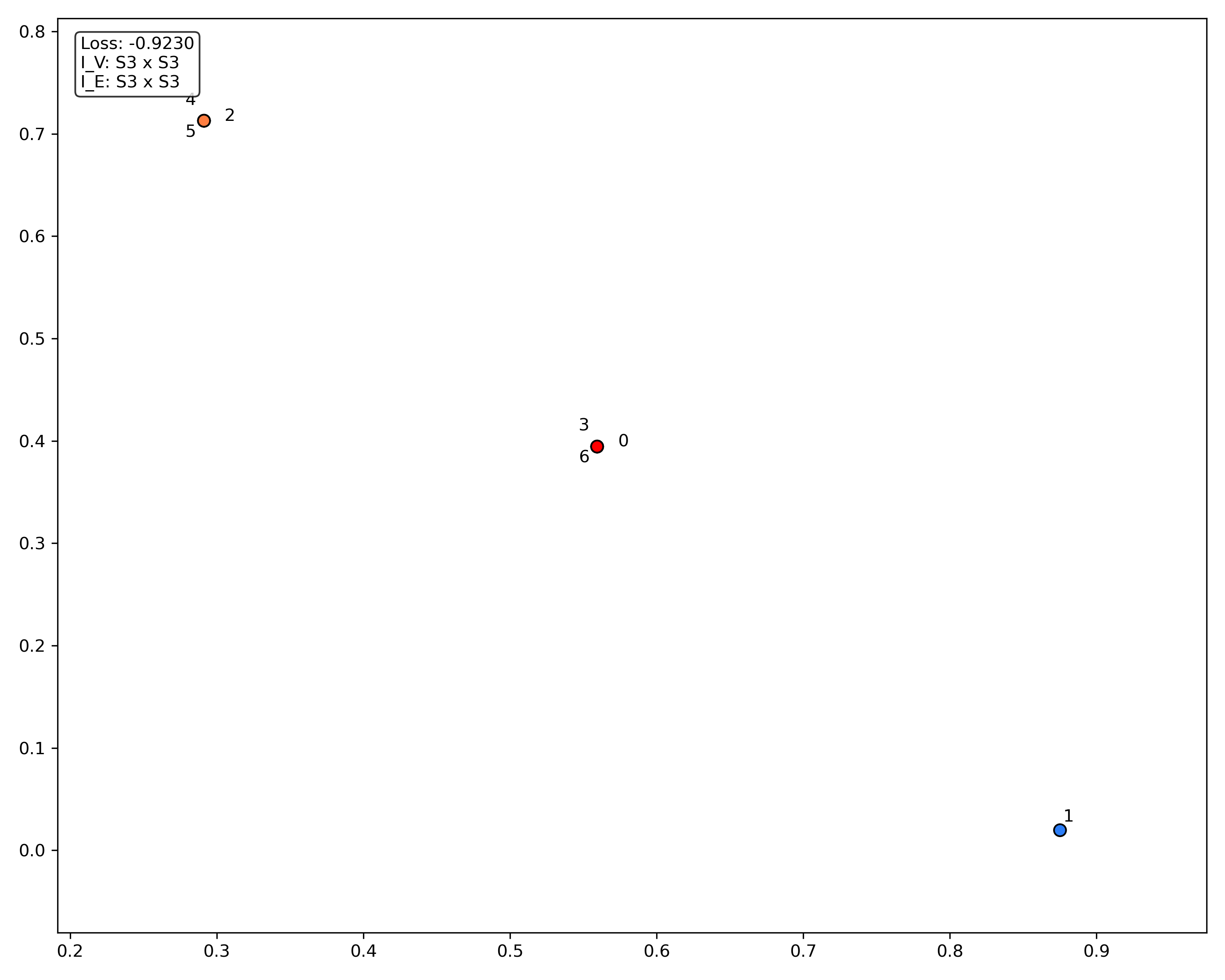}
\caption{}
\end{minipage}

\end{figure}
\begin{figure}[H]
\centering
\begin{minipage}{0.45\textwidth}
\centering
\includegraphics[width=\linewidth]{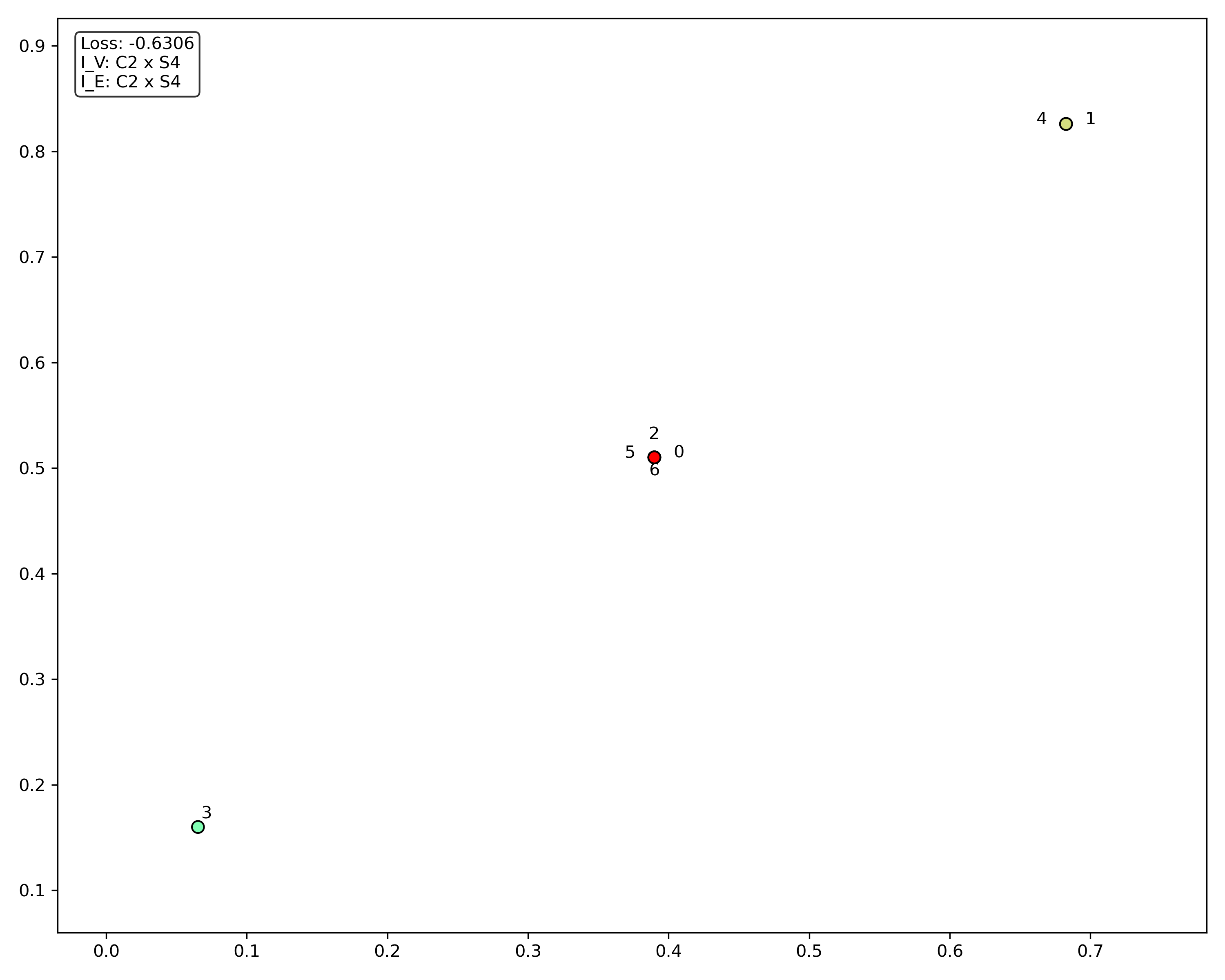}
\caption{}
\end{minipage}\hfill
\begin{minipage}{0.45\textwidth}
\centering
\includegraphics[width=\linewidth]{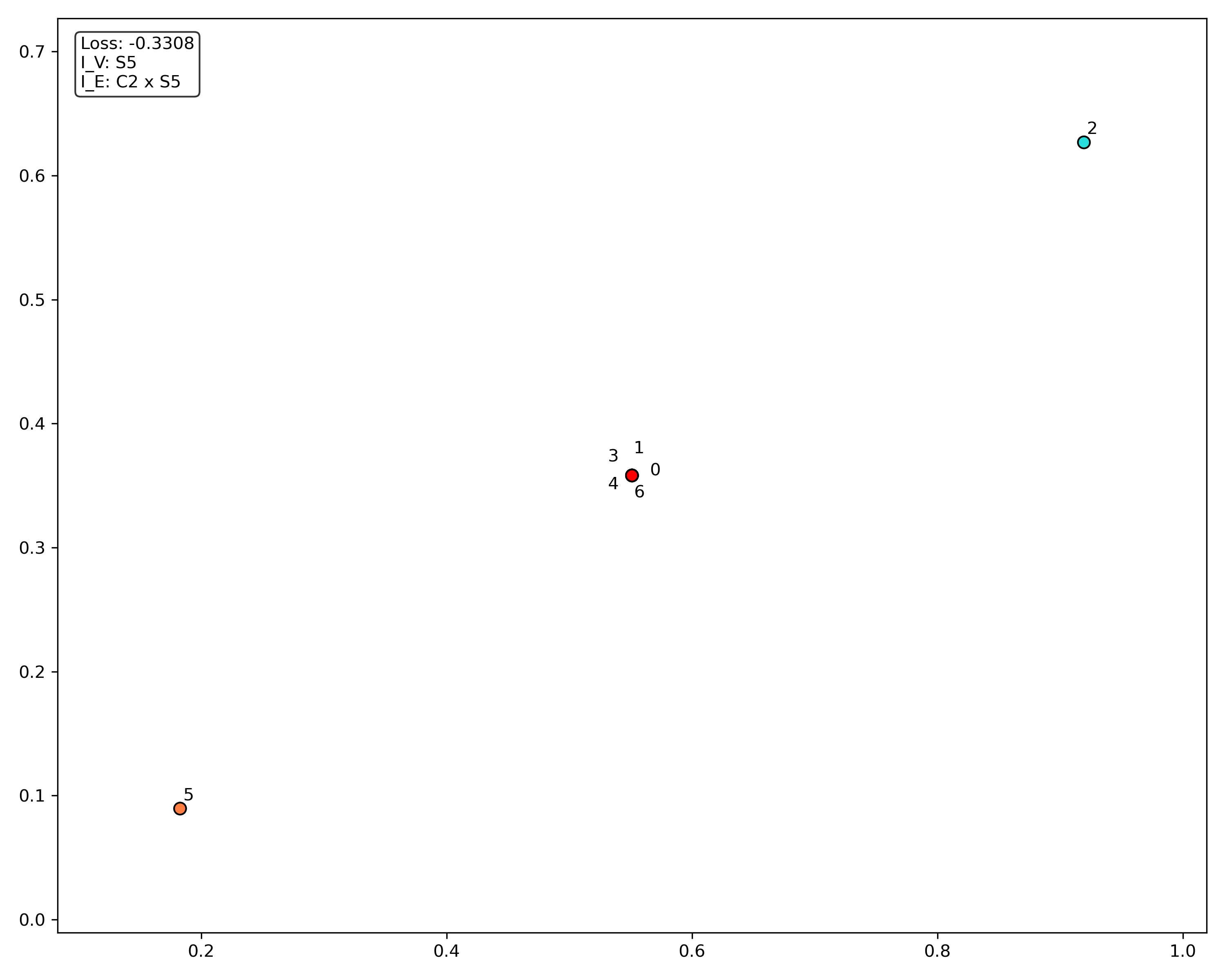}
\caption{}
\end{minipage}
\end{figure}
\subsection{Plots for 4 particles Experiments (kernel $d^2+\frac{1}{d^2}$)}
The plots are ordered in the order of the experiments in the table.\\

\begin{figure}[H]
\caption{Visualization of critical points from Table~\ref{table:experiment_results_repelling_particles_4}}
\label{fig:plots_particle_attarciton_n_4_repelling_ker}% \centering
\centering
% Row 1
% plots_particle_attarciton_n_7_repelling_ker
\begin{minipage}{0.45\textwidth}
\centering
\includegraphics[width=\linewidth]{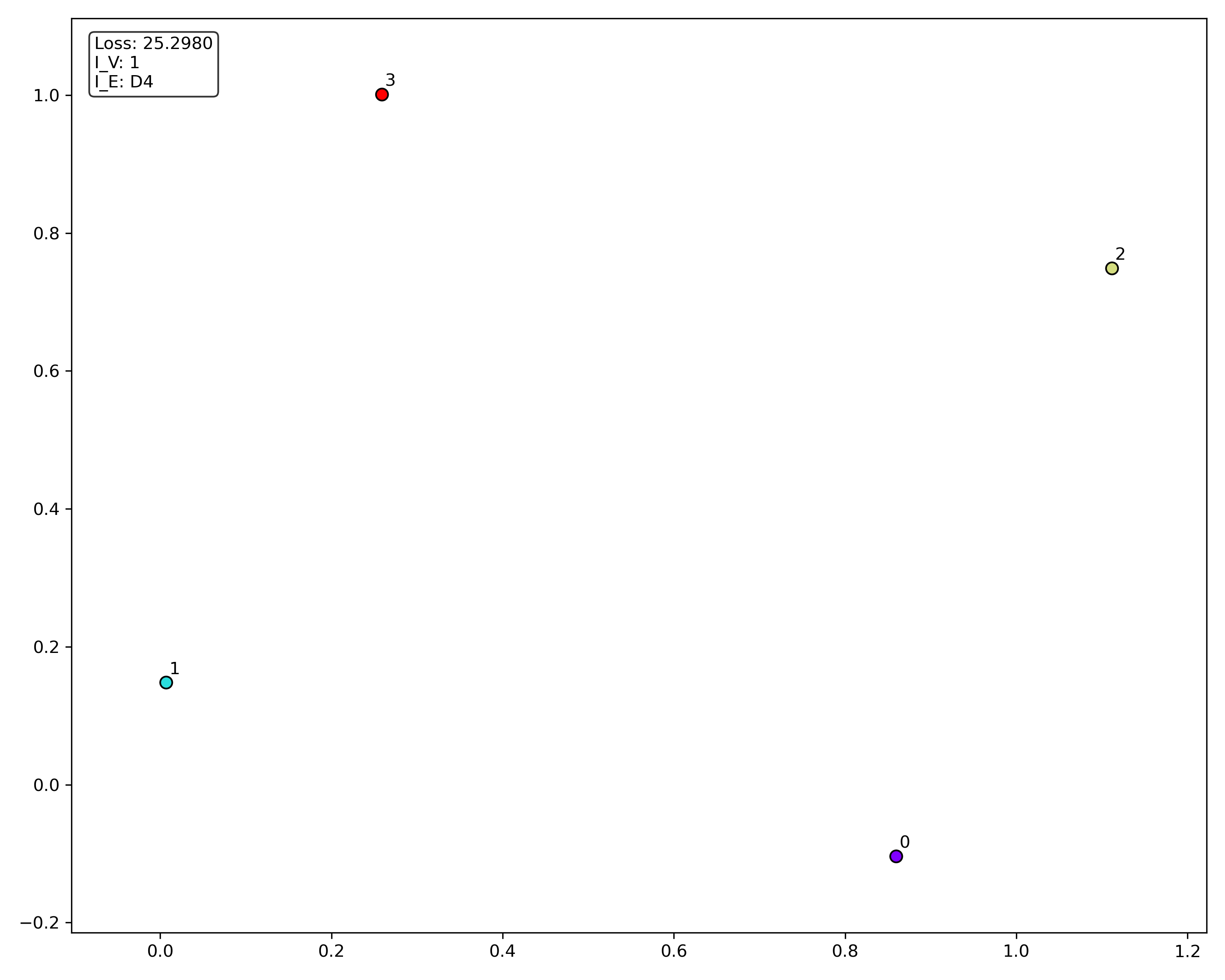}
\caption{}
\end{minipage}\hfill
\begin{minipage}{0.45\textwidth}
\centering
\includegraphics[width=\linewidth]{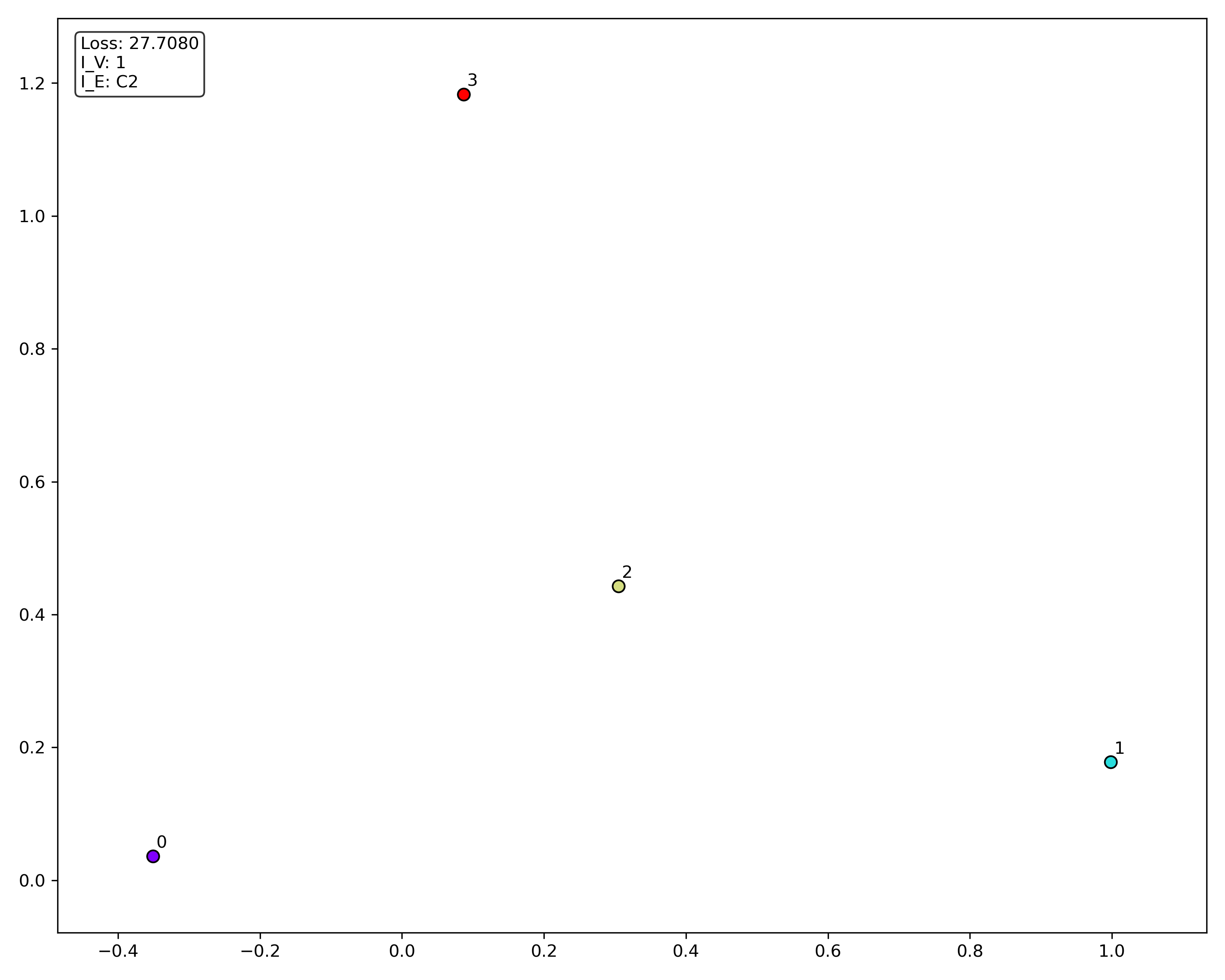}
\caption{}
\end{minipage}
\end{figure}
\begin{figure}[H]
\centering
% Row 2
\begin{minipage}{0.45\textwidth}
\centering
\includegraphics[width=\linewidth]{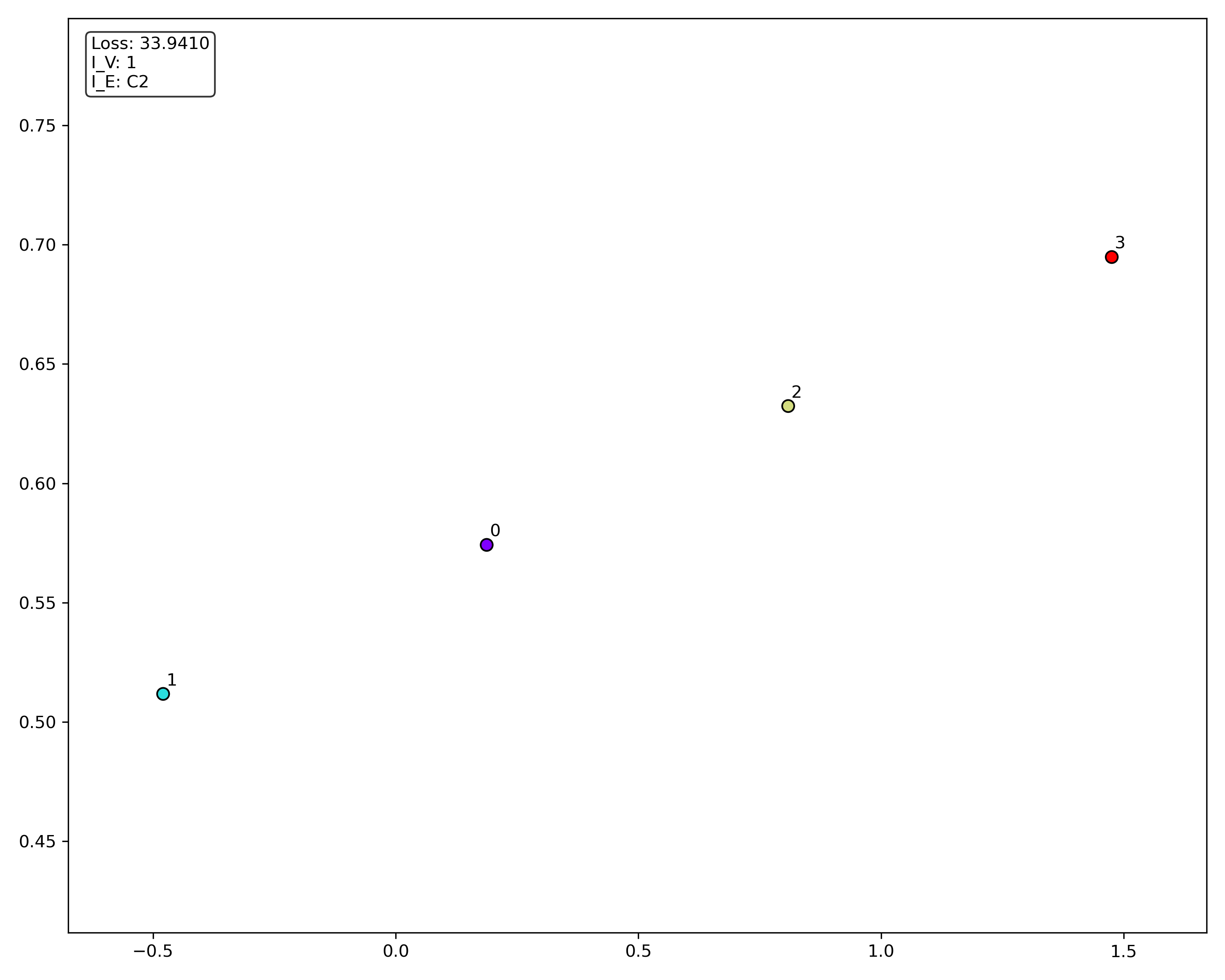}
\caption{}
\end{minipage}\hfill
\end{figure}
\subsection{Plots for 7 particles Experiments (kernel $d^2+\frac{1}{d^2}$)}
The plots are ordered in the order of the experiments in the table.\\
\begin{figure}[H]
\caption{Visualization of critical points from Table~\ref{table:experiment_results_repelling_particles_7}}
\label{fig:plots_particle_attarciton_n_7_repelling_ker}% \centering
\centering
% Row 1
\begin{minipage}{0.45\textwidth}
\centering
\includegraphics[width=\linewidth]{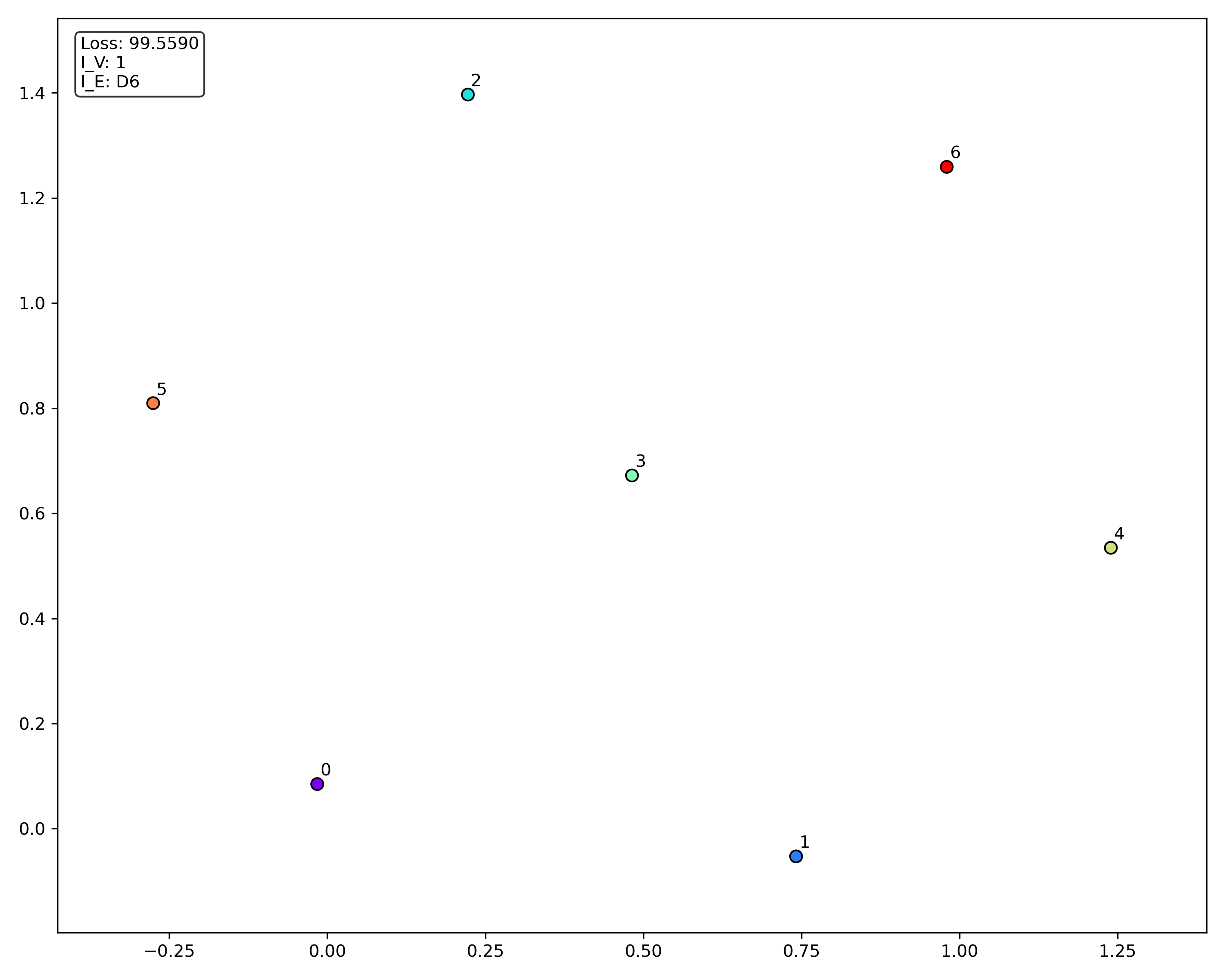}
\caption{}
\end{minipage}\hfill
\begin{minipage}{0.45\textwidth}
\centering
\includegraphics[width=\linewidth]{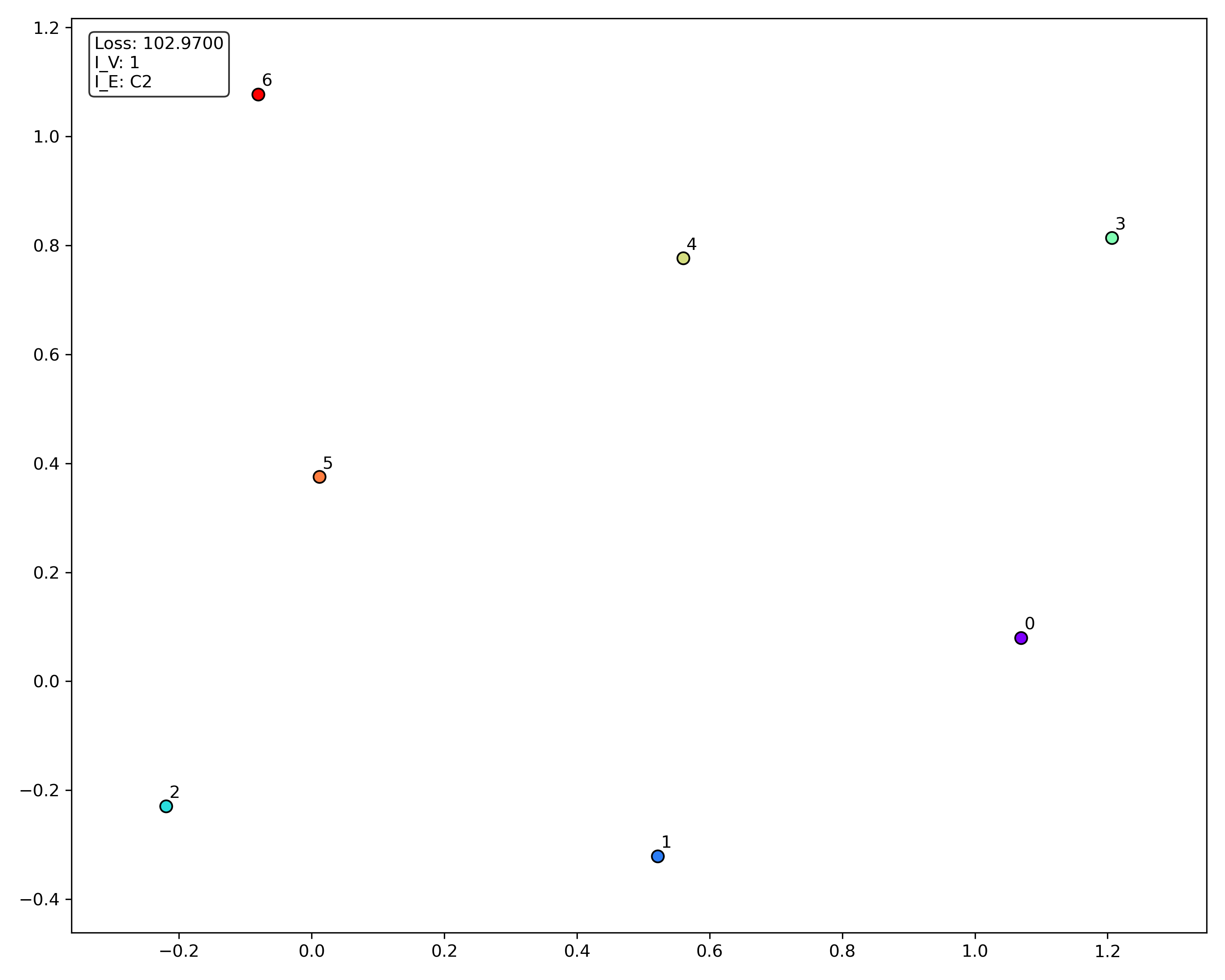}
\caption{}
\end{minipage}
\end{figure}
\begin{figure}[H]
\centering
% Row 2
\begin{minipage}{0.45\textwidth}
\centering
\includegraphics[width=\linewidth]{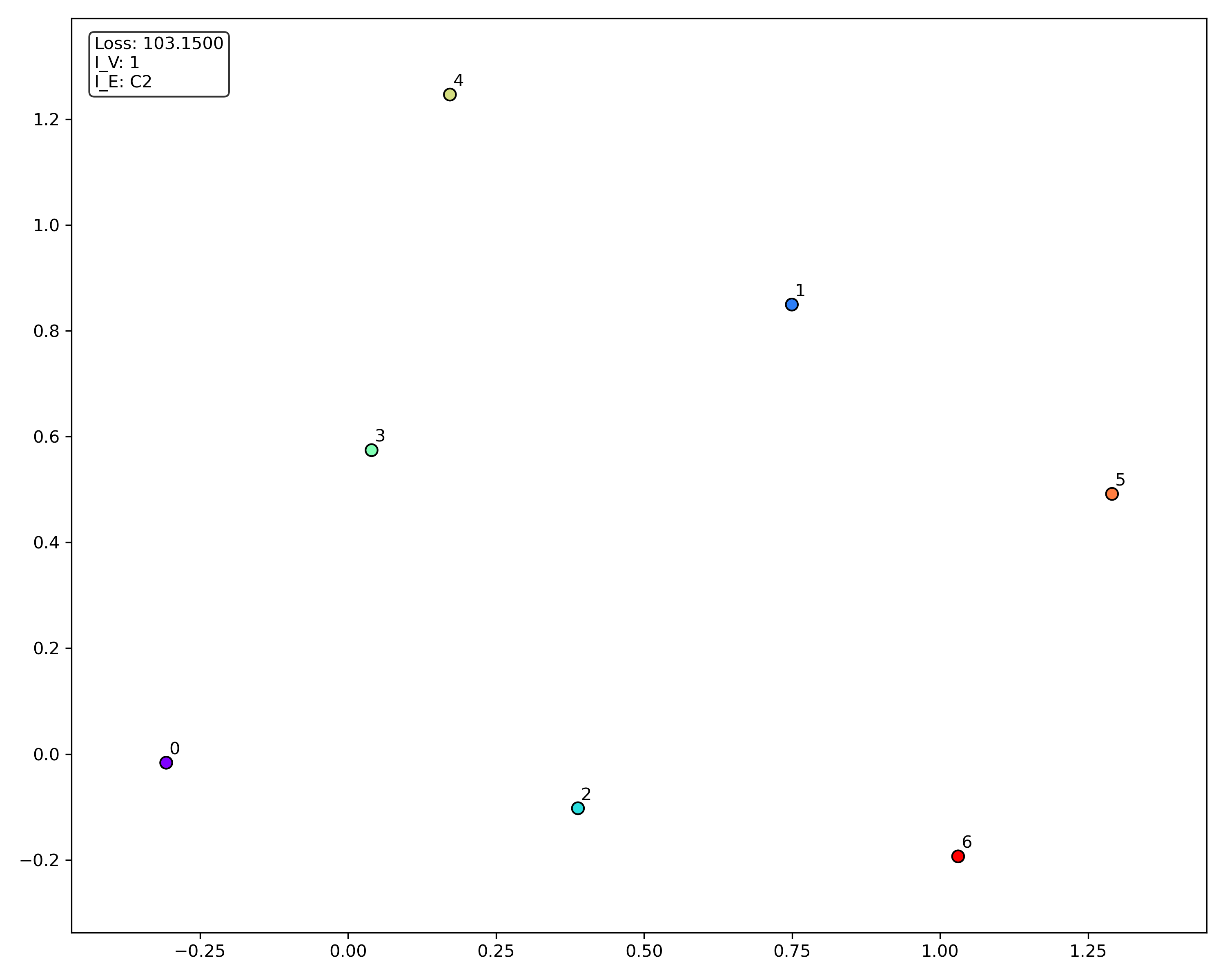}
\caption{}
\end{minipage}\hfill
\begin{minipage}{0.45\textwidth}
\centering
\includegraphics[width=\linewidth]{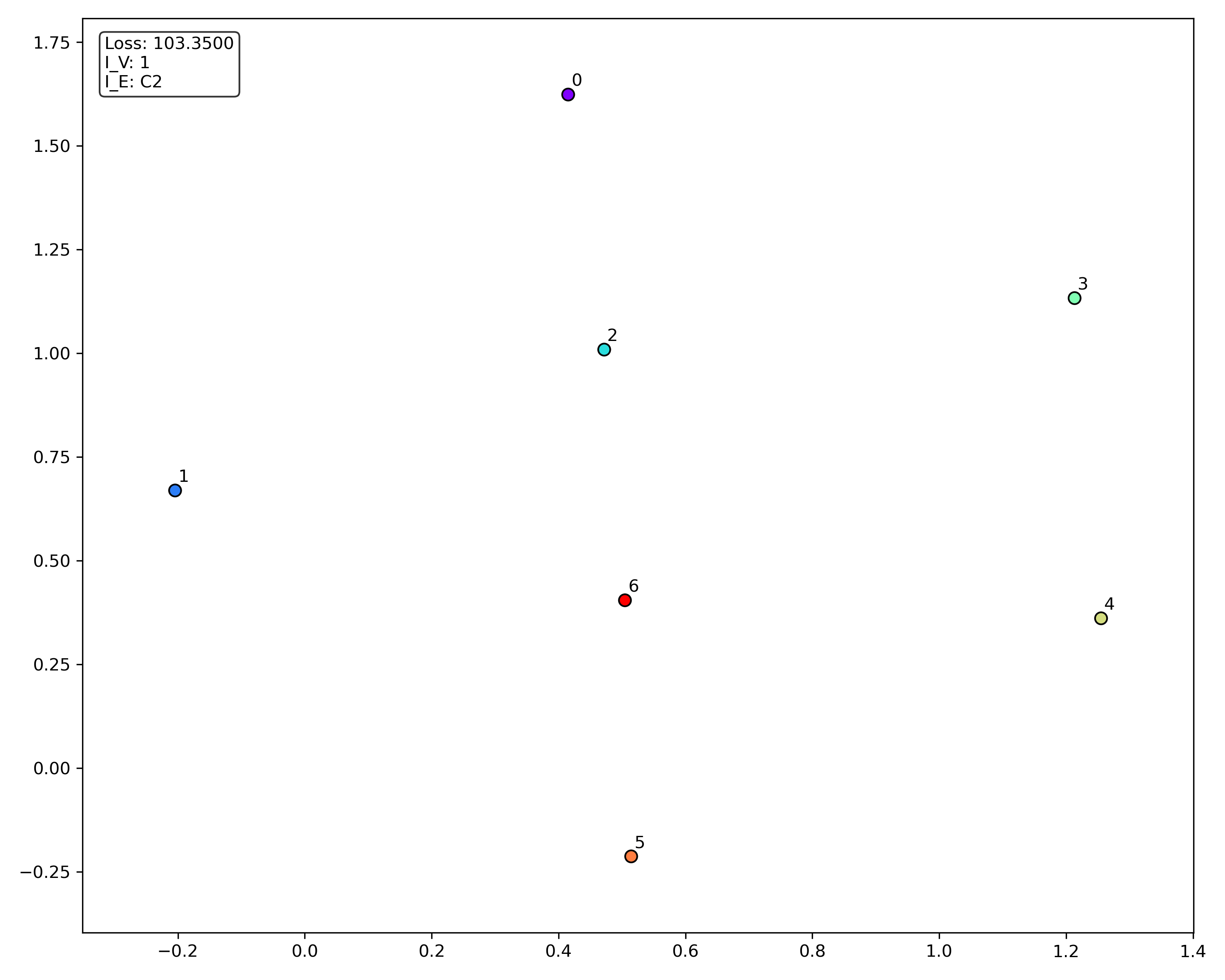}
\caption{}
\end{minipage}
\end{figure}
\begin{figure}[H]
\centering
% Row 3
\begin{minipage}{0.45\textwidth}
\centering
\includegraphics[width=\linewidth]{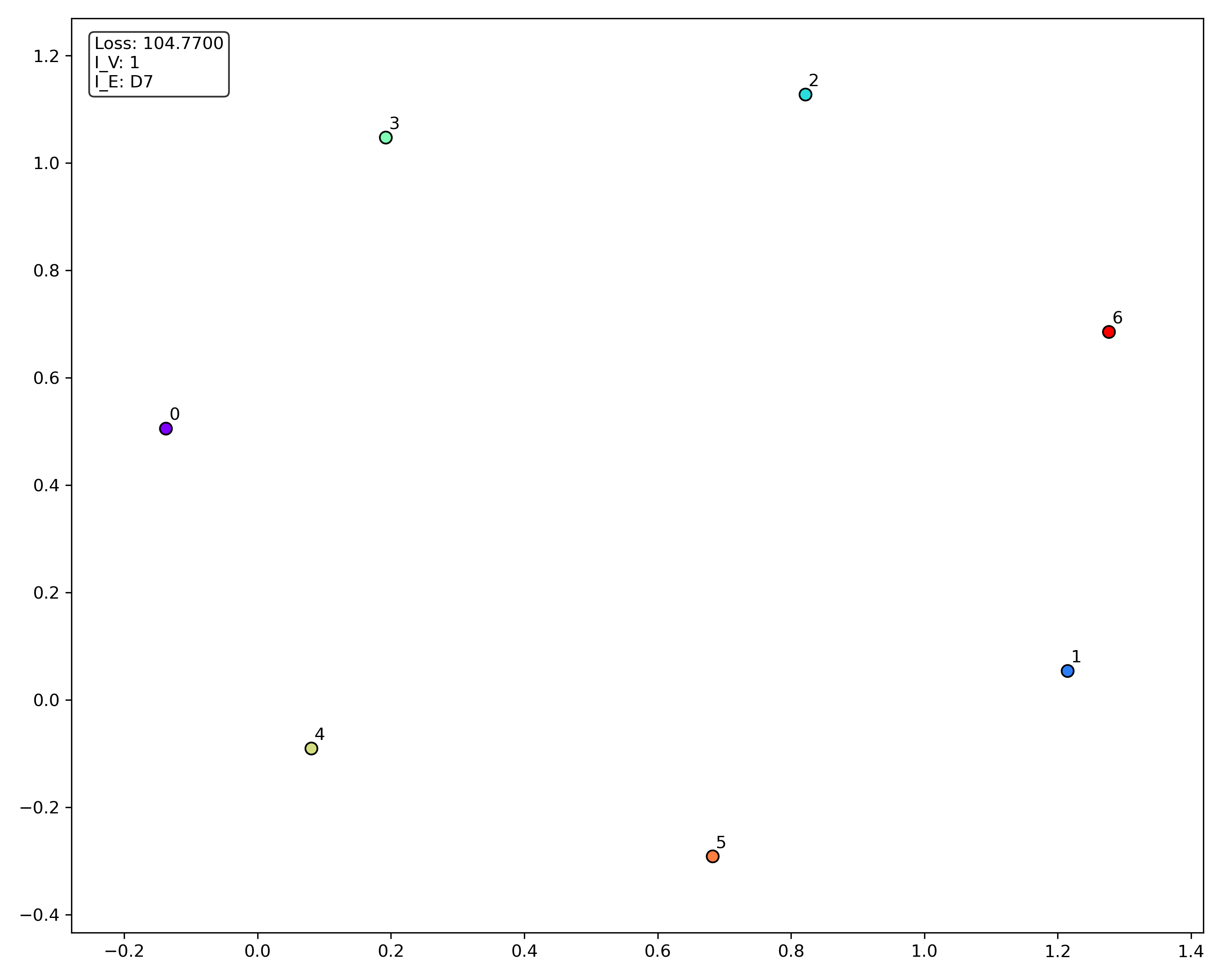}
\caption{}
\end{minipage}\hfill
\begin{minipage}{0.45\textwidth}
\centering
\includegraphics[width=\linewidth]{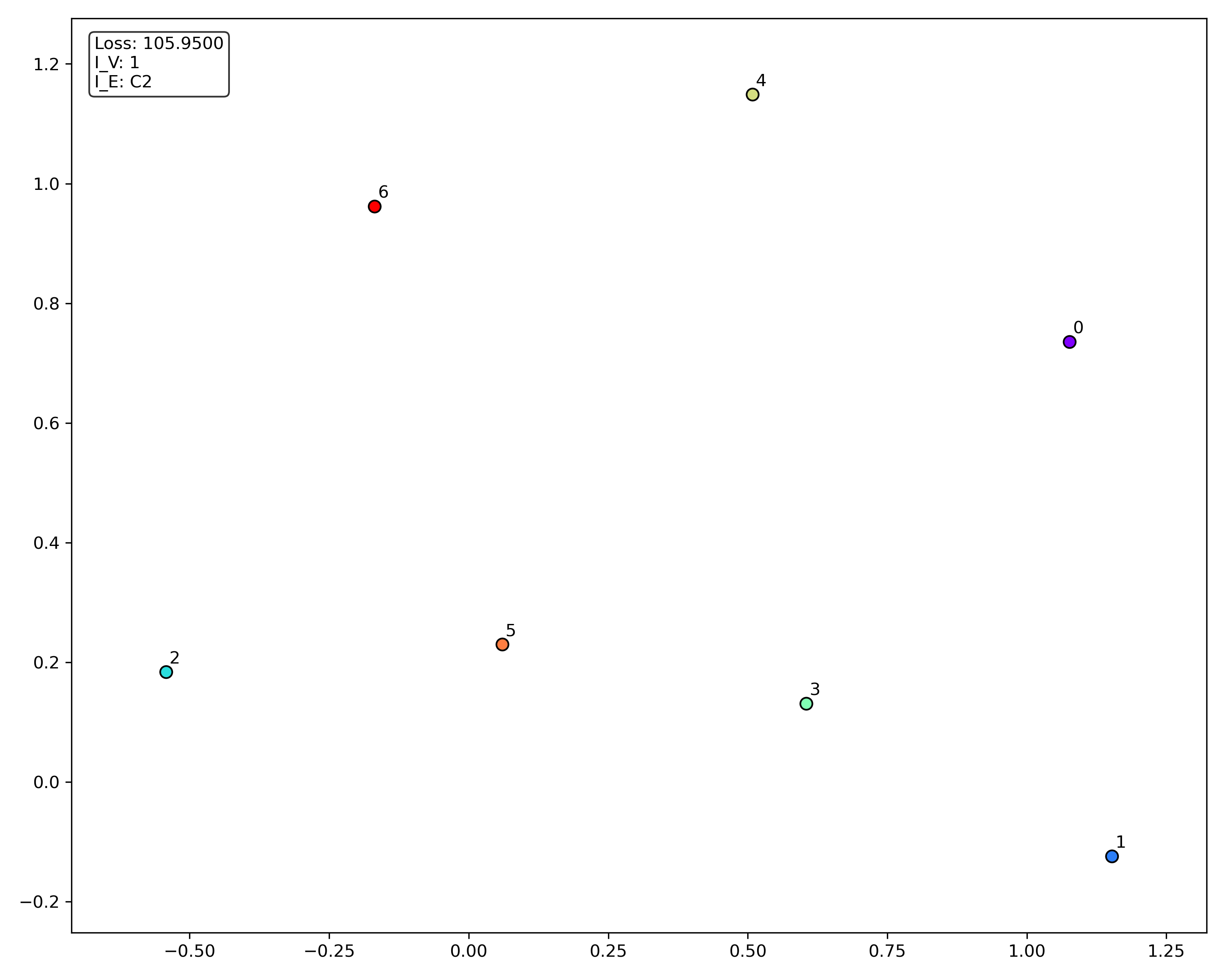}
\caption{}
\end{minipage}

\end{figure}
\begin{figure}[H]
\centering
% Row 1
\begin{minipage}{0.45\textwidth}
\centering
\includegraphics[width=\linewidth]{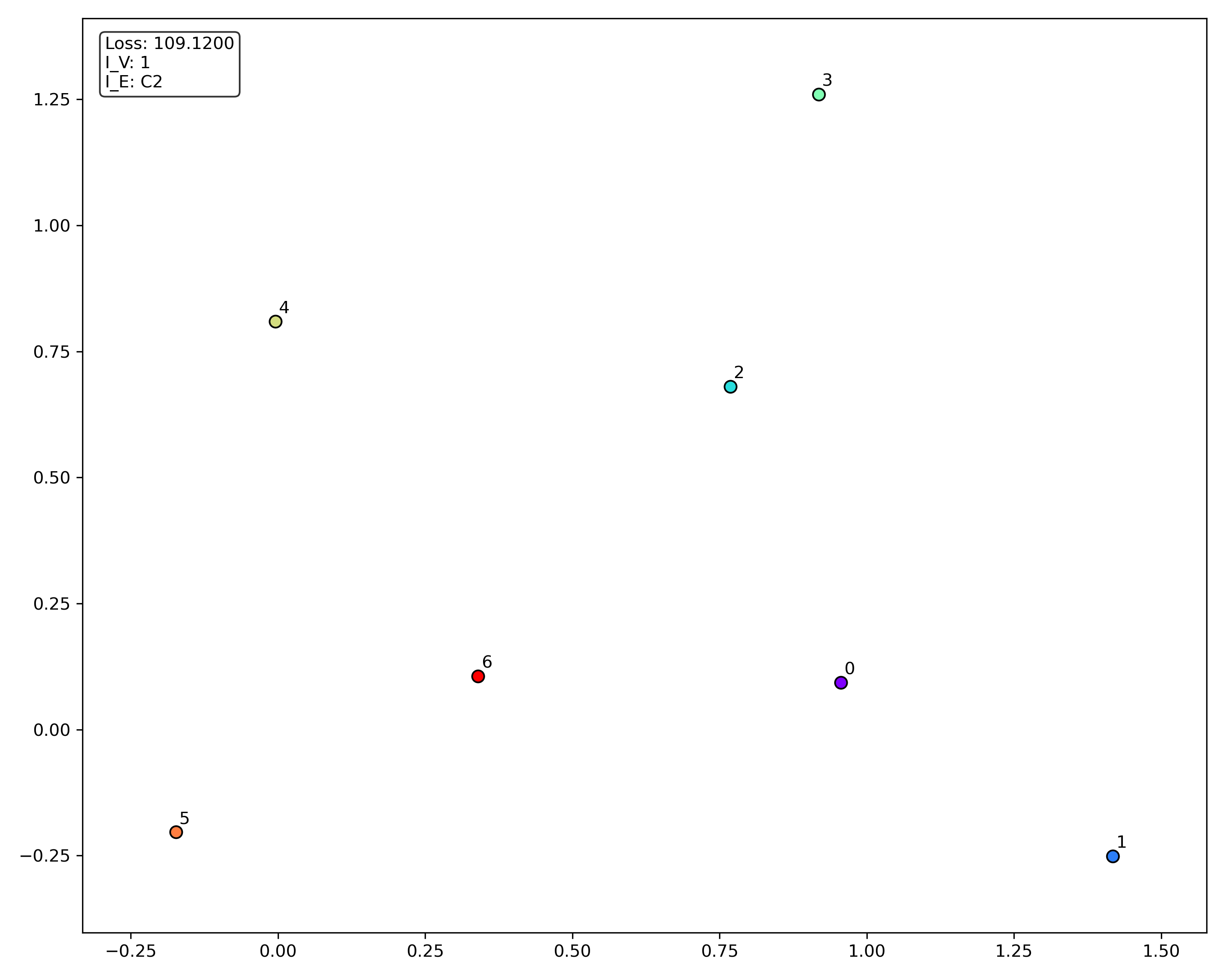}
\caption{}
\end{minipage}\hfill
\begin{minipage}{0.45\textwidth}
\centering
\includegraphics[width=\linewidth]{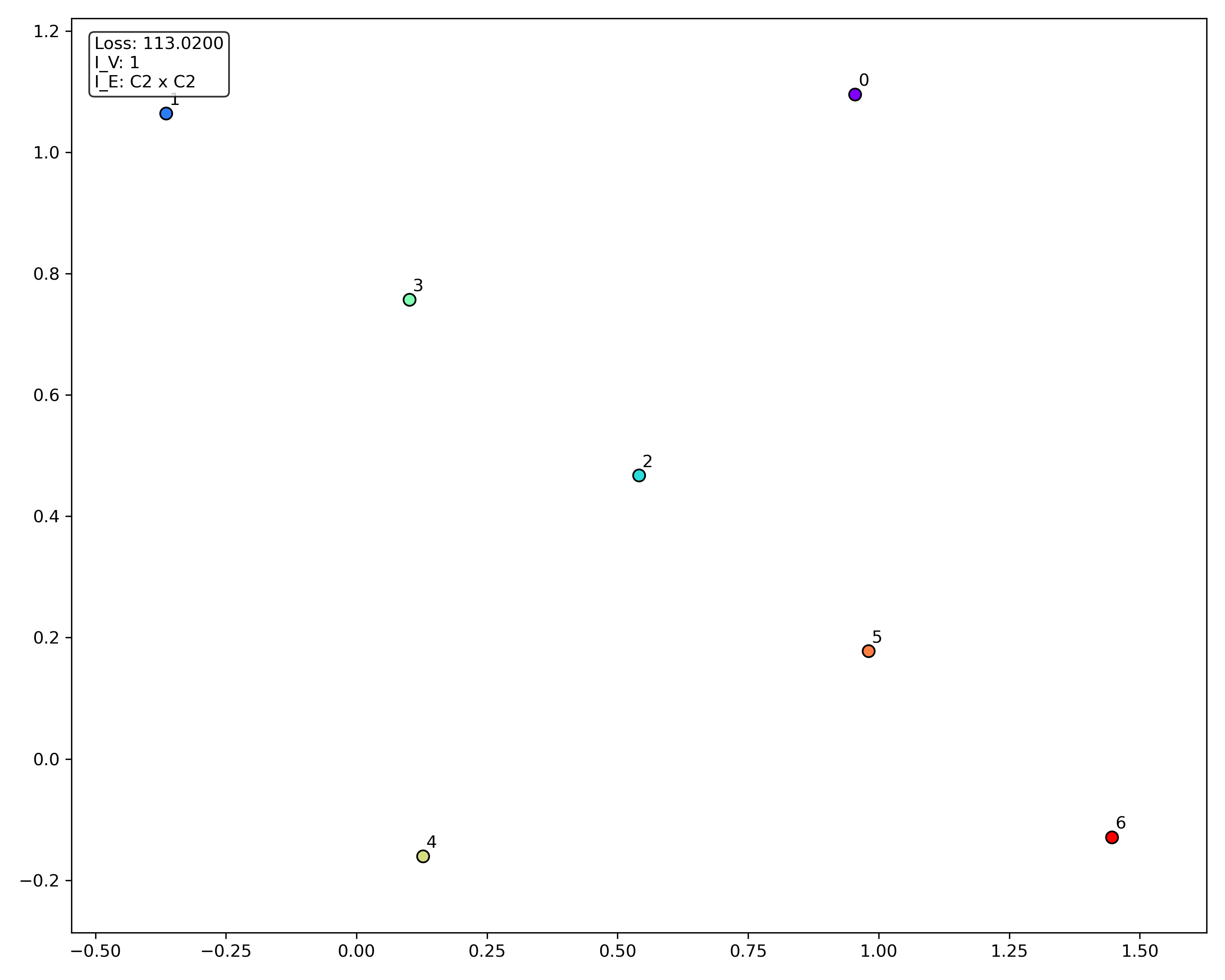}
\caption{}
\end{minipage}
\end{figure}
\begin{figure}[H]
\centering
% Row 2
\begin{minipage}{0.45\textwidth}
\centering
\includegraphics[width=\linewidth]{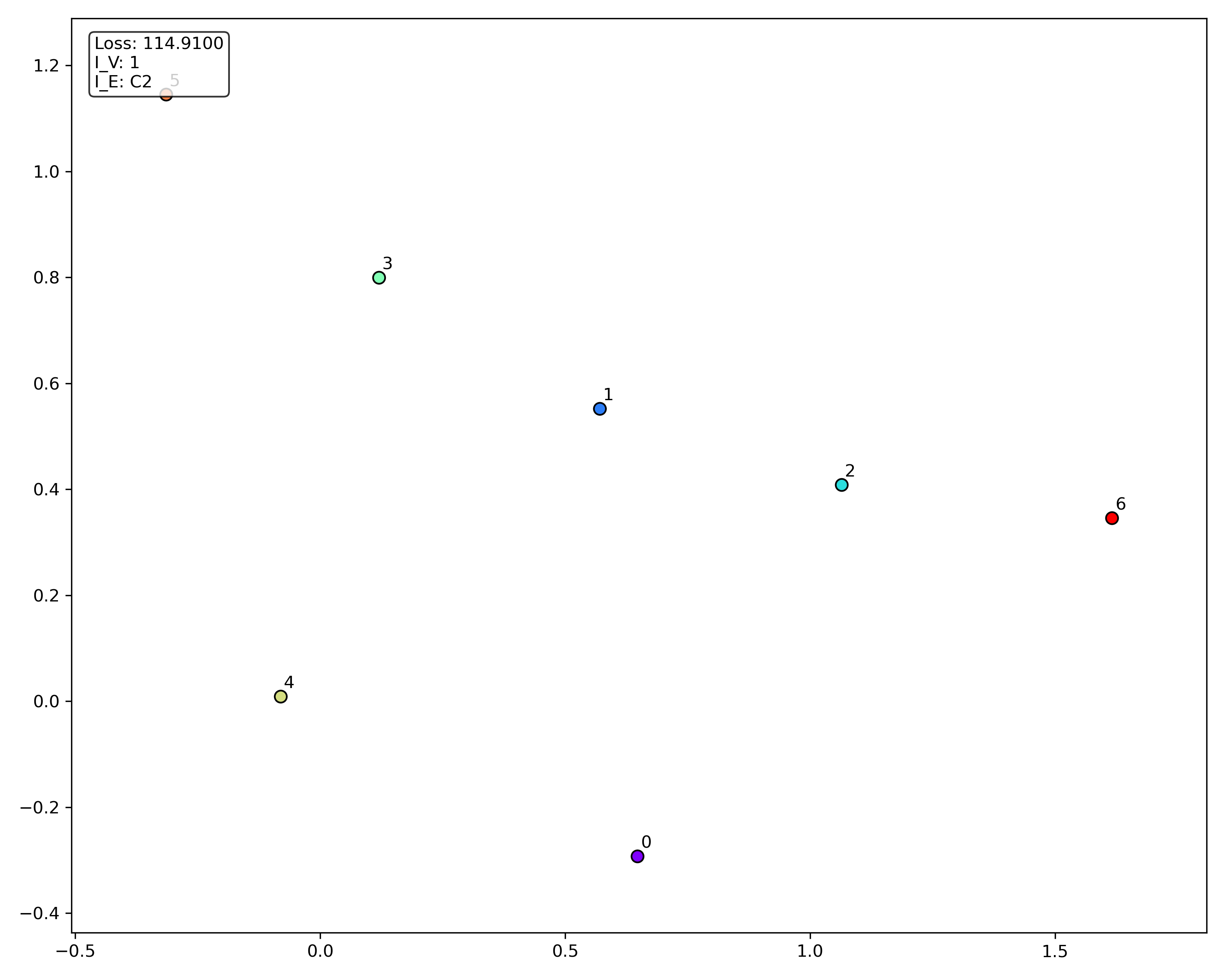}
\caption{}
\end{minipage}\hfill
\begin{minipage}{0.45\textwidth}
\centering
\includegraphics[width=\linewidth]{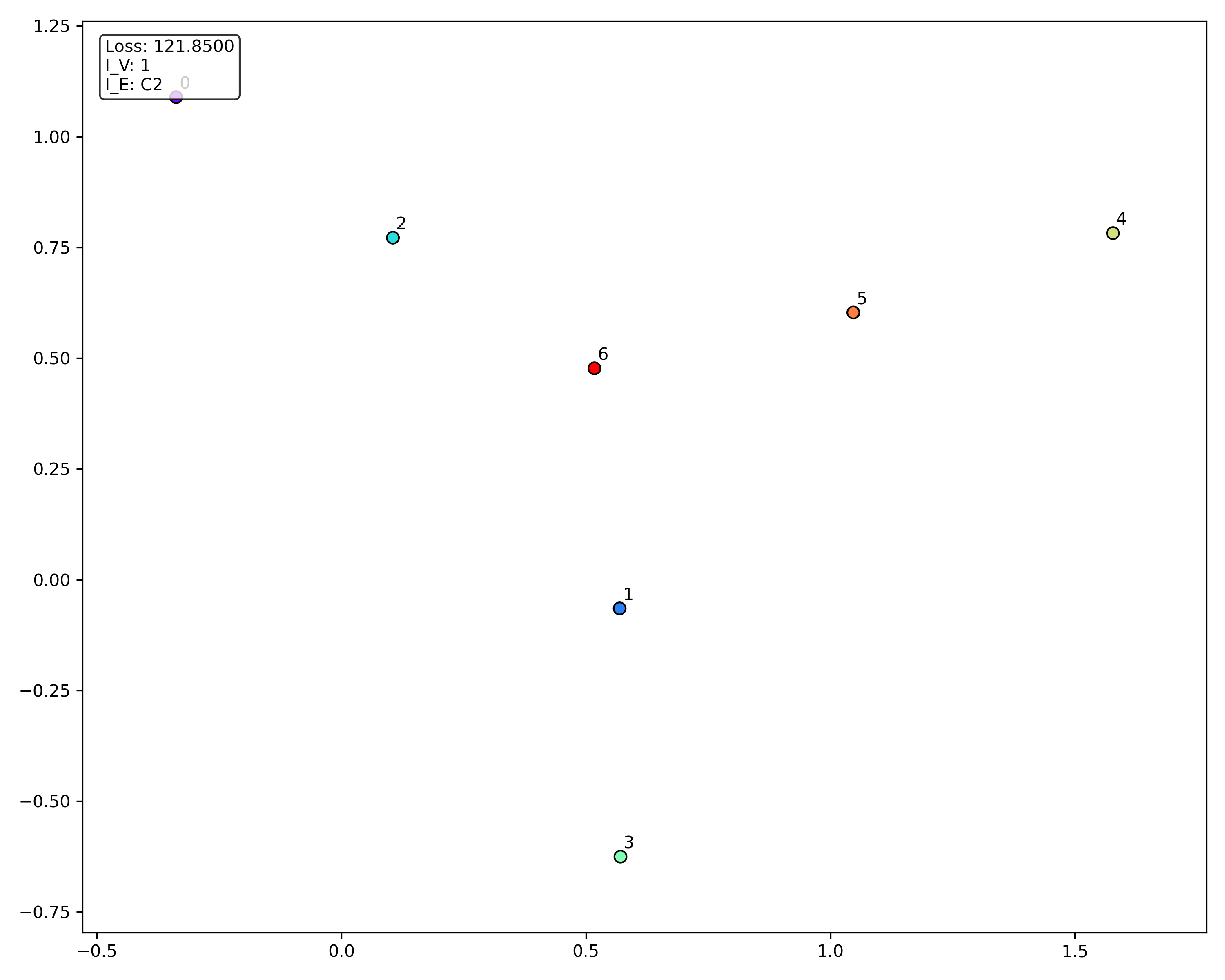}
\caption{}
\end{minipage}
\end{figure}
\begin{figure}[H]
\centering
% Row 3
\begin{minipage}{0.45\textwidth}
\centering
\includegraphics[width=\linewidth]{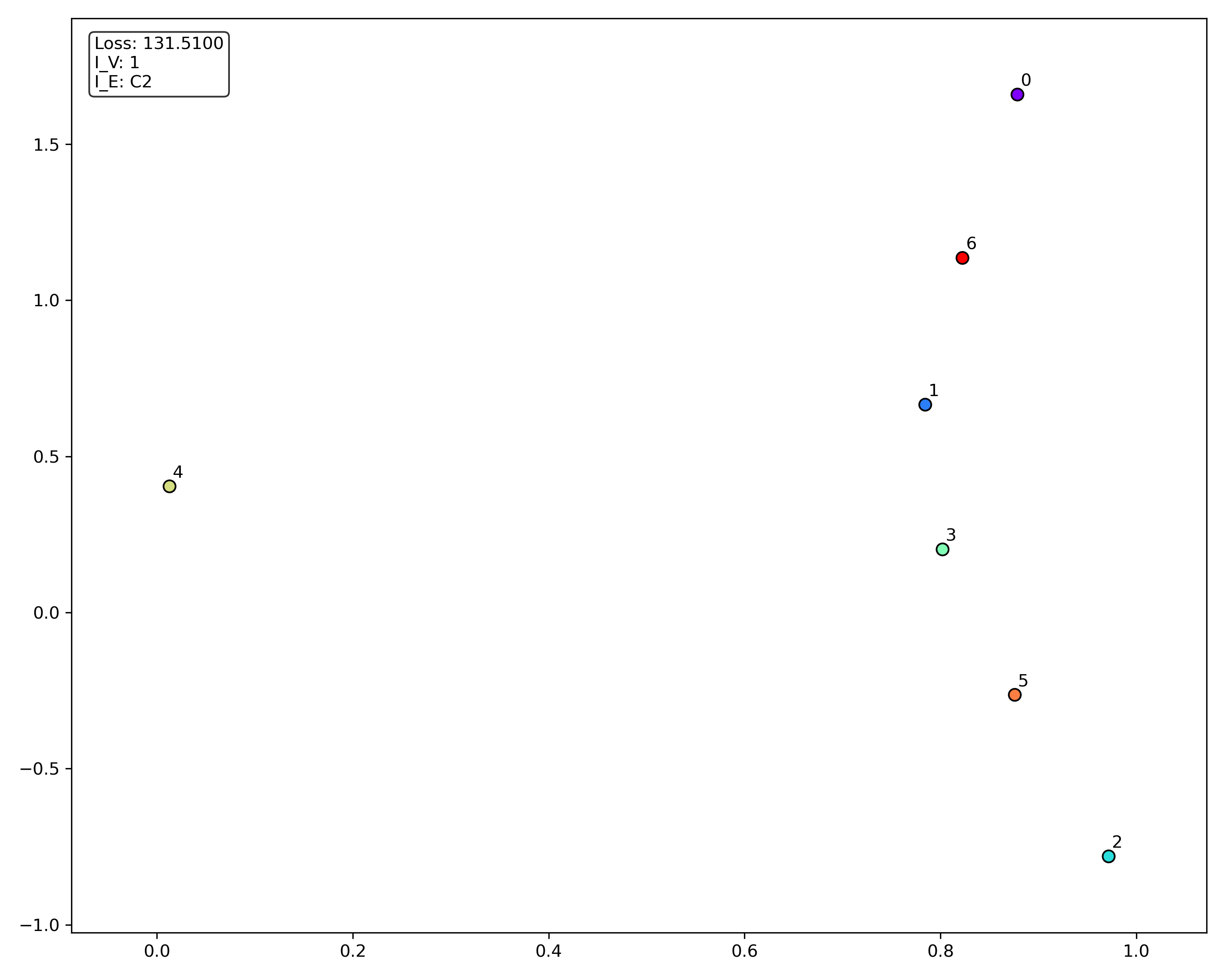}
\caption{}
\end{minipage}\hfill
\begin{minipage}{0.45\textwidth}
\centering
\includegraphics[width=\linewidth]{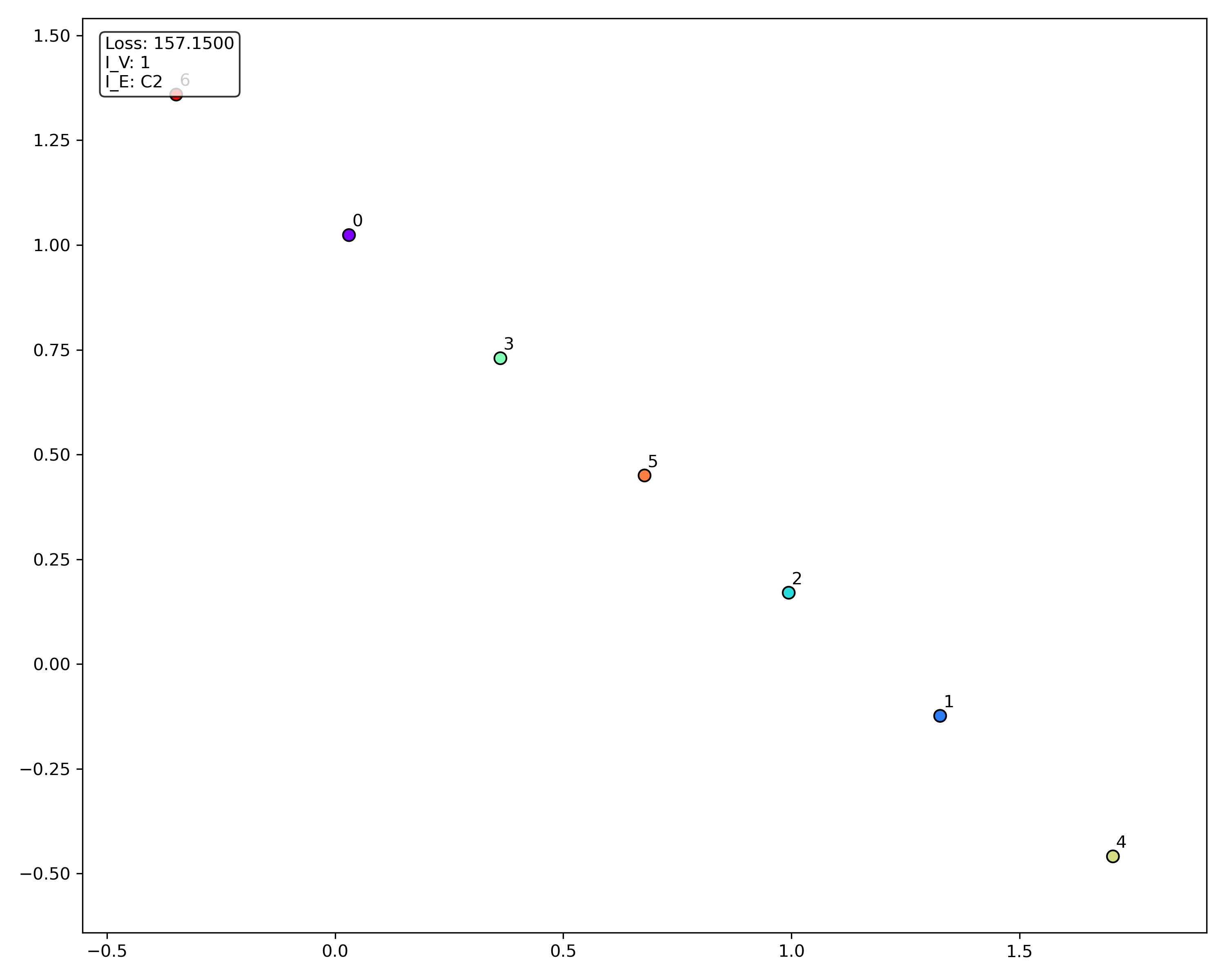}
\caption{}
\end{minipage}

\end{figure}
\section{Symmetry in Loss Landscapes: Particle Attraction vs. Random Polynomial Experiments}\label{app:30MillionExperiments}

In this section, we present two complementary experiments. In the first experiment, we study a symmetric particle attraction loss function, and in the second, we examine a random homogeneous polynomial with similar bounded-below properties. The purpose is to compare the number of observable local minima in both settings and to discuss how symmetry may serve as a mechanism to limit the effective number of minima.

\section{Symmetric Particle Attraction Experiment} \label{app:RandomPolynomialExperiment1}

Consider points \(x_1,\ldots,x_4 \in \mathbb{R}^2\) and define the kernel
\[
\kappa(x_i,x_j) = \langle x_i,x_j \rangle^{8} - \langle x_i,x_j \rangle^3,
\]
with the total energy (loss) function given by
\[
L(x_1,\ldots,x_4) = \sum_{i,j} \kappa(x_i,x_j).
\]
Since the potential is invariant under any 2D isometry applied simultaneously to all particles (due to the inner product), the loss is invariant under the full orthogonal group \(O_2\). To eliminate the trivial isometry symmetry, we fix the second coordinate of the first particle to be zero. This choice reduces the ambient space from \(\mathbb{R}^8\) (with four particles in \(\mathbb{R}^2\)) to \(\mathbb{R}^7\) and provides a canonical representative under the \(O_2\) action. In addition, a sign symmetry remains (multiplying all \(x\)- or all \(y\)-coordinates by \(-1\)), and the particle permutation symmetry group \(S_4\) (restricted to those permutations that preserve the fixed second coordinate of the first particle) continues to act on the configuration.

Using \(30\times 10^6\) randomly generated initial points, Gradient Descent converged to only 4 distinct orbits of local minima. In other words, the system exhibits 4 types of local minima, and when the group actions are taken into account, there are a total of 54 observable local minima. The following is a summary of the minima types:

\begin{itemize}
    \item \textbf{Type $P_1$:} 2 minima with 
    \[
    P_1 = \begin{bmatrix} 0.906 & 0 \\ 0.906 & 0 \\ 0.906 & 0 \\ 0.906 & 0 \end{bmatrix}, \quad L(P_1) = -5.55.
    \]
    \item \textbf{Type $P_2$:} 12 minima with 
    \[
    P_2 = \begin{bmatrix} 0.906 & 0 \\ 0.906 & 0 \\ 0 & 0.906 \\ 0 & 0.906 \end{bmatrix}, \quad L(P_2) = -2.77.
    \]
    \item \textbf{Type $P_3$:} 16 minima with 
    \[
    P_3 = \begin{bmatrix} 0.906 & 0 \\ 0.906 & 0 \\ 0.906 & 0 \\ 0 & 0.906 \end{bmatrix}, \quad L(P_3) = -3.46.
    \]
    \item \textbf{Type $P_4$:} 24 minima with 
    \[
    P_4 = \begin{bmatrix} 0.90146749 & 0 \\ 0.90146749 & 0 \\ -0.34181185 & -0.77588592 \\ -0.34181185 & \;0.77588592 \end{bmatrix}, \quad L(P_4) = -1.51.
    \]
\end{itemize}

Thus, a total of 54 distinct local minima (accounting for all symmetry-related orbits) were observed in the symmetric case.

\section{Particle Attraction Case For Random 4 degree kernel}\label{app:RandomKer}

We conducted several experiments using different fourth-degree polynomial loss functions. Each experiment uses a different polynomial, detailed in the Appendix. The results are summarized in Table~\ref{tab:results}.
\begin{table}[H]
\centering
\caption{Experimental Results}
\label{tab:results}
\begin{tabular}{@{}cccc@{}}
\toprule
\textbf{Experiment} & \textbf{Isotropy Group} & \textbf{Final Positions} & \textbf{Final Loss} \\ \midrule

\textbf{1} &
\begin{tabular}{l}
Order: 8 \\
Structure: $D_4$
\end{tabular} &
\begin{tabular}{l}
$v_0 = 0.68269$ \\
$v_1 = 0.28440$ \\
$v_2 = 0.68269$ \\
$v_3 = 0.68269$ \\
$v_4 = 0.53480$ \\
$v_5 = 0.68269$
\end{tabular} &
$-9.009$ \\ \midrule

\textbf{2} &
\begin{tabular}{l}
Order: 6 \\
Structure: $S_3$
\end{tabular} &
\begin{tabular}{l}
$v_0 = -0.55875$ \\
$v_1 = 0.93261$ \\
$v_2 = 0.93261$ \\
$v_3 = 0.93261$ \\
$v_4 = -0.55875$ \\
$v_5 = -0.55875$
\end{tabular} &
$-14.536$ \\ \midrule

\textbf{3} &
\begin{tabular}{l}
Order: 48 \\
Structure: $C_2 \times S_4$
\end{tabular} &
\begin{tabular}{l}
$v_0 = -0.31517$ \\
$v_1 = -0.31517$ \\
$v_2 = -0.31517$ \\
$v_3 = -0.31517$ \\
$v_4 = -0.31517$ \\
$v_5 = -0.31517$
\end{tabular} &
$-5.7053$ \\ \midrule

\textbf{4} &
\begin{tabular}{l}
Order: 48 \\
Structure: $C_2 \times S_4$
\end{tabular} &
\begin{tabular}{l}
$v_0 = 0.29563$ \\
$v_1 = 0.29563$ \\
$v_2 = 0.29563$ \\
$v_3 = 0.29563$ \\
$v_4 = 0.29563$ \\
$v_5 = 0.29563$
\end{tabular} &
$-2.4287$ \\ \midrule

\textbf{5} &
\begin{tabular}{l}
Order: 6 \\
Structure: $S_3$
\end{tabular} &
\begin{tabular}{l}
$v_0 = 0.50970$ \\
$v_1 = 0.50970$ \\
$v_2 = 0.50970$ \\
$v_3 = -0.37541$ \\
$v_4 = -0.37541$ \\
$v_5 = -0.37541$
\end{tabular} &
$-2.1555$ \\ \bottomrule

\end{tabular}
\end{table}

\subsection*{Loss Functions for Each Experiment}

In each experiment, the loss function \( L \) is a fourth-degree polynomial defined over the variables \( a_1, a_2, b_1, b_2 \), corresponding to the edge vectors in the graph. The coefficients and monomials used in each experiment are listed below.
We chose 30 monomials out of the 70 monomials of degree 4 in 4 variables. The coefficients were randomly chosen from a standard normal distribution.
Where for the 4 degree coefficients is positive and the exponents of 4 degree monomials is even, I used this conditions to ensure the polynomial is bounded below.\\
\textbf{Experiment 1:}
\[
\begin{array}{rcl}
f(a,b) &=& -0.3391\,b_2 + 0.4791\,a_1^4 + 0.1360\,a_1b_1^2 + 0.2717\,a_2b_1b_2^2 - 0.5321\,a_1b_2 - 0.9043\,b_1^2 \\
&& +\, 0.7399\,a_2b_2 + 0.1840\,b_2^3 + 0.6584\,b_1b_2 - 0.8382\,a_1b_2^2 + 0.8152\,a_2^2b_2 - 0.1054\,a_1 \\
&& +\, 0.5557\,a_1^3b_2 + 0.1230\,a_1a_2^2b_1 - 0.7825\,a_2^2b_1 + 0.5794\,a_1a_2b_1b_2 - 0.8400\,a_2^3 \\
&& -\, 0.9640\,b_2 + 0.8577\,a_2 - 0.4882\,b_1 + 0.4577\,a_1^2b_2 + 0.3470\,b_2^2 + 0.4239\,a_1^3 \\
&& -\, 0.3558\,a_2^2 + 0.4574\,b_2^4 - 0.1283\,b_1^2b_2 + 0.9085\,b_2 - 0.5772\,a_1a_2 - 0.8940\,a_1^2a_2 \\
&& +\, 0.2102\,a_2^2b_1.
\end{array}
\]

\textbf{Experiment 2:}
\[
\begin{array}{rcl}
f(a,b) &=& 0.3677\,a_1^2a_2b_1 - 0.0404\,a_2b_2^2 + 0.8227\,a_1^4 + 0.7633\,a_1 + 0.6529\,a_2 + 0.0127\,a_2^2b_1 \\
&& +\, 0.6695\,a_2^3b_1 + 0.7561\,a_2b_1^3 + 0.2934\,a_1^3b_1 - 0.8204\,a_2b_2^2 + 0.7910\,a_2^2b_2^2 \\
&& +\, 0.5344\,a_1 - 0.8581\,a_2^2 + 0.5033\,b_1^4 - 0.9312\,a_1^3 + 0.5393\,b_1b_2^2 + 0.9664\,a_2b_1b_2^2 \\
&& +\, 0.1014\,a_2b_2 - 0.3991\,b_2^3 - 0.8326\,b_2 + 0.2821\,a_1b_2^2 + 0.0106\,a_1a_2 - 0.7970\,a_1 \\
&& +\, 0.8012\,a_1a_2^2 + 0.9380\,a_1a_2b_1^2 - 0.2501\,a_2^2b_2 + 0.3412\,b_2^3 + 0.9164\,a_1b_2 \\
&& -\, 0.5956\,a_2 + 0.2598\,a_2^2b_2.
\end{array}
\]

\textbf{Experiment 3:}
\[
\begin{array}{rcl}
f(a,b) &=& 0.9736\,b_1^2b_2 + 0.4452\,a_1^2a_2b_2 + 0.6006\,b_1^4 + 0.3478\,a_2^2b_1b_2 - 0.0799\,b_2 \\
&& +\, 0.7705\,b_1b_2^3 - 0.0882\,a_1b_1 + 0.2263\,a_1^2b_1^2 + 0.0047\,a_2b_1b_2 - 0.6979\,a_2b_2^2 \\
&& +\, 0.8158\,a_1b_1^3 - 0.5612\,a_1b_2 - 0.7512\,a_1b_2^2 + 0.6146\,a_2b_1^2b_2 + 0.9060\,a_1^3b_2 \\
&& +\, 0.3293\,a_1b_1b_2^2 - 0.4857\,a_1b_2 + 0.6202\,a_1 + 0.7917\,a_1^2b_2^2 + 0.7527\,b_1^2b_2 \\
&& +\, 0.4912\,a_1a_2^2b_1 + 0.6479\,a_1a_2^2b_2 - 0.0854\,a_2^3 + 0.4726\,a_2b_1^3 - 0.1148\,b_1 \\
&& +\, 0.2498\,b_2^4 + 0.2668\,a_2^3b_1 + 0.2609\,a_1a_2b_1^2 + 0.3574\,a_1b_1^2b_2 + 0.9839\,a_1^2a_2b_1.
\end{array}
\]

\textbf{Experiment 4:}
\[
\begin{array}{rcl}
f(a,b) &=& 0.8046\,a_2^3b_1 - 0.0903\,a_1a_2b_1 - 0.3022\,a_2b_1^2 + 0.7107\,a_1a_2b_2^2 + 0.3738\,a_1^2a_2^2 \\
&& +\, 0.4659\,a_1a_2b_1 - 0.5364\,a_2b_1b_2 + 0.4696\,a_1b_1^2b_2 + 0.2321\,a_1b_2 + 0.4971\,a_2^2b_1b_2 \\
&& +\, 0.6020\,a_1a_2^3 - 0.8185\,a_1 - 0.9552\,a_1^2b_1 - 0.9612\,a_2^3 + 0.1544\,a_2^2b_1 + 0.9042\,b_1^2b_2^2 \\
&& +\, 0.4775\,b_2 + 0.4483\,a_1a_2b_2 + 0.8384\,a_1b_2 - 0.8172\,a_1a_2^2 + 0.5101\,b_1^3 \\
&& +\, 0.3645\,b_1b_2^2 + 0.8480\,a_1b_1^3 + 0.2504\,a_1a_2^2b_1 - 0.7310\,a_1b_2 + 0.4990\,b_2^3 \\
&& +\, 0.3686\,a_2b_1b_2 - 0.6202\,b_1^3 - 0.2180\,a_2b_2^2 + 0.0190\,a_1.
\end{array}
\]

\textbf{Experiment 5:}
\[
\begin{array}{rcl}
f(a,b) &=& 0.6114\,a_1a_2^2b_1 - 0.1573\,a_1a_2 - 0.4525\,b_2^2 + 0.5590\,b_1b_2^3 + 0.5904\,a_1b_2 \\
&& +\, 0.4621\,a_2^2 + 0.9355\,a_1^4 - 0.7042\,a_1a_2b_2 - 0.5839\,a_1^2 + 0.4718\,a_1b_1 \\
&& +\, 0.1064\,b_1^2b_2^2 - 0.3174\,a_1b_1^2 - 0.3987\,b_2 + 0.7050\,a_1^3a_2 + 0.5159\,a_1a_2b_1b_2 \\
&& +\, 0.2817\,a_2b_2^3 + 0.3154\,a_1^2a_2^2 + 0.9684\,b_1 - 0.3450\,a_2 + 0.9796\,a_1b_1^2 \\
&& +\, 0.9411\,a_2^4 + 0.9193\,a_1b_1b_2^2 + 0.6704\,a_2^2b_1b_2 + 0.1101\,a_1b_1^3 \\
&& +\, 0.1491\,a_1b_1 - 0.6584\,a_1 + 0.9249\,a_1b_1b_2 + 0.6863\,a_1b_1^2b_2 + 0.3818\,a_1b_2^3 \\
&& +\, 0.0884\,a_2b_2.
\end{array}
\]
\bibliographystyle{plainnat}  % This is important for author-year citations
\bibliography{Isotropy_Groups_in_Optimization}   % references is the name of your .bib file without extension
\end{document}